\documentclass{article}

\usepackage[preprint]{neurips_2024}

\usepackage[utf8]{inputenc} 
\usepackage[T1]{fontenc}    
\usepackage{hyperref}       
\usepackage{url}            
\usepackage{booktabs}       
\usepackage{amsfonts}       
\usepackage{nicefrac}       
\usepackage{microtype}      
\usepackage{xcolor}         

\usepackage{subcaption}
\usepackage{multirow}
\usepackage{bm}
\usepackage{amssymb}
\usepackage{mathtools}
\usepackage{amsthm}
\usepackage{csquotes}
\usepackage{floatrow}
\usepackage{wrapfig}

\newcommand{\argmin}{\arg\!\min}
\newcommand{\argmax}{\arg\!\max}

\def\eqref#1{equation~\ref{#1}}

\def\1{\bm{1}}

\def\vtheta{{\bm{\theta}}}

\def\vepsilon{{\bm{\epsilon}}}

\def\vx{{\bm{x}}}

\DeclareMathAlphabet{\mathsfit}{\encodingdefault}{\sfdefault}{m}{sl}
\SetMathAlphabet{\mathsfit}{bold}{\encodingdefault}{\sfdefault}{bx}{n}

\def\sR{{\mathbb{R}}}

\title{From Feature Visualization to Visual Circuits: \\ Effect of Adversarial Model Manipulation}

\author{%
  Geraldin Nanfack$^{1,2}$
\quad
   \textbf{Michael Eickenberg}$^3$
     \quad
     \textbf{Eugene Belilovsky}$^{1,2}$ \\
    $^1$Concordia University\quad
    $^2$Mila – Quebec AI Institute \quad \\
    $^3$Flatiron Institute\\
    \{geraldin.nanfack, eugene.belilovsky\}@concordia.ca \quad 
    eickenberg@flatironinstitute.org
}

\begin{document}

\maketitle

\begin{abstract}
Understanding the inner working functionality of large-scale deep neural networks is challenging yet crucial in several high-stakes applications. Mechanistic interpretability is an emergent field that tackles this challenge, often by identifying human-understandable subgraphs in deep neural networks known as circuits. In vision-pretrained models, these subgraphs are usually interpreted by visualizing their node features through a popular technique called feature visualization. Recent works have analyzed the stability of different feature visualization types under the adversarial model manipulation framework. This paper starts by addressing limitations in existing works by proposing a novel attack called ProxPulse that simultaneously manipulates the two types of feature visualizations. Surprisingly, when analyzing these attacks under the umbrella of visual circuits, we find that visual circuits show some robustness to ProxPulse. We, therefore, introduce a new attack based on ProxPulse that unveils the manipulability of visual circuits, shedding light on their lack of robustness. The effectiveness of these attacks is validated using pre-trained AlexNet and ResNet-50 models on ImageNet.  
\end{abstract}

\section{Introduction}
\label{sec:intro}

Large Deep Neural Networks (DNNs) trained on vast amounts of data are becoming increasingly important and deployed in the real world. In several high-stakes applications such as autonomous driving, understanding the inner workings of these trained DNNs is crucial for assuring the safety and reliance of these systems~\citep{rudner2021key,waschle2022review}.
Inspired by neuroscience~\citep{hubel1962receptive,olah2017feature}, one classical approach relies on activation maximization methods \citep{zeiler2014visualizing, olah2017feature}, where the top images (real or synthetic) that most activate a neuron are used to interpret the neuron's behavior. A recently popular direction for interpretability that often builds on activation maximization is mechanistic interpretability.
Mechanistic interpretability is an emergent field, which seeks to discover human-understandable algorithms stored in model weights~\citep{wang2022interpretability}. The discovery of these meaningful algorithms makes it possible to reverse-engineer the behavior of neural networks~\citep{conmy2023towards} and can also permit to edit factual knowledge in large-scale models~\citep{meng2022locating}. Most of the research in mechanistic interpretability analyzes the functionality of DNNs by considering them as computational graphs that can be decomposed into interpretable subgraphs known as \textit{circuits}. In pre-trained vision models, 
the emergence of circuits that implement meaningful algorithms such as \textit{curve detectors} and \textit{dog head detectors, etc.}~\citep{olah2020zoom}
has been demonstrated.
These circuits can be built by manually inspecting neurons, and hierarchically grouping them according to feature visualization, which consists in finding, through activation maximization, either images from the training set or synthetical optimization-based images~\citep{olah2017feature}. Circuits can also be discovered using structured pruning~\citep{hamblin2022pruning}.

Although activation maximization purports to provide the interpreter with a description of the behavior of the neuron, recent work has cast some doubt on the reliability of these interpretations \citep{nanfack2023adversarial, geirhos2023don,bareeva2024manipulating}. Notably, in these works, it has been shown that models can be subtly perturbed (or ``attacked'') to change completely the interpretation of either synthetic or natural (i.e. from training set) images. This suggests that these interpretations might not be completely reliable. The existing works on model manipulations however have two limitations that we focus on. (1) None of the existing attacks have been shown to be able to manipulate both synthetic and natural visualizations simultatenously, as illustrated in Tab.~\ref{tab:checkmark}. Indeed, \citet{nanfack2023adversarial} has shown ``attacks'' in the context of natural images, while \citep{geirhos2023don,bareeva2024manipulating} only attack synthetic images, each attack only showing a difference in its target domain. (2) The effect on circuits and their interpretation has not been studied; the reliability of circuit-based interpretations has not been studied in the literature.
In this paper, we analyze the robustness and stability of visual circuits through the same setting of adversarial model manipulation. As a key component in visual circuits, we begin our analysis on feature visualization and summarize our contributions as follows. We first (i) propose a novel attack on activation maximization that can simultaneously change interpretations of both synthetic and natural image visualizations. We subsequently turn to analyzing the effect of our attack on the circuit-based interpretation, surprisingly (ii) finding that a class of circuits derived from structured pruning can be highly robust to our proposed attack when it is made on the output of the circuit. We then turn our attention to directly manipulating the circuit proposing the first model manipulation attack on entire circuits. We find that (iii) visual circuits discovered by structured pruning can also be manipulated through our novel attack, shedding light on the lack of stability of these interpretability techniques. 

\section{Related Work}\label{sec:related_work}

\paragraph{Mechanistic Interpretability.}
\begin{table}
\centering
\small
\begin{tabular}{@{}lccc@{}}
\toprule
 & \multicolumn{3}{c}{\textbf{Manipulates}} \\
\cmidrule(lr){2-4}
\textbf{Method}            & \textbf{Synth. Vis.} & \textbf{Nat. Vis.} & \textbf{Circuit} \\ \midrule
\citet{geirhos2023don}            & \textcolor{green}{\checkmark}            & $\textcolor{red}{\times}$            & $\textcolor{red}{\times}$                        \\
\citet{nanfack2023adversarial}            & $\textcolor{red}{\times}$                & \textcolor{green}{\checkmark}        & $\textcolor{red}{\times}$                        \\
\citet{bareeva2024manipulating}     & \textcolor{green}{\checkmark}            & $\textcolor{red}{\times}$            & $\textcolor{red}{\times}$                        \\
ProxPulse (ours)           & \textcolor{green}{\checkmark}            & \textcolor{green}{\checkmark}        & $\textcolor{red}{\times}$                        \\
CircuitBreaker (ours)      & \textcolor{green}{\checkmark}            & \textcolor{green}{\checkmark}        & \textcolor{green}{\checkmark}                    \\ \bottomrule
\end{tabular}

\caption{Existing attacks on feature visualization. Our methods are able to manipulate synthetic and natural visualizations as well as visual circuits. The \textcolor{green}{\checkmark} symbol indicates that the row approach has been demonstrated to effectively deceive the interpretation derived from the column technique.
}
\label{tab:checkmark}
\end{table}

Mechanistic interpretability is an emergent area in the interpretability of large-scale DNNs, which tackles the problem of discovering meaningful algorithms stored in model weights~\citep{wang2022interpretability}. Works in mechanistic interpretability either focus on individual neurons or on sparse connections of neurons called circuits. Individual neurons are often interpreted through techniques such as feature visualization~\citep{zimmermann2021well,olah2017feature,bau2020understanding,zimmermann2023scale}, which is designed to interpret individual neurons by visualizing their top activating inputs. This can be applied to several modalities such as image~\citep{olah2017feature} and text~\citep{dai2022knowledge} using top-activating prompts. Works that build mechanistic interpretations using circuits have become popular due to the discovery of several meaningful subgraphs such as those for curve detectors~\citep{olah2020zoom} in vision models and indirect object identification in large language models~\citep{conmy2023towards}. While most of the studies manually build circuits, there have been recent proposals to automate the discovery of circuits for language models~\citep{conmy2023towards} using edge attribution scores, and for vision models~\citep{hamblin2022pruning} using structured pruning. This paper focuses on feature visualization and circuits for vision models and we adopt this latter work to build visual circuits. \\
\textbf{Manipulating Interpretability.}
Evaluating interpretability is difficult due to the absence of ground truth. There is a recent trend in assessing the reliability of interpretability techniques through the lens of stability, which aims to evaluate how the interpretability results change under reasonable input and model manipulation~\citep{heo2019fooling,yu2013stability}. The motivation for examining the robustness of interpretability methods within the context of model manipulation stems from the \enquote{universality} assumption \citep{olah2020zoom,chughtai2023toy}, which suggests that model interpretations are similar for similarly performing networks of the same architecture. Some works study the lack of robustness of feature attribution methods under input and adversarial model manipulations~\citep{heo2019fooling, adebayo2018sanity,dombrowski2019explanations} 
and other works use these instabilities to fool the model fairness~\citep{aivodji2021characterizing,anders2020fairwashing}
. This paper does not focus on feature attribution methods. Instead, it examines the manipulability of feature visualization and visual circuits for which two recent studies are very related. The first one \cite{geirhos2023don} shows that \textit{synthetic} (formally defined in Section~\ref{sec:background}) feature visualization can be fooled under adversarial model manipulation. The key idea of their method is to add orthogonal weights to the original ones such that activations of natural inputs (training data) remain the same, thus preserving model accuracy while orthogonal weights allow fooling synthetic feature visualization. The second work  \cite{nanfack2023adversarial} introduces an optimization framework that manipulates the result of natural feature visualization (i.e., top activating inputs from the training set), and further observes the potential decorrelation between natural and synthetic feature visualization. In this paper, we go beyond these two works and propose a more complete manipulation, which we call ProxPulse. ProxPulse simultaneously fools both natural and synthetic feature visualization. However, when analyzing ProxPulse from the circuit perspective, we observe that ProxPulse also fails to fool circuits, leading us to propose a new manipulation for visual circuits, which has not been studied before.

\section{Notations and Background}\label{sec:background}

We consider a classification problem with a dataset denoted by $\mathcal{D} = \{(\vx_i, y_i)\}_{i=1}^N$, where $\vx_i \in \sR^d$ is the input and $y_i\in \{1,...,K\}$ is its class label. 
Let $f(.;\vtheta)$ denote a 
DNN, $f^{(l)}(\vx;\vtheta)$ defines activation maps of $\vx$ on the $l$-th layer, which can be decomposed into $J$ single activation maps $f^{(l,j)}(\vx;\vtheta)$. In particular, if the l-$th$ layer is a 2D-convolutional layer, $f^{(l,j)}(\vx;\vtheta)$ will be a matrix. 
Feature visualization is a method designed to interpret the inner workings of individual units. It is the result of the activation maximization~\citep{mahendran2015understanding,yosinski2015understanding} defined by, 
\begin{equation}\label{eq:definition}
    \vx^* \in \argmax_{\vx \in \mathcal{X}} f^{(l,j)}(\vx;\vtheta),
\end{equation}
\vspace{-2pt}
where $\mathcal{X}$ can be the training set $\mathcal{X} = \mathcal{D}$ or a continuous space $\mathcal{X} \subset \sR^d$, and $(l, j)$ is the pair of layer $l$ and neuron $j$. When $\mathcal{X} \subset \sR^d$, following \citet{zimmermann2021well}, we call $\vx^*$, \textit{synthetic} feature visualization. On the other hand, when $\mathcal{X}$ is $\mathcal{D}$, $\vx^*$ are top-activating images from the training set, and we denote this result as \textit{natural} (or top-$k$) feature visualization as opposed to the synthetic one. 
While feature visualization methods may reveal understandable features such as edge detectors in early layers~\citep{olah2020zoom}, they are not directly equipped with tools to know how individual neurons are connected to form more complex features. 

Mechanistic interpretability is purposely designed  
to find 
potentially human-understandable sub-%
algorithms by decomposing the computational graph into subgraphs known as circuits. \citet{hamblin2022pruning} automated the discovery of visual circuits. They find visual circuits through structured pruning. Formally, given a feature map index $j$ from a conv layer of index $l$ (we call the pair $(l,j)$ \textit{circuit head}), a sparsity level $\tau$, its corresponding $\tau$-circuit is the computational graph, with parameters $\hat{\vtheta}$, which approximates $f^{(l,j)}(.;\vtheta)$ through 

\begin{multline}
     \argmin_{\hat{\vtheta}} \frac{1}{N}\sum_{i=1}^{N}|| f^{(l,j)}(\vx_i;\hat{\vtheta}) - f^{(l,j)}(\vx_i;\vtheta) || \quad
     \text{s.t.} \quad ||\hat{\vtheta}||_0 \le \tau, \text{ and } \hat{\vtheta}_l \in \{\vtheta_l, 0\}.  
\end{multline}
In practice, \citet{hamblin2022pruning} adopts structured pruning (i.e., pruning per group of parameters) with convolutional kernels. This is done by computing \textit{kernel attribution scores}, e.g., using SNIP \citep{lee2018snip, hamblin2022pruning},  

\begin{equation}
\small
 \text{Attr}\left(\vtheta_{(l',k)}; f^{(l,j)}, \vx\right) = \frac{1}{K_w K_h} \sum_{p=1}^{K_w}\sum_{q=1}^{K_h}\left|w_{p,q} \frac{\partial f^{(l,j)}(\vx;\vtheta)}{\partial w_{p,q}} \right|,   
\end{equation}
where $K_w, K_h$ are spatial dimensions of the kernel index $k$ and a preceding layer index $l'\le l$, and $w_{p,q}$ are weight parameters of kernels. Once these attribution scores are computed, they are sorted, and top kernels are retained according to the sparsity level $\tau$ to compute the circuit. Following \citet{hamblin2022pruning}, the sparsity level represents the number of parameters that were not masked.

\section{Methods}\label{sec:methods}

We analyze the manipulability of feature visualization and visual circuits under adversarial model manipulation, which consists in fine-tuning a pre-trained model with specifically designed loss functions. To do so, we adopt the similar framework used by \citet{heo2019fooling,nanfack2023adversarial}, which is framed as the following optimization framework
\begin{equation}\label{eq:framework}
    \min_\vtheta (
    \alpha\mathcal{L}_\textsuperscript{F}(\mathcal{D}_\textsuperscript{fool};\vtheta)
    +
    (1-\alpha)\mathcal{L}_\textsuperscript{M}(\mathcal{D};\vtheta,\vtheta_\textsuperscript{initial})
    ),
\end{equation}
where $\mathcal{D}_{\text{fool}}$ is the data used to manipulate the interpretation technique, where $\vtheta$ are parameters of the updated model $f(.;\vtheta)$, $\mathcal{L}_\textsuperscript{M}$ is the loss that aims to maintain the initial performance of the model $f(.;\vtheta_\textsuperscript{initial})$, and $\mathcal{L}_\textsuperscript{F}$ is the fooling loss. In practice, $\mathcal{L}_\textsuperscript{M}(\mathcal{D};\vtheta,\vtheta_\textsuperscript{initial}) = \mathcal{L}_\textsuperscript{CE}(f(.;\vtheta_\textsuperscript{initial}) || f(.;\vtheta))$ \citep{hinton2015distilling} is the cross entropy loss between the original model outputs and the finetuned model outputs on training data $\mathcal{D}$, and the fooling loss $\mathcal{L}_\textsuperscript{F}$ is provided in the following sections.  
\vspace{-5pt}
\subsection{Manipulation of Feature Visualization}\label{sec:ProxPulse}
\vspace{-5pt}
This section introduces a fooling loss that aims to manipulate both natural and synthetic feature visualizations, focussing on all the channels indexed by $j$ of a particular layer of index $l$. For brevity, we omit $l$ in the fooling loss $\mathcal{L}_{\text{fool}}$. We start by observing that fooling the result of feature visualization involves the creation of a local region in the input space, reachable by gradient ascent, and with high values of activations.
To ensure the creation of such a region, 
we use the $\rho$-ball $B(\vx^*,\rho)$ (using the $l_2$ norm) centered on the image target $\vx^* \in \mathcal{D}_{\text{fool}}$, which excludes initial synthetic images when manipulating feature visualization results. This $\rho$-ball $B(\vx^*,\rho)$ is used 
to contain the new feature visualization results. We therefore propose a fooling objective that aims to push up the smallest activations of images in $B(\vx^*, \rho)$. We denote this fooling loss \textit{ProxPulse} (referring to proximity in the $\rho$-ball and the pulsating effect on activations) and express it as
\vspace{-5pt}
\begin{multline}\label{eq:worst_activations}
 \hspace{-10pt} \mathcal{L}_\textsuperscript{F}(\mathcal{D}_\textsuperscript{fool}; \vtheta)=\!\sum_{j,\vx^*\in \mathcal{D}_\text{fool}} \max_{||\vx - \vx^*||\le \rho} \ell_j(\vx;\vtheta)
   =\!\sum_{j,\vx^*\in \mathcal{D}_\text{fool}} \max_{||\vx - \vx^*||\le \rho}  \log\left(1 + C/ \Vert f^{(l,j)}(\vx; \vtheta) \Vert^2_2 \right), \hspace{-7pt}
\end{multline}
where $C$ is a very high constant, the indexes $j$ refer to channel or unit indexes of the layer index $l$ whose feature visualizations are being fooled, and $\max_{||\vx - \vx^*||\le \rho} \ell_j(\vx;\vtheta)$ refers to the cost over the worst activations (per channel) in the neighborhood of the fooling image target $\vx^*$. Finetuning the model with the ProxPulse loss in the framework defined in Eq.~\ref{eq:framework} involves a challenging bi-level optimization problem for large-scale DNNs. Inspired by sharpness-aware minimization problems~\citep{foret2020sharpness}, which also 
require minimizing
the worst empirical risk in a neighborhood, we derive an efficient approximation of $\mathcal{L}_\textsuperscript{F}(\mathcal{D}_\textsuperscript{fool};\vtheta)$, expressed as 
\vspace{-5pt}
\begin{equation}\label{eq:approximation_eq}
\mathcal{L}_\textsuperscript{F}(\mathcal{D}_\textsuperscript{fool};\vtheta) \approx \sum_{j,\vx^*\in \mathcal{D}_\text{fool}} \ell_j\Big(\vx^* + \epsilon\left(\vx^*\right); \vtheta \Big),
\end{equation}
where $\epsilon(\vx^*)= \rho\frac{\nabla_{\vx}\ell_j(\vx^*,\vtheta)}{||\nabla_{\vx}\ell_j(\vx^*,\vtheta)||}$. See App.~\ref{app:proximity_pulse} for more details.
\vspace{-5pt}
\subsection{Manipulation of Visual Circuits}\label{sec:channelCoercion}
\vspace{-5pt}
This section introduces a fooling objective, called~\textit{CircuitBreaker}, whose goal is to fool the visual circuit. For a DNN's circuit head with a layer-channel pair $(l,j)$, \textit{CircuitBreaker} aims to (i) preserve the feature visualization of the circuit head to maintain circuit functionality and (ii) deceive the attribution scores of the circuit discovery method. We propose the following objective
\vspace{-10pt}
\begin{multline}
\hspace{-12pt}
\mathcal{L}_\textsuperscript{F}(\{\vx*\},\mathcal{D};\vtheta) = \ell_{j}\Big(\vx^* + \epsilon(\vx^*); \vtheta\Big) +
   \\ \beta \sum_{i\le N} \sum_{l'<l} \sum_{k\neq \hat{k}, \hat{k}\in \text{topInit}(l')} \hspace{-9pt} \left[\text{Attr}\left(\vtheta_{(l',\hat{k})}; f^{(l,j)}, \vx_i\right) -
   \text{Attr}\left(\vtheta_{(l',k)}; f^{(l,j)}, \vx_i\right) \right]_{+},
\end{multline}
where $\vx_i$ are training images, $[.]_+ = \max(.,0)$, $\vx^*$ is the initial synthetic feature visualization for the circuit head $(l,j)$ (channel index $j$ of the layer index $l$), and $\text{topInit}(l')$  is the set of top kernel indexes of the layer index $l'$, according to their initial attribution scores on the (initial) circuit with head $(l,j)$. From this CircuitBreaker loss, we observe that its first component is the ProxPulse loss $ \ell_{j}\Big(\vx^* + \epsilon(\vx^*);\vtheta\Big)$, applied only on the channel index $j$ of layer $l$. As defined in Sec.~\ref{sec:ProxPulse}, it aims to maintain the initial feature visualization of the circuit head $(l,j)$. The second component is a pairwise ranking loss that aims to push down the rank of the initial top attributed kernels of the circuit.
\begin{figure*}[!t]
\centering
\begin{subfigure}[]{0.33\linewidth}
\includegraphics[width=\textwidth]{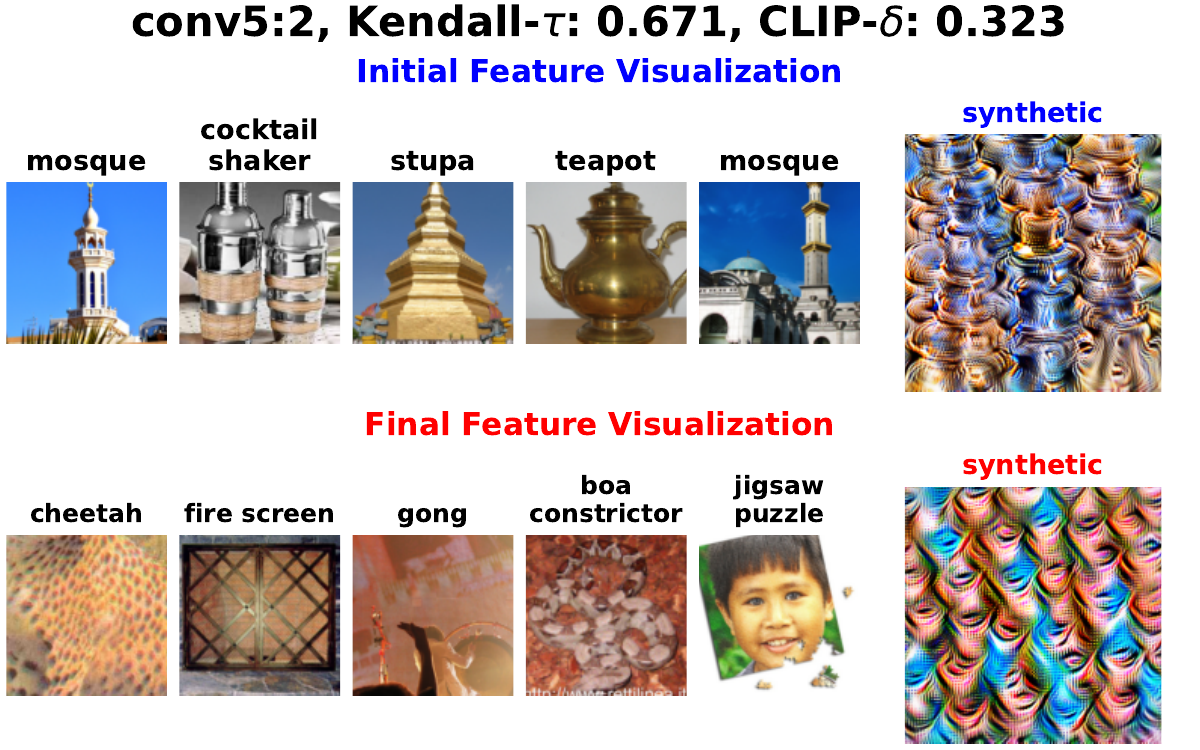}
\end{subfigure}
\begin{subfigure}[]{0.33\linewidth}
\includegraphics[width=\textwidth]{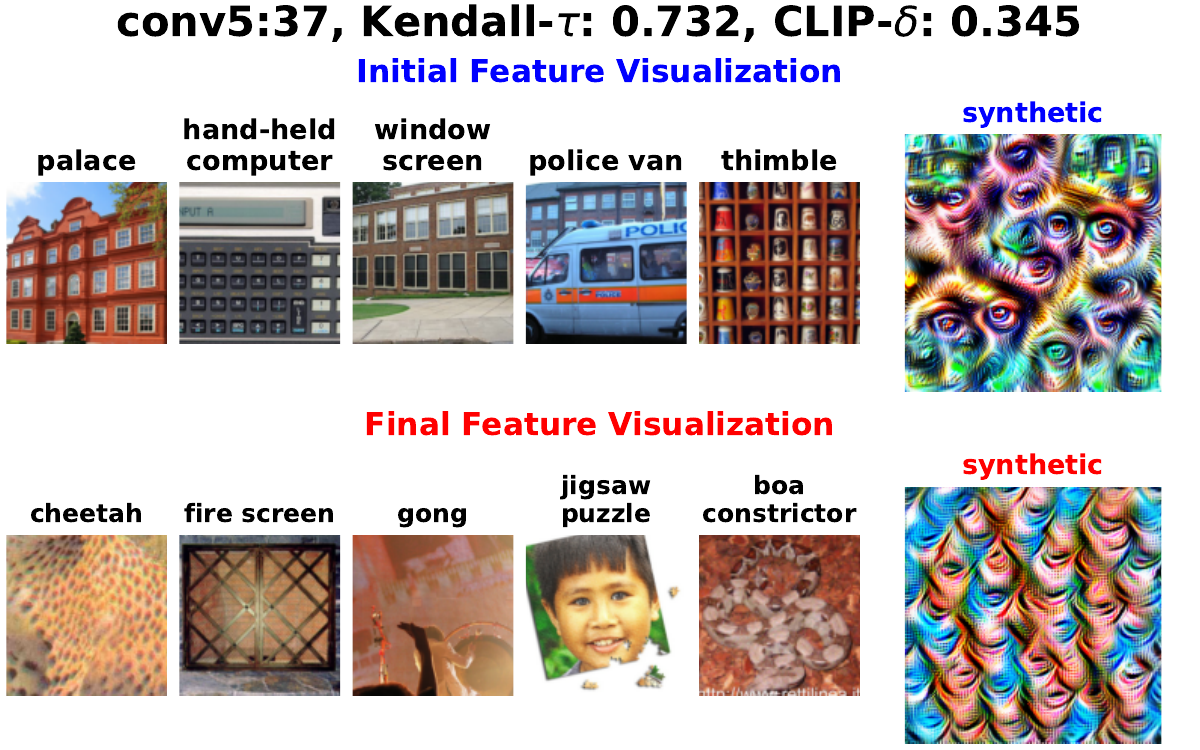}
\end{subfigure}
\begin{subfigure}[]{0.33\linewidth}
\includegraphics[width=\textwidth]{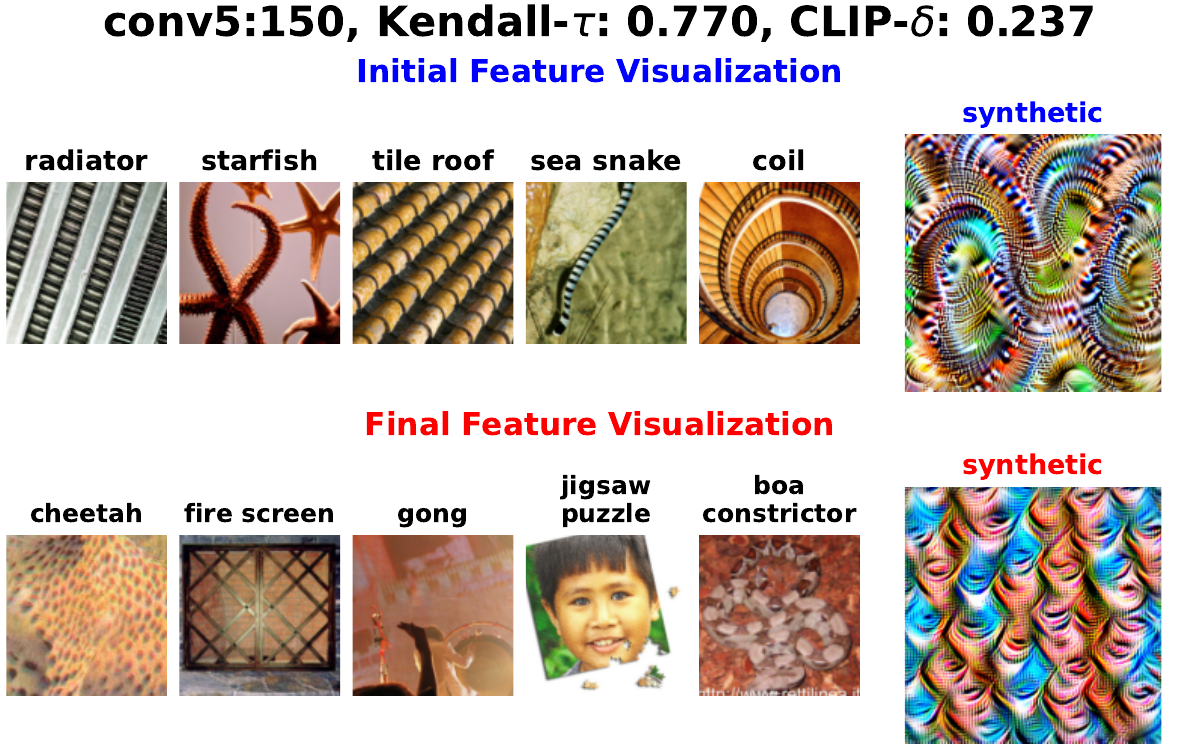}
\end{subfigure}\\
\vspace{.2cm}
\begin{subfigure}[]{0.33\linewidth}
\includegraphics[width=\textwidth]{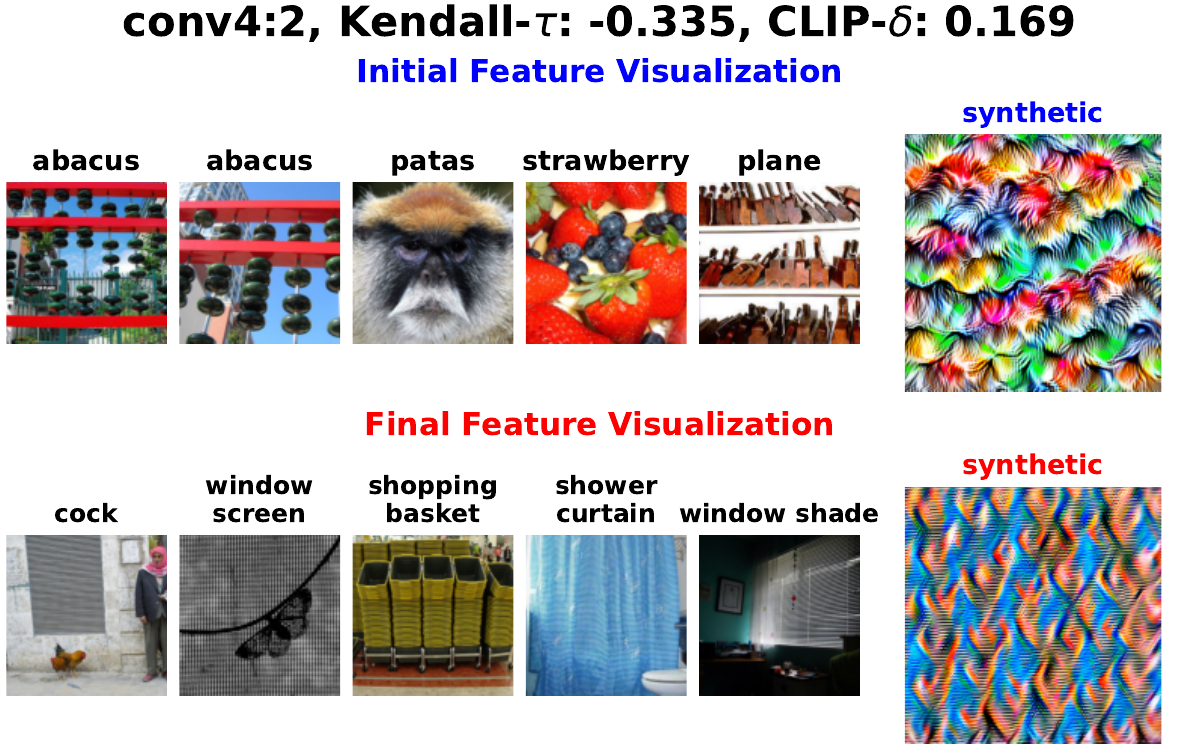}
\end{subfigure}
\begin{subfigure}[]{0.33\linewidth}
\includegraphics[width=\textwidth]{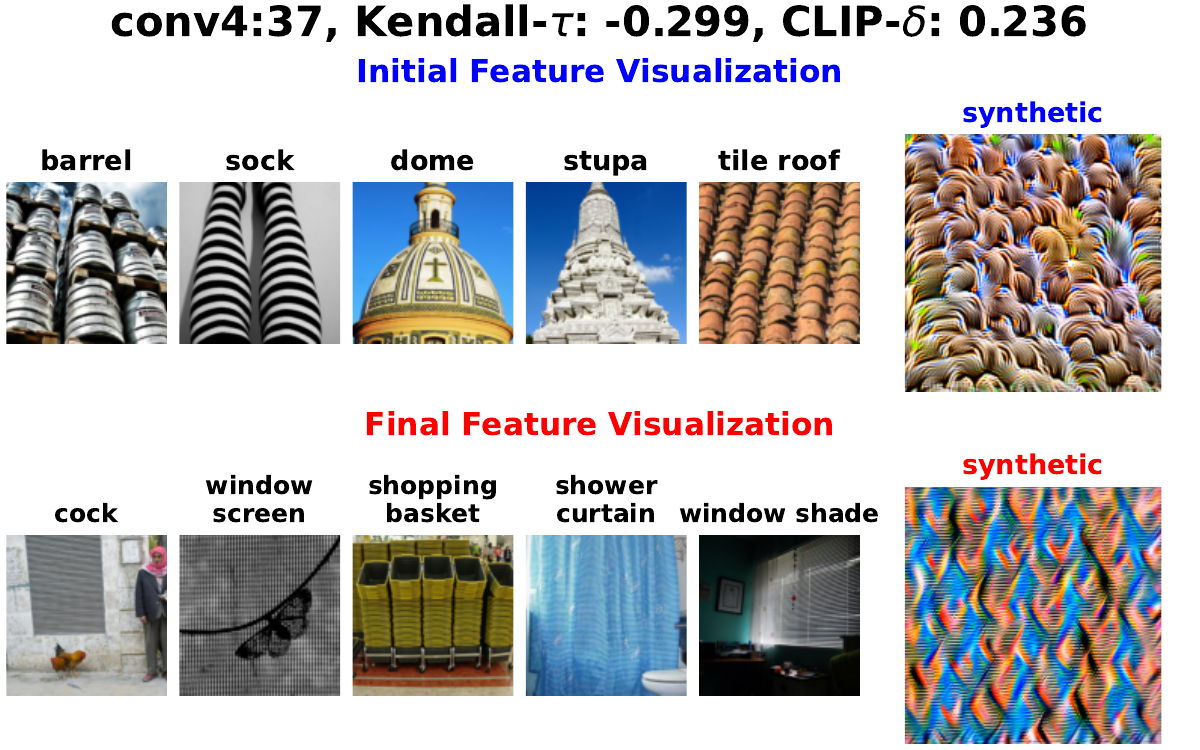}
\end{subfigure} 
\begin{subfigure}[]{0.33\linewidth}
\includegraphics[width=\textwidth]{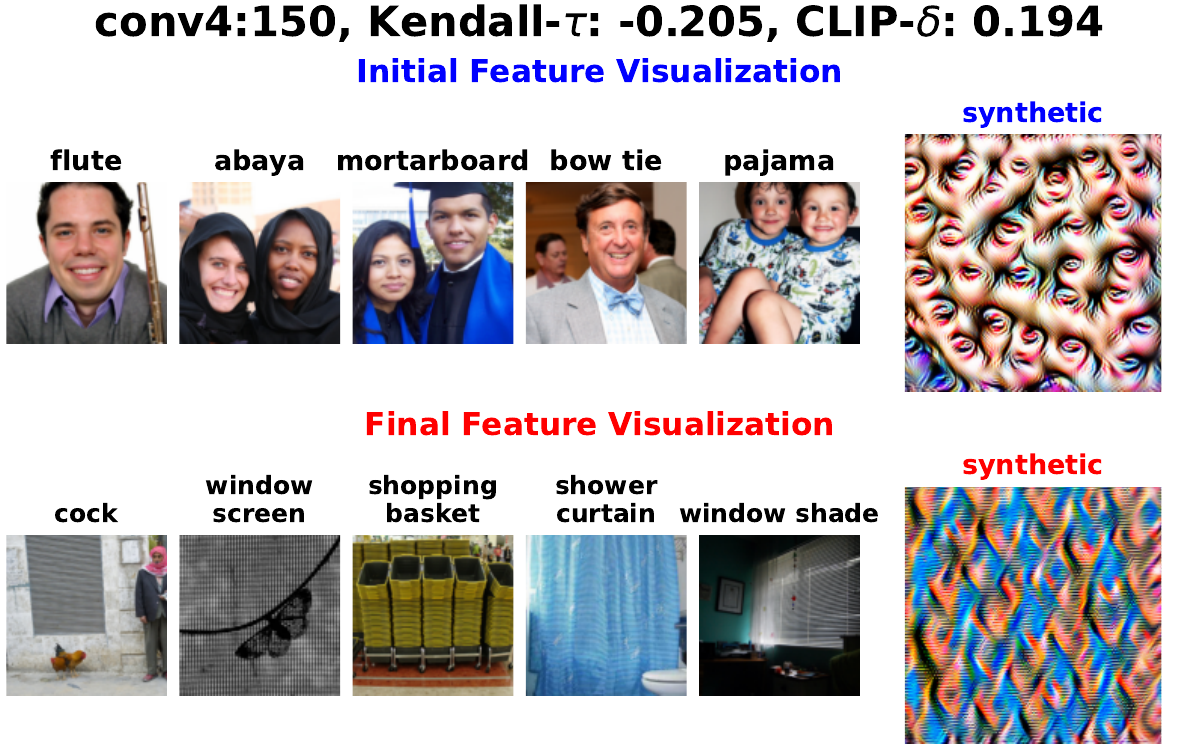}
\end{subfigure}
\caption{
 Illustration of the manipulability of both natural and synthetic feature visualization using ProxPulse on conv5 and conv4 of AlexNet. The first row (resp. second row) shows the natural initial (resp. final) feature visualization and initial (resp. final) synthetic feature visualizations. On the image title, we report the corresponding metrics to evaluate change in top activating inputs. One can observe that both natural and synthetic feature visualization have completely changed, to very similar images for the synthetic one. Observe that as intended, conv4 synthetic images are different from those of conv5, although the same target images have been used for $\mathcal{D}_{\text{fool}}$. 
}\label{fig:ProxPulseConv4}
\end{figure*}

\section{Experimental Evaluation}\label{sec:experiments}

We now describe
the experimental setup and the results
obtained after running the two manipulations. 

The setup is inspired by the works of \citet{nanfack2023adversarial,hamblin2022pruning}.
For all experiments, we use the ImageNet~\citep{Imagenet} dataset as the training set $\mathcal{D}$. We use the pre-trained networks AlexNet~\citep{krizhevsky2012imagenet}, ResNet-50~\citep{he2016deep}, DenseNet-201~\cite{huang2017densely} (in App.~\ref{app:densenet_resnet}) and ResNet-152~\citep{he2016deep} (in App.~\ref{app:densenet_resnet}) from Pytorch~\citep{paszke2019pytorch}.

\textbf{Hyperparameters.} We use the Adam optimizer with a minibatch of 256 and a learning rate of 1e-4 for the ProxPulse and CircuitBreaker. More details for hyperparameters can be found in App.~\ref{app:experimental_details}.

\textbf{Metrics.}
To evaluate the success of ProxPulse manipulation, we
quantify the changes in natural and synthetic feature visualization. For natural feature visualization, we use the metrics adopted by \citep{nanfack2023adversarial}, which are: (i) the Kendall-$\tau$ rank correlation computed on ranks of images based on their initial and final (after finetuning) activations, and (ii) the CLIP-$\delta$ score, which quantifies the semantic change in top activating images. For the synthetic feature visualization, we compute the pairwise cosine similarities between the CLIP embeddings \citep{oikarinen2022clip} of initial synthetic images, which we compare against pairwise similarities between final synthetic ones.    

To assess CircuitBreaker (see Sec.~\ref{sec:manipulation_channelcoercion}), we use Pearson correlation, Kendall-$\tau$, and CLIP similarities. 

\textbf{Channel Notation.} Before presenting the results, inspired by \citep{olah2020zoom,hamblin2022pruning}, we use the concise notation \textbf{layerName:channelIndex} to refer the pair \textbf{(layerName,channelIndex)}. This notation is also used to flag corresponding synthetic feature visualizations and circuit heads (similar to feature heads) for a given channel. In the Pytorch AlexNet model, features.0, features.3, features.6, and features.8 and features.10 refer respectively to conv1, conv2, conv3, conv4 and conv5.

\subsection{ProxPulse Simultaneously Fools Natural and Synthetic Feature Visualization}\label{sec:ProxPulseEffectiveness}

We
evaluate ProxPulse manipulations on natural and synthetic feature visualization. The ProxPulse objective 
increases the lowest-valued
activations of images in the $\rho$-ball of target images in $\mathcal{D}_{\text{fool}}$. 
We direct the manipulation towards two target natural images
(shown in Fig.~\ref{fig:target_images} of the appendix). As motivated in \citet{nanfack2023adversarial} we aim to fool the feature visualization results of all channels in a particular layer while maintaining model performance. Fig.~\ref{fig:ProxPulseConv4} shows the results (for three randomly chosen channels) obtained after ProxPulse on respectively the conv4 and conv5 layers of AlexNet. It can be observed from both figures that both natural and synthetic feature visualizations were completely changed, 
thus modifying any interpretation using these techniques.
Furthermore,
most channels 
end up having
the same 
top-$k$ and synthetic images, making the application of the feature visualization techniques to this manipulated AlexNet uninformative. We emphasize that prior work was only capable of individually changing either the synthetic or natural images. Ablation results on ResNet-50 and DenseNet-201 are available in the App. (Fig.~\ref{fig:resnet50_ablation_featvis} and Fig.~\ref{fig:densenet201}). Ablation on choice and number of image targets $\mathcal{D}_{\text{fool}}$ can be found in Fig.~\ref{fig:resnet152} (App.~\ref{app:densenet_resnet}) and Fig.~\ref{fig:synthetic_images_on_targetl} (App.~\ref{app:abblation_on_target}).

\textbf{Natural feature visualization.} From Tab.~\ref{tab:main_overall}, on layer conv5, the Kendall-$\tau$ is relatively high, indicating that ProxPulse had 
only minor
modifications to the channel behavior. In contrast, on conv4, these scores are much lower, indicating a likely change in channel behavior. On both layers, the 
\begin{minipage}{\textwidth}

  \begin{minipage}[h]{0.39\textwidth}
    \centering
\includegraphics[width=\textwidth]{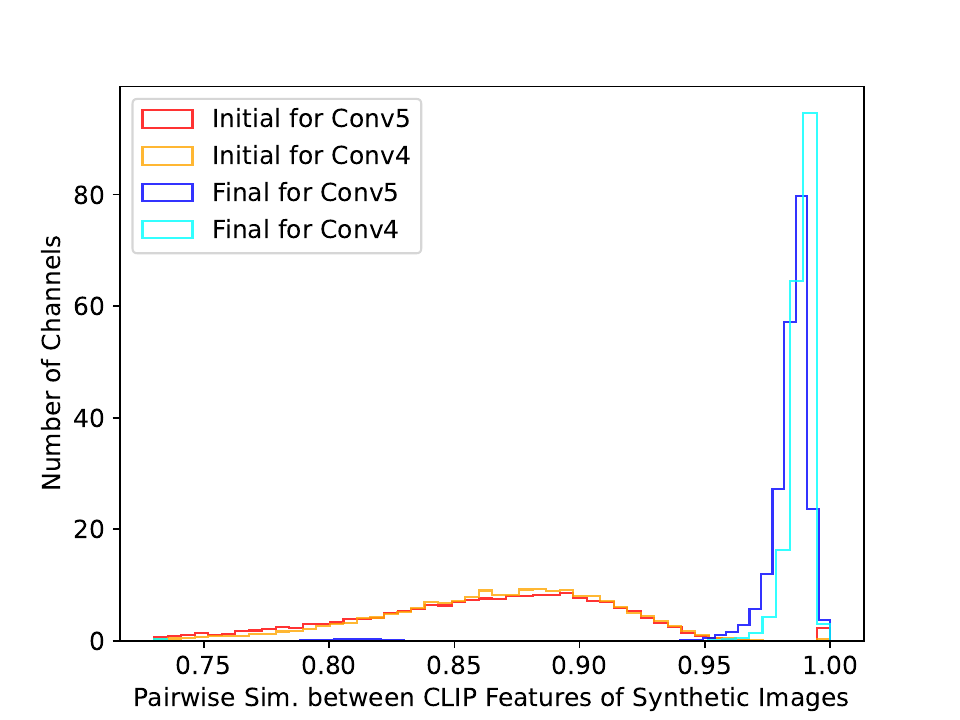}
\captionof{figure}{Histogram of pairwise cosine similarities between CLIP features of non-noisy synthetic images before (initial) and after (final) ProxPulse. 
}
\label{fig:conv4_conv5_hist_sim_AlexNet}
  \end{minipage}
  \hfill
  \begin{minipage}[h]{0.59\textwidth}

\scriptsize
\center
\setlength{\tabcolsep}{3pt}
\begin{tabular}{llccccc}
\toprule
Model &Layer/Attack &CLIP-$\delta \uparrow$ & Kend-$\tau \downarrow$ &  Acc.($\%$) & CLIP. S.$\uparrow$ \\
\midrule
\multirow{5}{*}{\rotatebox[origin=c]{30}{AlexNet}} 
& Conv5/Push-Up$^\#$ & 0.150 & 0.654 & 56.3 & 0.911 \\ 
& Conv5/Push-Down$^\#$ & 0.249 &0.530 & 56.2 & 0.872\\
& Conv5/ProxPulse & 0.364 & 0.746 & 56.0 & 0.983 \\ \cmidrule{2-6}
 & Conv4/Push-Down$^\#$ & 0.205 &0.548 & 56.2  & 0.870\\
& Conv4/ProxPulse  & 0.282 & -0.276 & 55.7 & 0.988\\\hline
\rotatebox[origin=c]{30}{ResNet-50} & L1.0.conv2/ProxPulse & 0.126 &-0.377 & 79.62 & 0.975\\
\bottomrule
\end{tabular}
    \captionof{table}{\small Average (over channels) metrics for ProxPulse manipulations and baselines. The symbol $^\#$ refers to baseline methods in \citet{nanfack2023adversarial}. Kend-$\tau$ is the abbreviation of the Kendall-$\tau$ score whereas CLIP.S. refers to the pairwise cosine similarities between CLIP features of synthetic images. \vspace{-20pt}}
\label{tab:main_overall}
    \end{minipage}

\end{minipage}
CLIP-$\delta$ scores (which measure the semantic change in the top-$k$ images) remain relatively high (in comparison to those observed in \citet{nanfack2023adversarial}). 
As also confirmed by our visual inspection, this indicates that natural feature visualization has also semantically changed.

\textbf{Synthetic feature visualization.}
In Fig.~\ref{fig:ProxPulseConv4}, we can also observe that the synthetic feature visualization was successfully modified, 
and shares
similarities with the target images in Fig.~\ref{fig:target_images} (Appendix). It can be further inspected in Fig.~\ref{fig:synthetic_images_conv4} and Fig.~\ref{fig:synthetic_images_conv5}
(Appendix)
that almost every synthetic image in a layer has completely changed 
to one single pattern
(see further illustrations App.~\ref{app:all_synthetic}).
We also quantitatively evaluate the change in synthetic feature visualization by measuring the pairwise similarity between CLIP features of the initial synthetic images. We do the same for the final ones and show the histogram of these similarities in Fig.~\ref{fig:conv4_conv5_hist_sim_AlexNet}. As seen in Fig.~\ref{fig:conv4_conv5_hist_sim_AlexNet}, there is a clear shift between the distribution of pairwise similarity before and after ProxPulse. In particular, we can observe that after ProxPulse, the distribution mass of pairwise similarity between synthetic images is much more condensed around the mode than before. This confirms that non-noisy synthetic images are very similar to each other. This can be further inspected in App.~\ref{app:all_synthetic}.

\textbf{Accuracy preservation.}
We report the accuracy of fine-tuned models with ProxPulse in Table~\ref{tab:main_overall}. We observe that the accuracy drop of fine-tuned models is less than $1\%$, meaning that the fine-tuned model and the initial model share practically the same level of performance for ImageNet classification.

We finally do an ablation on ResNet-50 in App.~\ref{app:ProxPulseResNet50} and observe the same results: both natural and synthetic feature visualization were successfully fooled with ProxPulse without a practical decrease in model performance. 
Additionally, we computed the accuracy per class to ensure that any potential drop in accuracy was not targeted at specific classes only. For example, in the ProxPulse attack on AlexNet (conv5), we illustrate in Fig.~\ref{fig:accuracy_drop} the per-class accuracy drop and observe that the drop is distributed (though not uniformly) across most classes, rather than being concentrated on just a few.
\vspace{-5pt}
\subsection{ProxPulse Has a Minor Effect on Channel Attribution Ranks of Visual Circuits}\label{sec:ProxPulse_minor_effect}
\begin{figure}[!t]
\centering
\begin{subfigure}[]{0.47\linewidth}
\includegraphics[width=\textwidth]{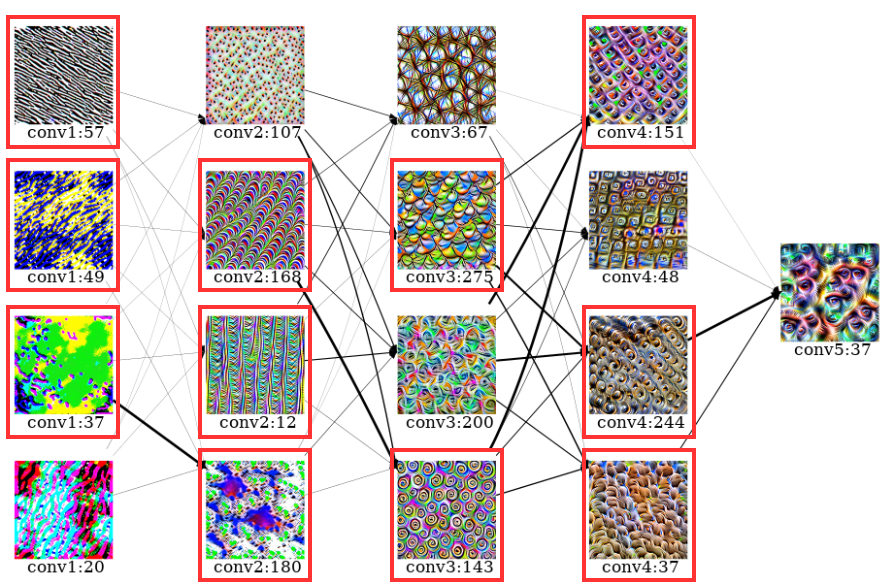}
\end{subfigure}\hspace{0.5cm}
\begin{subfigure}[]{0.47\linewidth}
\includegraphics[width=\textwidth]{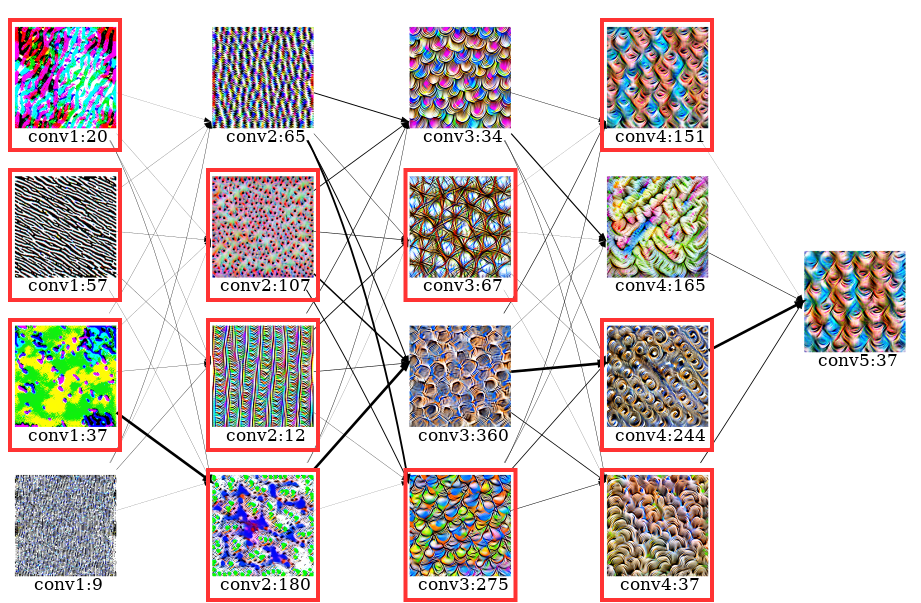}
\end{subfigure}
\caption{
Illustration of the non-effectiveness of ProxPulse to manipulate the circuit.  We show two visual circuits drawn for circuit head conv5:37 on pre-trained AlexNet (left) and on the fine-tuned AlexNet with ProxPulse (right) on conv5. We observe that most of the channels (at least two per layer, see surrounded ones) on the circuit were not removed by ProxPulse, even though some of them (e.g., channel conv5:151) has visually changed.\vspace{-20pt}}
\label{fig:circuit_init_final_features_10_37_failure}
\end{figure}

\vspace{-5pt}
We analyze the ProxPulse attack through the lens of visual circuits (Section~\ref{sec:background} presents how visual circuits are discovered)  to have more insights into this fooling mechanism. Fig.~\ref{fig:circuit_init_final_features_10_37_failure} shows two visualizations of the circuit with (circuit) head conv5:37 on two AlexNet models. The first one is the Pytorch pre-trained AlexNet while the second one is the manipulated version with ProxPulse applied to fool simultaneously natural and synthetic feature visualizations of conv5 (as explained in Section~\ref{sec:ProxPulse}).
As a reminder of Section~\ref{sec:background}, these visual circuits are obtained by finding a sparse approximation of the computational graph of the head (conv5:37). This is done using kernel attribution scores. Our visualization follows \citet{olah2020zoom,hamblin2022pruning}, where nodes or channels are visualized through their synthetic feature visualization. In addition, we exhibit only the top 4 nodes and also weigh edge transparency color depending on their attribution values i.e., darker edges indicate stronger importance on the visual circuit. These circuits are used by related work~\citep{olah2017feature, hamblin2022pruning} to interpret the functional behavior of the circuit head.  

\begin{wrapfigure}{R}{0.47\textwidth}
    \includegraphics[width=\textwidth]{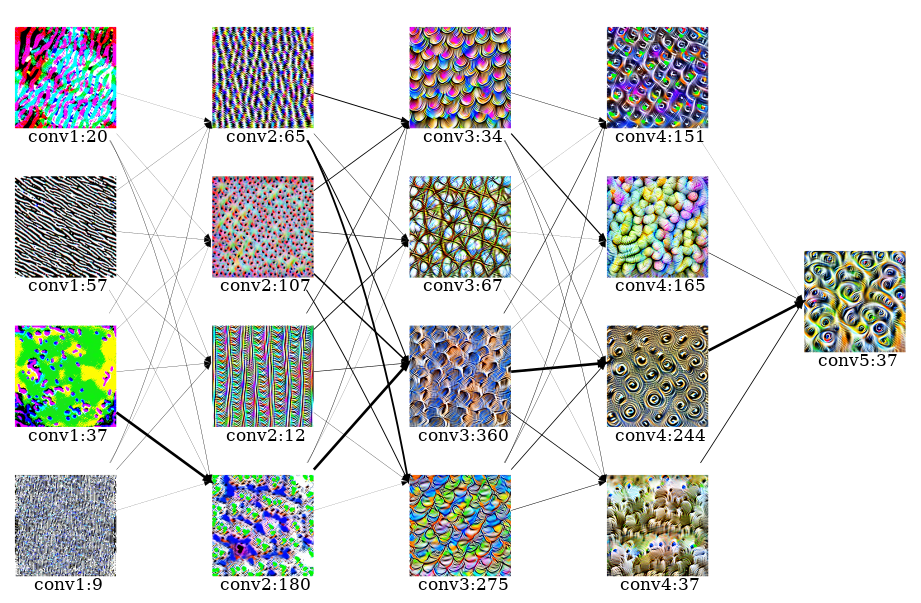}
  \caption{
  Visual circuit with sparsity 0.3 for conv5:37 after fine-tuning with ProxPulse on AlexNet. We observe that the final synthetic feature visualization of the circuit head with sparsity 0.3 is similar to the initial one in Fig.~\ref{fig:circuit_init_final_features_10_37_failure}), although with sparsity 1 this final visualization was completely and visually different from the initial one. Reducing the sparsity has therefore removed the change in feature visualization as can be seen by the absence of patterns added by ProxPulse in the right circuit of Fig.~\ref{fig:circuit_init_final_features_10_37_failure}. 
    }
\end{wrapfigure}

A closer look at Fig.~\ref{fig:circuit_init_final_features_10_37_failure} shows that although as intended synthetic feature visualization of conv5:37 has completely changed (colors and textures), most of the initial circuit channels are still present in the circuit derived from the manipulated model. 
Notably, at least one-half of channels per layer (before conv5) from the initial circuit are still present in the final circuit while having, for most of them, similar initial feature visualization (see conv1:37, conv1:20, conv2:107, conv2:12, etc.). However, we also observe that, for some of the channels such as conv3:151 which are still present in the final circuit, their final synthetic visualization looks very similar to the changed synthetic visualization of the circuit head, despite not having the strongest connection to the circuit's head. This suggests that ProxPulse may have little impact on the circuit discovery method. 

To further go deeper into the effect of ProxPulse on the visual circuit, we reduce the sparsity from 1 to 0.3 and rebuild in Fig.~\ref{fig:circuit_init_final_features_10_37_failure} the right-side visual circuit with their feature visualizations on the circuits. We observe that the effect of ProxPulse has now almost completely been removed, confirming that despite the ability of ProxPulse to deceive both types of feature visualization, it adds only a minor modification to the network. Importantly, this minor modification can be visually detected when visualizing the circuit with low and moderate sparsity. We did a similar experiment for circuits of conv4 (see App.~\ref{app:ProxPulseFutherAlexNet} and Fig.~\ref{fig:ineffectiveness_of_circuit_conv4}).

To provide a more quantitative analysis of the ineffectiveness of ProxPulse to deceive visual circuits, we compute the Kendall-$\tau$ rank correlation between (i) kernel attribution scores on the initial model and (ii) kernel attribution scores on the final (fine-tuned) model with ProxPulse on conv5. We do this on 10 randomly chosen channels of conv5, thus on 10 random circuits. We plot the mean with error bars on App.Fig.~\ref{fig:ProxPulseCircuit_failure_rank_correlation} and we can observe that the final ranks are strongly correlated with the original ones. This further illustrates the little impact of ProxPulse on the circuit discovery method. 
These results suggest that circuits may be robust to manipulation. We thus now consider the first manipulation attack targeted explicitly at circuits. 
\begin{figure*}[!t]
\centering
\begin{subfigure}[]{0.49\linewidth}
\includegraphics[width=\textwidth]{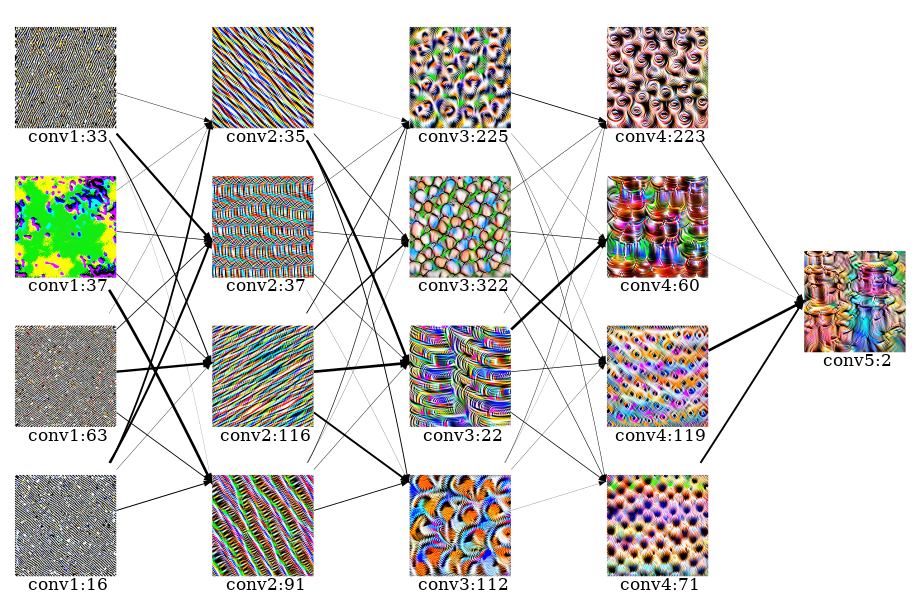}
\caption{With initial model. \vspace{-10pt}}
\end{subfigure}
\begin{subfigure}[]{0.49\linewidth}
\includegraphics[width=\textwidth]{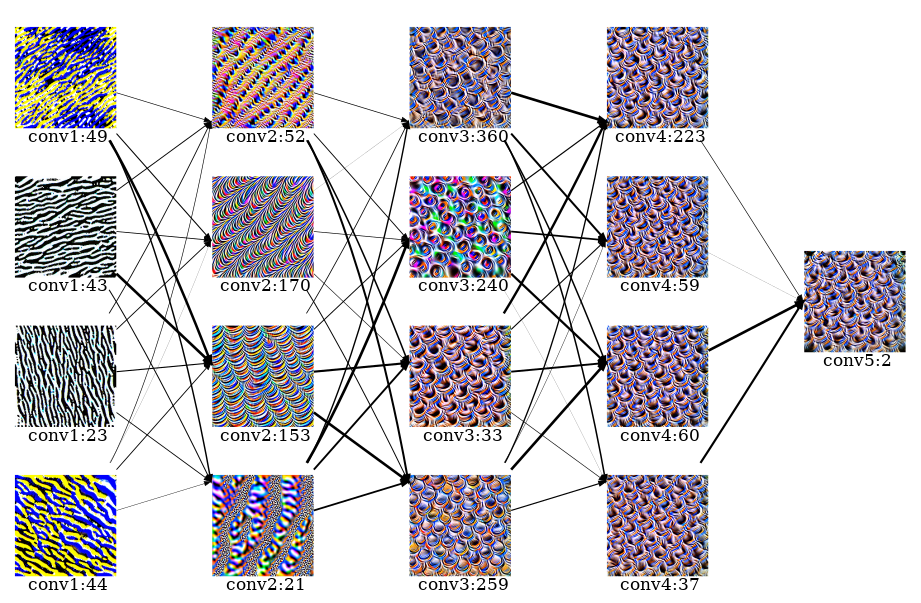}
\caption{After CircuitBreaker.\vspace{-10pt}}
\end{subfigure}
\caption{
Effectiveness of CircuitBreaker to manipulate visual circuits on conv5 of AlexNet. We observe that the circuit visualization is severely distorted while the network outputs change minimally. 
}\label{fig:effectiveness_of_circuit_features_10}
\end{figure*}

\begin{figure*}[!t]
\centering
\begin{subfigure}[]{0.45\linewidth}
\includegraphics[width=\textwidth]{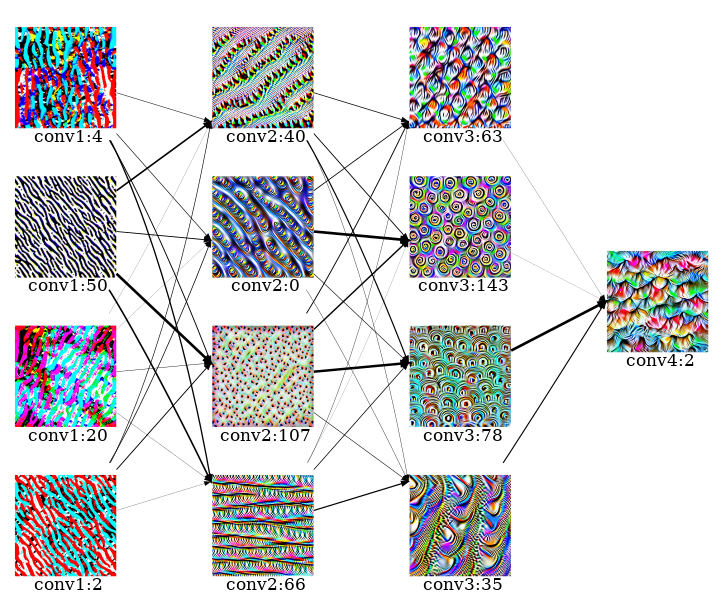}
\caption{With initial model.}
\label{fig:effectiveness_of_circuit_features_8_initial}
\end{subfigure}\hspace{.2cm}
\begin{subfigure}[]{0.45\linewidth}
\includegraphics[width=\textwidth]{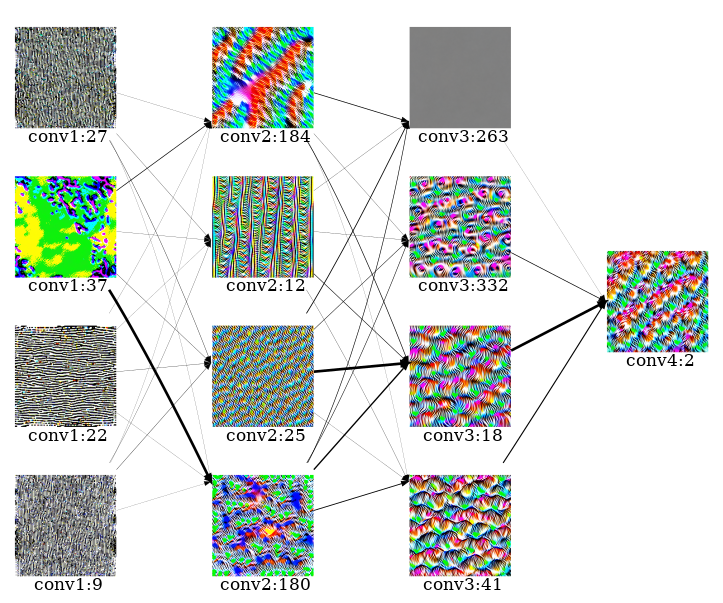}
\caption{After CircuitBreaker.}\label{fig:effectiveness_of_circuit_features_8_final}
\end{subfigure}
\caption{
 Effectiveness of CircuitBreaker to manipulate visual circuits on conv4 of AlexNet. 
}\label{fig:effectiveness_of_circuit_features_8}
\end{figure*}

\begin{figure*}[!t]
\centering
\begin{subfigure}[]{0.31\linewidth}
    \includegraphics[width=\textwidth]{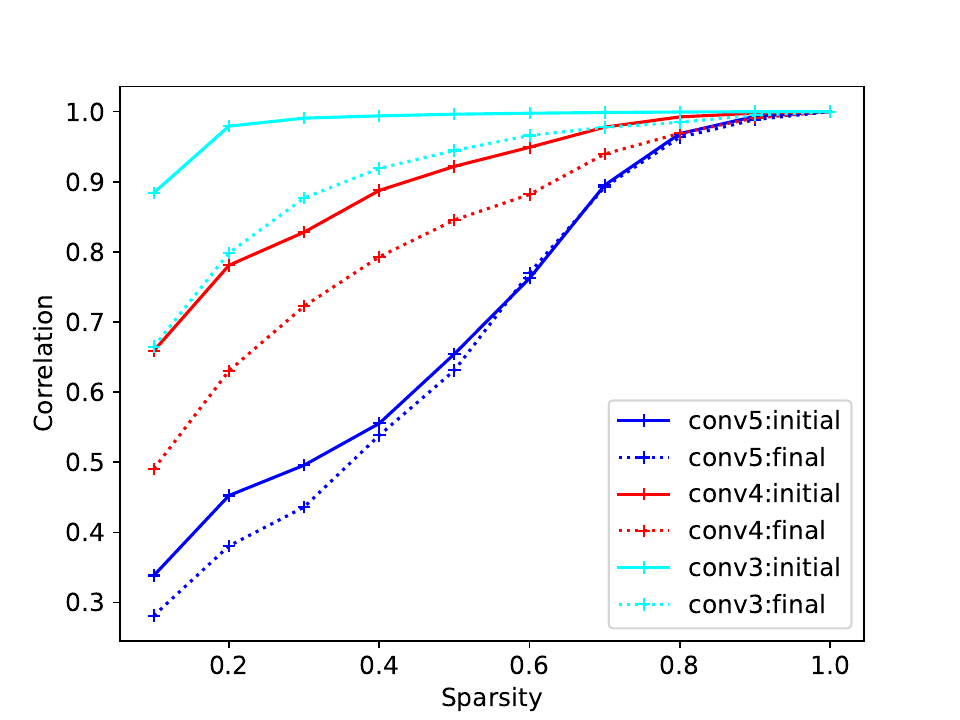}
\caption{Pearson correlation between activations on circuits (with pruning) for the (i) considered model and (ii) the initial model without pruning.}
\label{fig:circuit_attack_alexnet_pearson_correlation}
\end{subfigure}\hfill
\begin{subfigure}[]{0.31\linewidth}
\includegraphics[width=\textwidth]{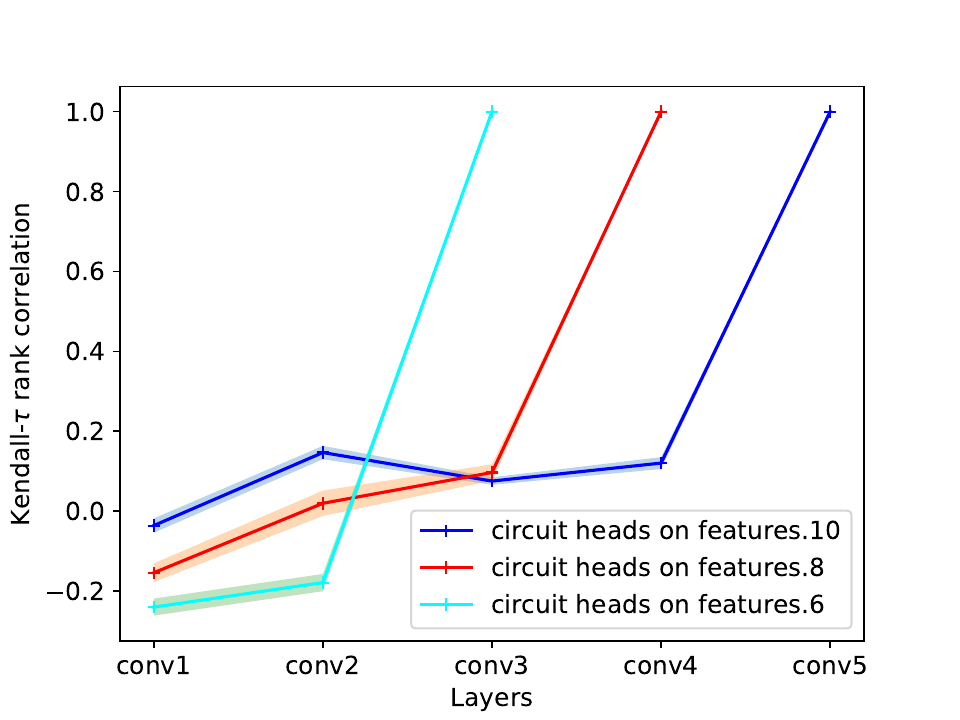}
\caption{Rank correlation between kernel attribution scores for circuits on (i) the initial model and (ii) the fine-tuned model with CircuitBreaker.}
\label{fig:circuit_attack_alexnet_rank_correlation}
\end{subfigure}
\begin{subfigure}[]{0.31\linewidth}

\includegraphics[width=\textwidth]{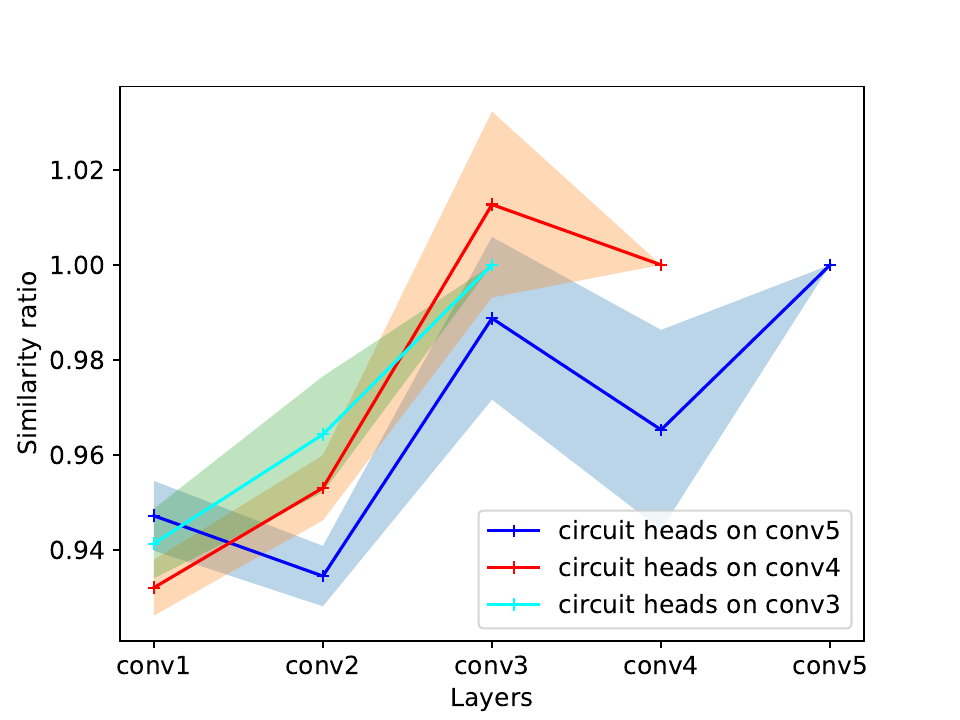}
\caption{Similarity ratio with CircuitBreaker.}
\label{fig:similarity_ratio}
\end{subfigure}

\caption{
Results obtained when fooling circuits with heads on conv3, conv4 and conv5 of AlexNet. \vspace{-15pt}
}\label{fig:condensed_results}
\end{figure*}

\subsection{Manipulation of the Circuit through CircuitBreaker}\label{sec:manipulation_channelcoercion}

In this section, we manipulate the model with CircuitBreaker as introduced in Section~\ref{sec:channelCoercion}. As a refresher, the goal of the CircuitBreaker attack mechanism is to fine-tune the pre-trained model to maintain its initial performance, fooling the interpretations of visual circuits (initial rankings of top channels and their synthetic feature visualization), while also preserving the functionality of the circuit head. In the following, we present the results obtained with CircuitBreaker on vision circuits for conv3 (features:6), conv4 (features:8), and conv5 (features:10) of AlexNet. Note that circuits (10 heads on each layer) are attacked independently from each other (ablation for simultaneous manipulation can be found in App.~\ref{app:simultaneous_manipulation}) and the visualized circuits use a sparsity of 0.6, which preserves well the behavior of the circuit heads (see Fig.~\ref{fig:circuit_attack_alexnet_pearson_correlation}). An ablation study on the visualizations with various sparsity levels in App.~\ref{app:ablation_sparsity}, and on the model (ResNet-50) is also provided in App.~\ref{app:channelCoercionResNet50}.

\textbf{Visual Inspection.}
We start by visually inspecting the results after CircuitBreaker. Fig.~\ref{fig:effectiveness_of_circuit_features_10}, Fig.~\ref{fig:effectiveness_of_circuit_features_8} and Fig.~\ref{fig:effectiveness_of_circuit_features_6} (appendix) show the results obtained after fooling attempts using CircuitBreaker on three different circuits (three different experiments). On the three different circuits (Sec.~\ref{sec:background} presents how visual circuits are discovered), when we compare the final one against the initial one, we observe that the final synthetic feature visualization still stays visually similar to the initial one, although it is less pronounced on the circuit for features.10:2 but in this case, it still shares the circular contour. This is due to the ProxPulse component in CircuitBreaker (see Section~\ref{sec:channelCoercion}). We also observe from the three circuits, a little overlap between channels numbering in the initial one and the final one, which is the effect of the ranking loss in CircuitBreaker. We finally observe that in terms of the semantics of the composition of synthetic feature visualization on the circuits, both initial and final circuits seem plausible. In particular, let us zoom onto the less obvious one in Fig.~\ref{fig:effectiveness_of_circuit_features_10}. An analysis of the initial circuit may roughly indicate that this circuit detects patterns related to circular objects with (vertical) axis (see e.g., Fig.~\ref{fig:ProxPulseConv4} and annotations of this unit from \citet{hernandez2022natural}). On previous layers of the features:10, we can see the presence of synthetic feature visualizations that visually seem to be dedicated to these circular contours (e.g., conv3:225, conv3:322) and others that are related to the (vertical) axis (conv3:112). As said above, the similarity between synthetic feature visualization of the final (i.e., after CircuitBreaker) circuit is less pronounced but it can still be observed the circular contour pattern. Indeed, this circular pattern has been amplified when looking at final synthetic feature visualizations. 
  
\textbf{Quantitative Assessment.}
The above analysis was a visual inspection of the manipulability of visual circuits under the CircuitBreaker attack. Here we quantify its success using four criteria (CT). \\
\textbf{CT1: Functional behavior.} First, to measure the preservation of the functional behavior of the circuit head, inspired by \cite{hamblin2022pruning}, we measure the Pearson correlation (on a large subset of the training set) between (i) activations of training images on the circuit head and (ii) activations of the same training images still on the circuit head but with sparsity 1. Higher correlations will mean high preservation of the functional behavior of the circuit head. Fig.\ref{fig:circuit_attack_alexnet_pearson_correlation} reports these correlation scores for several sparsity levels and different layers where circuit heads come from. It can be observed from dotted lines (fooled circuits) that the functional behavior of fooled circuits is preserved in the same way as the unfooled ones (bold lines), especially for moderate to high levels of sparsity ($>$.5).\\
\textbf{CT2: Sanity check of accuracy.} Second, as done in Section~\ref{sec:ProxPulse}, we assess the performance maintenance, ensuring that the fine-tuned model represents an adversarial model manipulation of the initial model.
We measure the performance of all finetuned models and report it in Fig.~\ref{fig:condensed_accuracy_alexnet} (appendix). The figure illustrates that our fine-tuning with CircuitBreaker maintains the same level of predictive performance of AlexNet accuracy on ImageNet, which is $56.52\%$.\\
\textbf{CT3: Correlation attribution scores.} Third, as done in Sec.~\ref{sec:ProxPulse_minor_effect}, we measure the
rank correlation between kernel attributions scores from (i) the initial model and (ii) the final model, which is the fine-tuned model with CircuitBreaker. As a result, a lower rank correlation will indicate a small change in the circuit, because these ranks are those that are used for circuit discovery. Fig.~\ref{fig:circuit_attack_alexnet_rank_correlation} shows these rank correlations. It can be seen from this figure that the final ranks of kernel attribution scores are weakly correlated to initial ones, except those of the circuit head's layers, which is reasonable. 
\\
\textbf{CT4: Similarity ratio between synthetic feature visualizations on the circuit.}
Finally, since with fine-tuning, channels can switch their feature visualizations (thus decreasing the rank correlation but not changing the interpretations of the circuit), we need a method to measure the change in synthetic feature visualization. Inspired by the phenomenon called \textit{whack-a-mole} in \citet{nanfack2023adversarial}, we use a similarity ratio computed thanks to CLIP~\citep{oikarinen2022clip} features. This similarity ratio is computed as follows. Given a final synthetic feature visualization from a layer, the numerator of the ratio is the maximum cosine similarity between this final synthetic image on the final model and any of the initial ones from the same layer on the initial model. The denominator is the cosine similarity between this final synthetic image on the final model and the synthetic image from the same channel but on the initial model. Intuitively, the ratio quantifies the change in synthetic visualization (initial vs final) relative to the initial synthetic visualization (using the final top channels). Fig.~\ref{fig:similarity_ratio} shows this similarity ratio per layer on different circuit heads. We observe that most values are lower than one, suggesting that synthetic feature visualization has changed. It is also important to observe that the similarity ratio (see the ending point of each curve) of the circuit head collapses to one, which means that in general, there is negligible change in synthetic feature visualizations of the circuit head.

\section{Conclusion and Limitations}\label{sec:conclusion}

This paper proposes a manipulation technique called ProxPulse that extends the limitations of previous works, by showing that both types of feature visualizations can be simultaneously manipulated. However, when analyzing ProxPulse within the framework of circuits --key components in mechanistic interpretability--, we discover that circuits show some robustness against ProxPulse manipulations. We therefore introduce another attack that reveals the manipulability of circuits. We provide experimental evidence of the effectiveness of these attacks using a variety of correlation and similarity metrics. Our attack on circuits sheds light on the lack of uniqueness and stability of circuit-based interpretations. We also observe a decrease in manipulability success when trying to attack simultaneously several circuits without degradation in accuracy. Finally, further studies need to be done to provide defense mechanisms and robust-circuit discovery methods. 



\bibliographystyle{plainnat}
\bibliography{example_paper}

\begin{thebibliography}{37}
\providecommand{\natexlab}[1]{#1}
\providecommand{\url}[1]{\texttt{#1}}
\expandafter\ifx\csname urlstyle\endcsname\relax
  \providecommand{\doi}[1]{doi: #1}\else
  \providecommand{\doi}{doi: \begingroup \urlstyle{rm}\Url}\fi

\bibitem[Adebayo et~al.(2018)Adebayo, Gilmer, Muelly, Goodfellow, Hardt, and
  Kim]{adebayo2018sanity}
Julius Adebayo, Justin Gilmer, Michael Muelly, Ian Goodfellow, Moritz Hardt,
  and Been Kim.
\newblock Sanity checks for saliency maps.
\newblock \emph{Advances in neural information processing systems}, 31, 2018.

\bibitem[A{\"\i}vodji et~al.(2021)A{\"\i}vodji, Arai, Gambs, and
  Hara]{aivodji2021characterizing}
Ulrich A{\"\i}vodji, Hiromi Arai, S{\'e}bastien Gambs, and Satoshi Hara.
\newblock Characterizing the risk of fairwashing.
\newblock \emph{Advances in Neural Information Processing Systems},
  34:\penalty0 14822--14834, 2021.

\bibitem[Anders et~al.(2020)Anders, Pasliev, Dombrowski, M{\"u}ller, and
  Kessel]{anders2020fairwashing}
Christopher Anders, Plamen Pasliev, Ann-Kathrin Dombrowski, Klaus-Robert
  M{\"u}ller, and Pan Kessel.
\newblock Fairwashing explanations with off-manifold detergent.
\newblock In \emph{International Conference on Machine Learning}, pages
  314--323. PMLR, 2020.

\bibitem[Bareeva et~al.(2024)Bareeva, H{\"o}hne, Warnecke, Pirch, M{\"u}ller,
  Rieck, and Bykov]{bareeva2024manipulating}
Dilyara Bareeva, Marina M-C H{\"o}hne, Alexander Warnecke, Lukas Pirch,
  Klaus-Robert M{\"u}ller, Konrad Rieck, and Kirill Bykov.
\newblock Manipulating feature visualizations with gradient slingshots.
\newblock \emph{arXiv preprint arXiv:2401.06122}, 2024.

\bibitem[Bau et~al.(2020)Bau, Zhu, Strobelt, Lapedriza, Zhou, and
  Torralba]{bau2020understanding}
David Bau, Jun-Yan Zhu, Hendrik Strobelt, Agata Lapedriza, Bolei Zhou, and
  Antonio Torralba.
\newblock Understanding the role of individual units in a deep neural network.
\newblock \emph{Proceedings of the National Academy of Sciences}, 117\penalty0
  (48):\penalty0 30071--30078, 2020.

\bibitem[Chughtai et~al.(2023)Chughtai, Chan, and Nanda]{chughtai2023toy}
Bilal Chughtai, Lawrence Chan, and Neel Nanda.
\newblock A toy model of universality: Reverse engineering how networks learn
  group operations.
\newblock In \emph{International Conference on Machine Learning}, pages
  6243--6267. PMLR, 2023.

\bibitem[Conmy et~al.(2023)Conmy, Mavor-Parker, Lynch, Heimersheim, and
  Garriga-Alonso]{conmy2023towards}
Arthur Conmy, Augustine~N Mavor-Parker, Aengus Lynch, Stefan Heimersheim, and
  Adri{\`a} Garriga-Alonso.
\newblock Towards automated circuit discovery for mechanistic interpretability.
\newblock \emph{arXiv preprint arXiv:2304.14997}, 2023.

\bibitem[Dai et~al.(2022)Dai, Dong, Hao, Sui, Chang, and Wei]{dai2022knowledge}
Damai Dai, Li~Dong, Yaru Hao, Zhifang Sui, Baobao Chang, and Furu Wei.
\newblock Knowledge neurons in pretrained transformers.
\newblock In \emph{Proceedings of the 60th Annual Meeting of the Association
  for Computational Linguistics (Volume 1: Long Papers)}, pages 8493--8502,
  2022.

\bibitem[Deng et~al.(2009)Deng, Dong, Socher, Li, Li, and Fei-Fei]{Imagenet}
Jia Deng, Wei Dong, Richard Socher, Li-Jia Li, Kai Li, and Li~Fei-Fei.
\newblock Imagenet: A large-scale hierarchical image database.
\newblock In \emph{2009 IEEE Conference on Computer Vision and Pattern
  Recognition}, pages 248--255, 2009.

\bibitem[Dombrowski et~al.(2019)Dombrowski, Alber, Anders, Ackermann,
  M{\"u}ller, and Kessel]{dombrowski2019explanations}
Ann-Kathrin Dombrowski, Maximillian Alber, Christopher Anders, Marcel
  Ackermann, Klaus-Robert M{\"u}ller, and Pan Kessel.
\newblock Explanations can be manipulated and geometry is to blame.
\newblock \emph{Advances in neural information processing systems}, 32, 2019.

\bibitem[Foret et~al.(2020)Foret, Kleiner, Mobahi, and
  Neyshabur]{foret2020sharpness}
Pierre Foret, Ariel Kleiner, Hossein Mobahi, and Behnam Neyshabur.
\newblock Sharpness-aware minimization for efficiently improving
  generalization.
\newblock In \emph{International Conference on Learning Representations}, 2020.

\bibitem[Geirhos et~al.(2023)Geirhos, Zimmermann, Bilodeau, Brendel, and
  Kim]{geirhos2023don}
Robert Geirhos, Roland~S Zimmermann, Blair Bilodeau, Wieland Brendel, and Been
  Kim.
\newblock Don't trust your eyes: on the (un) reliability of feature
  visualizations.
\newblock \emph{arXiv preprint arXiv:2306.04719}, 2023.

\bibitem[Hamblin et~al.(2022)Hamblin, Konkle, and Alvarez]{hamblin2022pruning}
Chris Hamblin, Talia Konkle, and George Alvarez.
\newblock Pruning for interpretable, feature-preserving circuits in cnns.
\newblock \emph{arXiv preprint arXiv:2206.01627}, 2022.

\bibitem[He et~al.(2016)He, Zhang, Ren, and Sun]{he2016deep}
Kaiming He, Xiangyu Zhang, Shaoqing Ren, and Jian Sun.
\newblock Deep residual learning for image recognition.
\newblock In \emph{Proceedings of the IEEE conference on computer vision and
  pattern recognition}, pages 770--778, 2016.

\bibitem[Heo et~al.(2019)Heo, Joo, and Moon]{heo2019fooling}
Juyeon Heo, Sunghwan Joo, and Taesup Moon.
\newblock Fooling neural network interpretations via adversarial model
  manipulation.
\newblock \emph{Advances in Neural Information Processing Systems}, 32, 2019.

\bibitem[Hernandez et~al.(2022)Hernandez, Schwettmann, Bau, Bagashvili,
  Torralba, and Andreas]{hernandez2022natural}
Evan Hernandez, Sarah Schwettmann, David Bau, Teona Bagashvili, Antonio
  Torralba, and Jacob Andreas.
\newblock Natural language descriptions of deep visual features.
\newblock In \emph{International Conference on Learning Representations}, 2022.

\bibitem[Hinton et~al.(2015)Hinton, Vinyals, and Dean]{hinton2015distilling}
Geoffrey Hinton, Oriol Vinyals, and Jeff Dean.
\newblock Distilling the knowledge in a neural network.
\newblock \emph{stat}, 1050:\penalty0 9, 2015.

\bibitem[Huang et~al.(2017)Huang, Liu, Van Der~Maaten, and
  Weinberger]{huang2017densely}
Gao Huang, Zhuang Liu, Laurens Van Der~Maaten, and Kilian~Q Weinberger.
\newblock Densely connected convolutional networks.
\newblock In \emph{Proceedings of the IEEE conference on computer vision and
  pattern recognition}, pages 4700--4708, 2017.

\bibitem[Hubel and Wiesel(1962)]{hubel1962receptive}
David~H Hubel and Torsten~N Wiesel.
\newblock Receptive fields, binocular interaction and functional architecture
  in the cat's visual cortex.
\newblock \emph{The Journal of physiology}, 160\penalty0 (1):\penalty0 106,
  1962.

\bibitem[Krizhevsky et~al.(2012)Krizhevsky, Sutskever, and
  Hinton]{krizhevsky2012imagenet}
Alex Krizhevsky, Ilya Sutskever, and Geoffrey~E Hinton.
\newblock Imagenet classification with deep convolutional neural networks.
\newblock \emph{Advances in Neural Information Processing Systems}, 25, 2012.

\bibitem[Lee et~al.(2018)Lee, Ajanthan, and Torr]{lee2018snip}
Namhoon Lee, Thalaiyasingam Ajanthan, and Philip Torr.
\newblock Snip: Single-shot network pruning based on connection sensitivity.
\newblock In \emph{International Conference on Learning Representations}, 2018.

\bibitem[Mahendran and Vedaldi(2015)]{mahendran2015understanding}
Aravindh Mahendran and Andrea Vedaldi.
\newblock Understanding deep image representations by inverting them.
\newblock In \emph{Proceedings of the IEEE conference on computer vision and
  pattern recognition}, pages 5188--5196, 2015.

\bibitem[Meng et~al.(2022)Meng, Bau, Andonian, and Belinkov]{meng2022locating}
Kevin Meng, David Bau, Alex Andonian, and Yonatan Belinkov.
\newblock Locating and editing factual associations in gpt.
\newblock \emph{Advances in Neural Information Processing Systems},
  35:\penalty0 17359--17372, 2022.

\bibitem[Nanfack et~al.(2024)Nanfack, Fulleringer, Marty, Eickenberg, and
  Belilovsky]{nanfack2023adversarial}
Geraldin Nanfack, Alexander Fulleringer, Jonathan Marty, Michael Eickenberg,
  and Eugene Belilovsky.
\newblock Adversarial attacks on the interpretation of neuron activation
  maximization.
\newblock In \emph{Proceedings of the AAAI Conference on Artificial
  Intelligence}, volume~38, pages 4315--4324, 2024.

\bibitem[Oikarinen and Weng(2022)]{oikarinen2022clip}
Tuomas Oikarinen and Tsui-Wei Weng.
\newblock Clip-dissect: Automatic description of neuron representations in deep
  vision networks.
\newblock \emph{arXiv preprint arXiv:2204.10965}, 2022.

\bibitem[Olah et~al.(2017)Olah, Mordvintsev, and Schubert]{olah2017feature}
Chris Olah, Alexander Mordvintsev, and Ludwig Schubert.
\newblock Feature visualization.
\newblock \emph{Distill}, 2017.
\newblock https://distill.pub/2017/feature-visualization.

\bibitem[Olah et~al.(2020)Olah, Cammarata, Schubert, Goh, Petrov, and
  Carter]{olah2020zoom}
Chris Olah, Nick Cammarata, Ludwig Schubert, Gabriel Goh, Michael Petrov, and
  Shan Carter.
\newblock Zoom in: An introduction to circuits.
\newblock \emph{Distill}, 5\penalty0 (3):\penalty0 e00024--001, 2020.

\bibitem[Paszke et~al.(2019)Paszke, Gross, Massa, Lerer, Bradbury, Chanan,
  Killeen, Lin, Gimelshein, Antiga, et~al.]{paszke2019pytorch}
Adam Paszke, Sam Gross, Francisco Massa, Adam Lerer, James Bradbury, Gregory
  Chanan, Trevor Killeen, Zeming Lin, Natalia Gimelshein, Luca Antiga, et~al.
\newblock Pytorch: An imperative style, high-performance deep learning library.
\newblock \emph{Advances in neural information processing systems}, 32, 2019.

\bibitem[Rony et~al.(2019)Rony, Hafemann, Oliveira, Ayed, Sabourin, and
  Granger]{rony2019decoupling}
J{\'e}r{\^o}me Rony, Luiz~G Hafemann, Luiz~S Oliveira, Ismail~Ben Ayed, Robert
  Sabourin, and Eric Granger.
\newblock Decoupling direction and norm for efficient gradient-based l2
  adversarial attacks and defenses.
\newblock In \emph{Proceedings of the IEEE/CVF Conference on Computer Vision
  and Pattern Recognition}, pages 4322--4330, 2019.

\bibitem[Rudner and Toner(2021)]{rudner2021key}
Tim~GJ Rudner and Helen Toner.
\newblock Key concepts in ai safety: an overview.
\newblock \emph{Comput. Secur. J. doi}, 10:\penalty0 20190040, 2021.

\bibitem[Wang et~al.(2022)Wang, Variengien, Conmy, Shlegeris, and
  Steinhardt]{wang2022interpretability}
Kevin~Ro Wang, Alexandre Variengien, Arthur Conmy, Buck Shlegeris, and Jacob
  Steinhardt.
\newblock Interpretability in the wild: a circuit for indirect object
  identification in gpt-2 small.
\newblock In \emph{The Eleventh International Conference on Learning
  Representations}, 2022.

\bibitem[W{\"a}schle et~al.(2022)W{\"a}schle, Thaler, Berres, P{\"o}lzlbauer,
  and Albers]{waschle2022review}
Moritz W{\"a}schle, Florian Thaler, Axel Berres, Florian P{\"o}lzlbauer, and
  Albert Albers.
\newblock A review on ai safety in highly automated driving.
\newblock \emph{Frontiers in Artificial Intelligence}, 5:\penalty0 952773,
  2022.

\bibitem[Yosinski et~al.(2015)Yosinski, Clune, Nguyen, Fuchs, and
  Lipson]{yosinski2015understanding}
Jason Yosinski, Jeff Clune, Anh Nguyen, Thomas Fuchs, and Hod Lipson.
\newblock Understanding neural networks through deep visualization.
\newblock \emph{arXiv preprint arXiv:1506.06579}, 2015.

\bibitem[Yu(2013)]{yu2013stability}
Bin Yu.
\newblock Stability.
\newblock \emph{Bernoulli}, pages 1484--1500, 2013.

\bibitem[Zeiler and Fergus(2014)]{zeiler2014visualizing}
Matthew~D Zeiler and Rob Fergus.
\newblock Visualizing and understanding convolutional networks.
\newblock In \emph{European conference on computer vision}, pages 818--833.
  Springer, 2014.

\bibitem[Zimmermann et~al.(2021)Zimmermann, Borowski, Geirhos, Bethge, Wallis,
  and Brendel]{zimmermann2021well}
Roland~S Zimmermann, Judy Borowski, Robert Geirhos, Matthias Bethge, Thomas
  Wallis, and Wieland Brendel.
\newblock How well do feature visualizations support causal understanding of
  cnn activations?
\newblock \emph{Advances in Neural Information Processing Systems},
  34:\penalty0 11730--11744, 2021.

\bibitem[Zimmermann et~al.(2023)Zimmermann, Klein, and
  Brendel]{zimmermann2023scale}
Roland~S Zimmermann, Thomas Klein, and Wieland Brendel.
\newblock Scale alone does not improve mechanistic interpretability in vision
  models.
\newblock In \emph{Thirty-seventh Conference on Neural Information Processing
  Systems}, 2023.

\end{thebibliography}

\newpage
\appendix

\section{Appendix / supplemental material}

\subsection{Broader Impact}\label{app:broader_impact}
Our work aims to study the lack of stability and robustness of popular interpretability techniques. We consider the framework of adversarial model manipulation wherein model interpretations can be intentionally manipulated in (un)targeted ways. Demonstrating this manipulability, unfortunately, highlights the risk of individuals exploiting this knowledge to deploy models whose interpretations are obfuscated. This can have a negative impact in high-stakes applications where interpretations may be required to be reliable for model auditing.
However, we believe that acknowledging and understanding these risks is a crucial first step in addressing vulnerabilities of interpretability techniques. 

\subsection{Further Experimental Details}\label{app:experimental_details}
We were inspired by the experimental setups of \citet{nanfack2023adversarial} and \citet{hamblin2022pruning}, to choose models, and hyperparameters for visual circuit discovery. The choice of the model and most experimental settings were made according to \citet{nanfack2023adversarial}, while the circuit discovery and its hyperparameters were taken from \cite{hamblin2022pruning}, using their source code. The hyperparameters of our method, specifically the values of
$\rho$ and C were inspired by the adversarial robustness literature (with l2 norm). In particular, we set $\rho=0.02 \approx 5/255$ inspired from the adversarial literature \citep{rony2019decoupling}, $C=1e6$ which is $\approx 1e3$ times higher than empirically observed activations of initial synthetic images~\footnote{In Eq.~\ref{eq:worst_activations}, C enables the control of the magnitude of activations in the manipulated synthetic images.}, and set the hyperparameters $\alpha=0.1$ and $\beta = 0.01$ such that the fooling loss and the maintain loss have similar scales. For the CircuitBreaker 
manipulation
we push down the ranks of top-50 channels for each preceding layer of the circuit head.

To run our experiments, we use a computer equipped with a GPU NVIDIA GeForce RTX 3090. Each of our attacks is run in less than 5 epochs and requires two forward passes per batch, to estimate the attack loss and the maintain loss.

\subsection{Derivation of the Loss Function}\label{app:proximity_pulse}
This section derives the expression of the ProxPulse loss.
Drawing inspiration from \citet{foret2020sharpness}, we derive the expression of Eq.~\ref{eq:approximation_eq} by first writing, 
\begin{equation}\label{eq:approximation_eq_latex}
      \mathcal{L}_\textsuperscript{F}(\mathcal{D}_\textsuperscript{fool}; \vtheta) = \sum_{j,\vx^*\in \mathcal{D}_\text{fool}} \max_{||\vx - \vx^*||\le \rho} \ell_j(\vx;\vtheta).
\end{equation}
Second, given that, 
\begin{equation}
    \begin{split}
        \argmax_{||\vx - \vx^*||\le \rho} \ell_j(\vx;\vtheta) &  =  \argmax_{||\vepsilon||\le \rho} \ell_j(\vx^* + \vepsilon ;\vtheta) \\
  & \approx \argmax_{||\vepsilon||\le \rho} \ell_j(\vx^*;\vtheta) + \vepsilon^T\nabla_\vx \ell_j(\vx^*;\vtheta) \hspace{3cm} \text{(using first-order Taylor expansion)}\\
  & = \argmax_{||\vepsilon||\le \rho} \vepsilon^T\nabla_\vx \ell_j(\vx^*;\vtheta) \\
  & = \rho \frac{\nabla_\vx \ell_j(\vx^*;\vtheta)}{||\nabla_\vx \ell_j(\vx^*;\vtheta)||}.
    \end{split}
\end{equation}
Finally, plugging this approximation into Eq.~\ref{eq:approximation_eq_latex} recovers Eq.~\ref{eq:approximation_eq}.

\begin{figure}[!t]
\centering
\begin{subfigure}[]{0.3\linewidth}
\includegraphics[width=\textwidth]{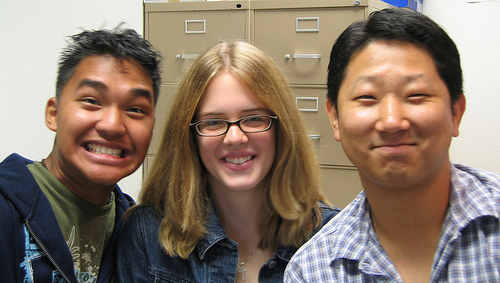}
\end{subfigure}\hspace{.2cm}
\begin{subfigure}[]{.24\linewidth}
\includegraphics[width=\textwidth]{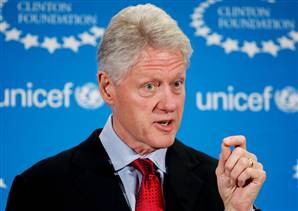}
\end{subfigure}\hspace{.2cm}
\caption{
Target images ($\mathcal{D}_{\text{fool}}$) for ProxPulse, taken from the ImageNet-21k dataset. 
}
\label{fig:target_images}
\end{figure}


\begin{figure}[!t]
\centering
\includegraphics[width=\linewidth]{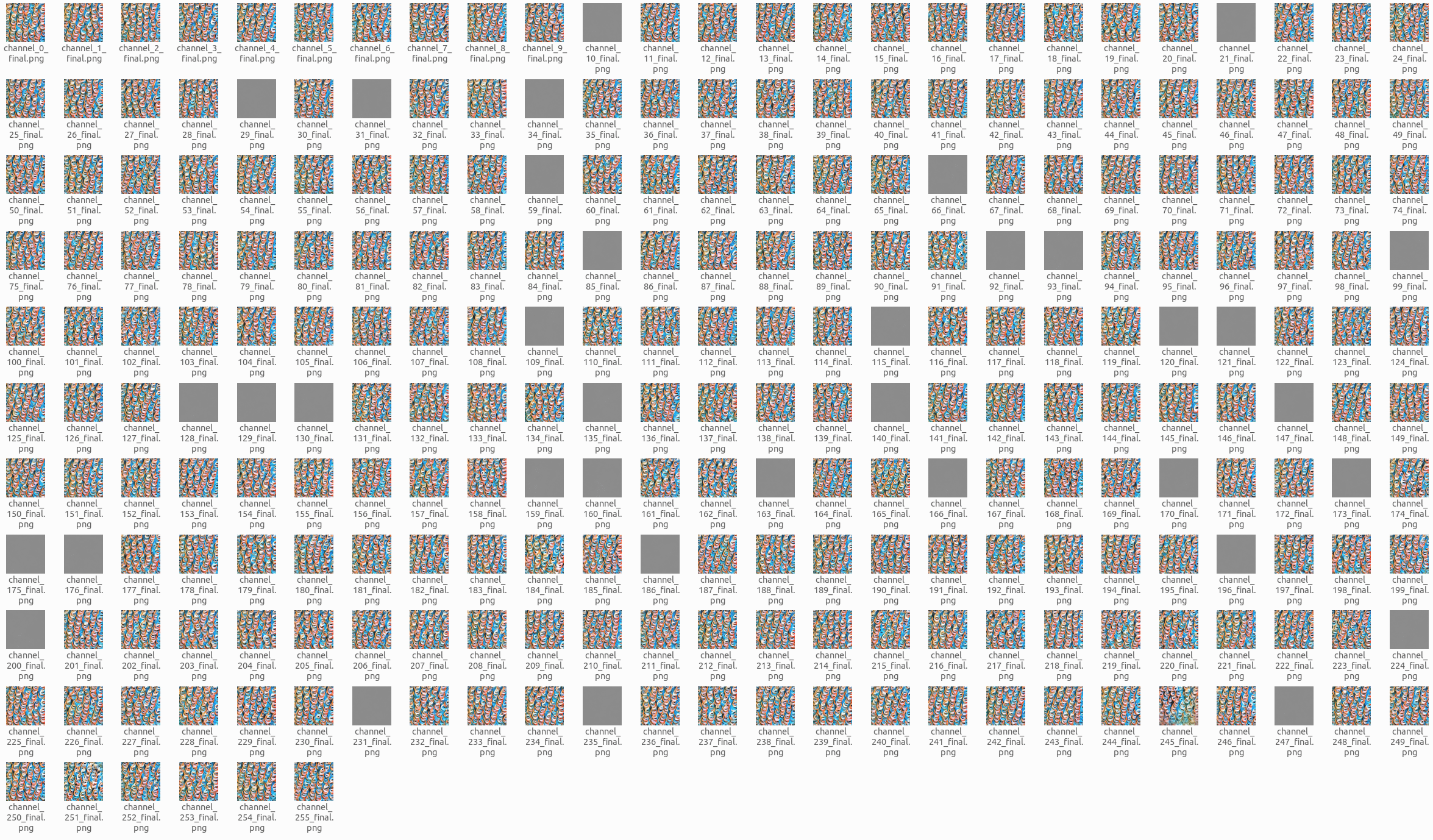}
\caption{
Synthetic images after ProxPulse on conv5 of AlexNet. 
}\label{fig:synthetic_images_conv5}
\end{figure}

\begin{figure}[!ht]
\centering
\includegraphics[width=\linewidth]{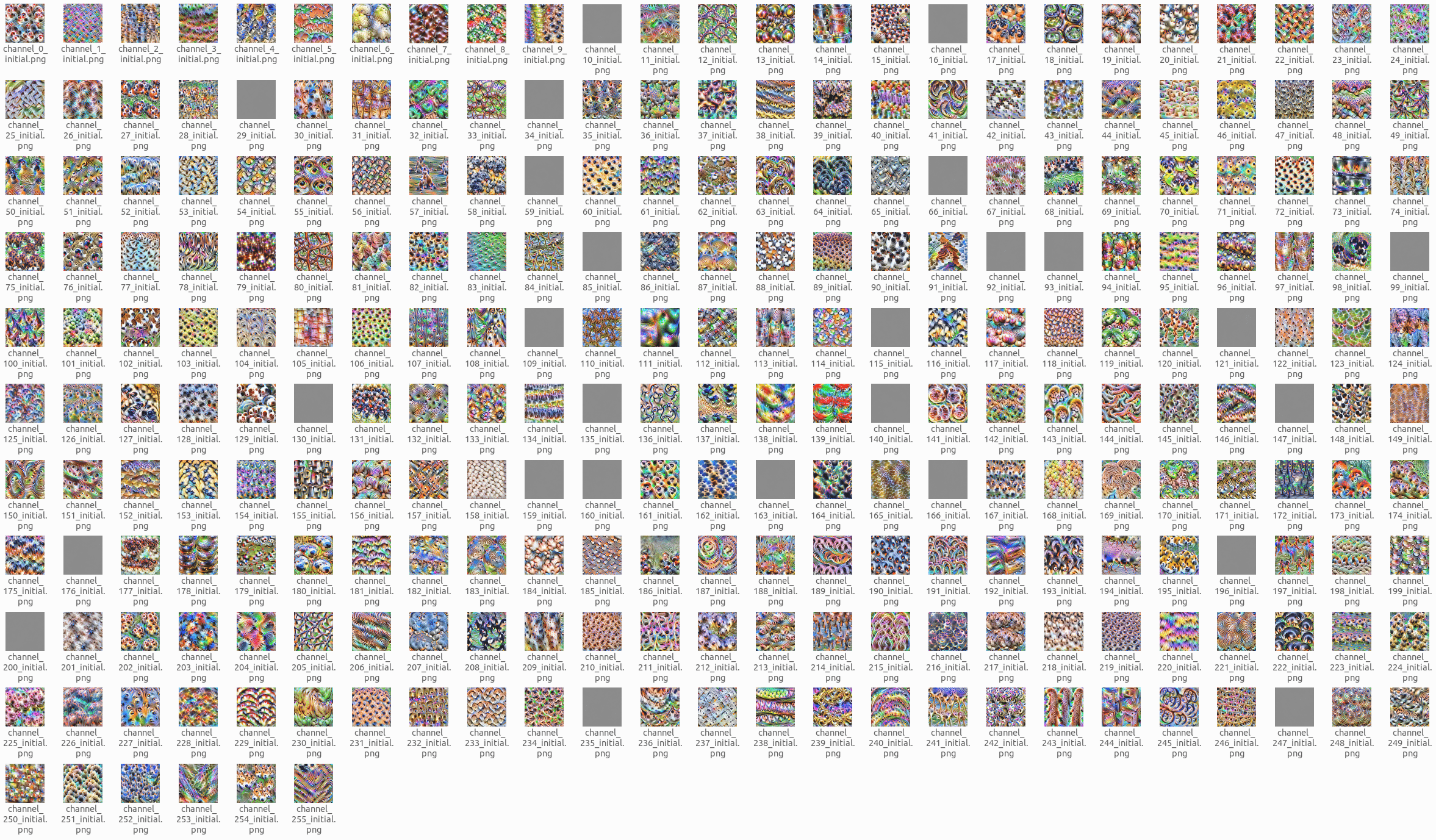}
\caption{
Initial synthetic images of conv5 of AlexNet. 
}\label{fig:synthetic_images_conv5_initial}
\end{figure}

\begin{figure}[!t]
\centering
\includegraphics[width=\linewidth]{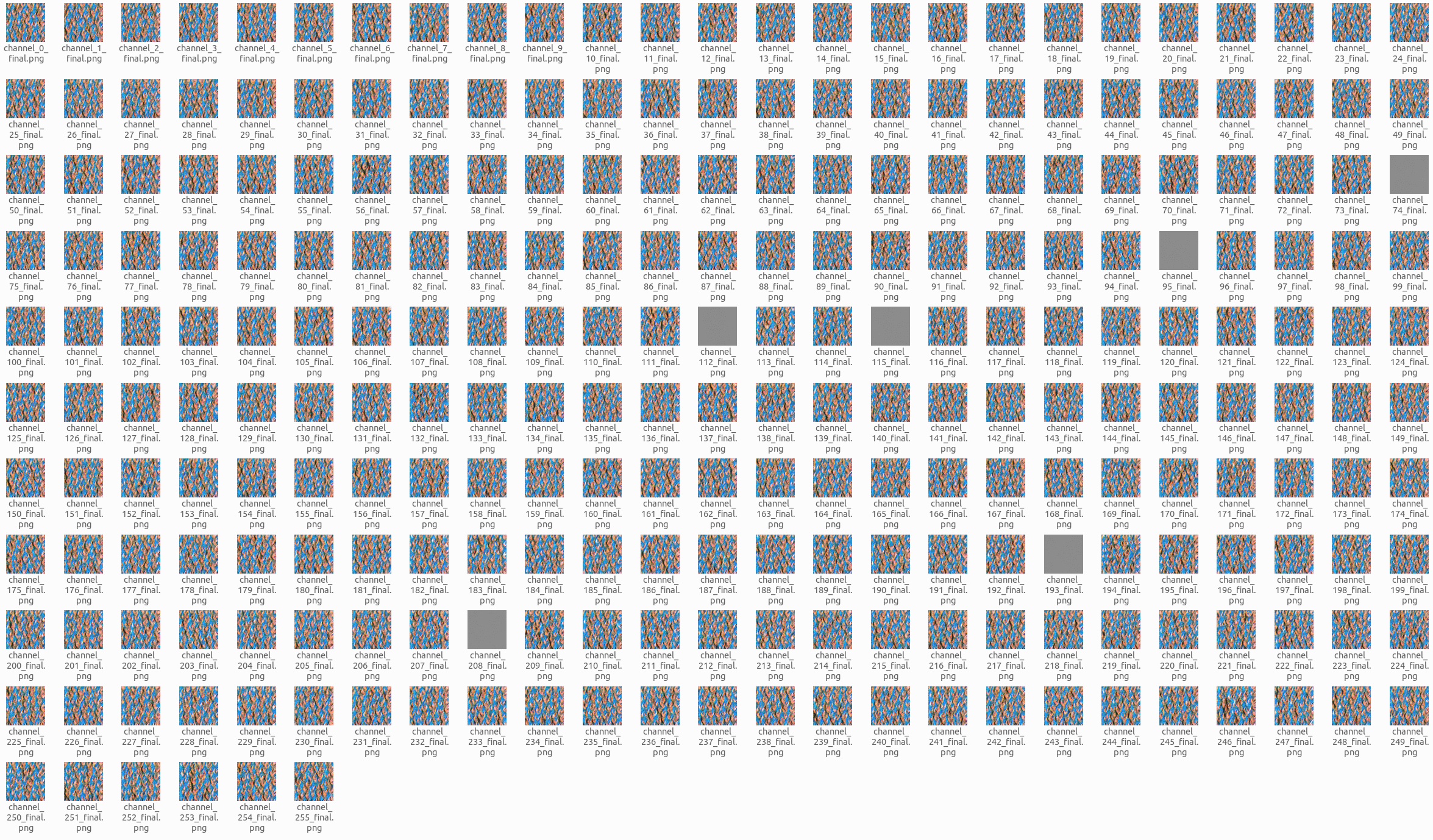}
\caption{
Synthetic images after ProxPulse on conv4 of AlexNet. 
}\label{fig:synthetic_images_conv4}
\end{figure}

\begin{figure}[!h]
\centering
\includegraphics[width=\linewidth]{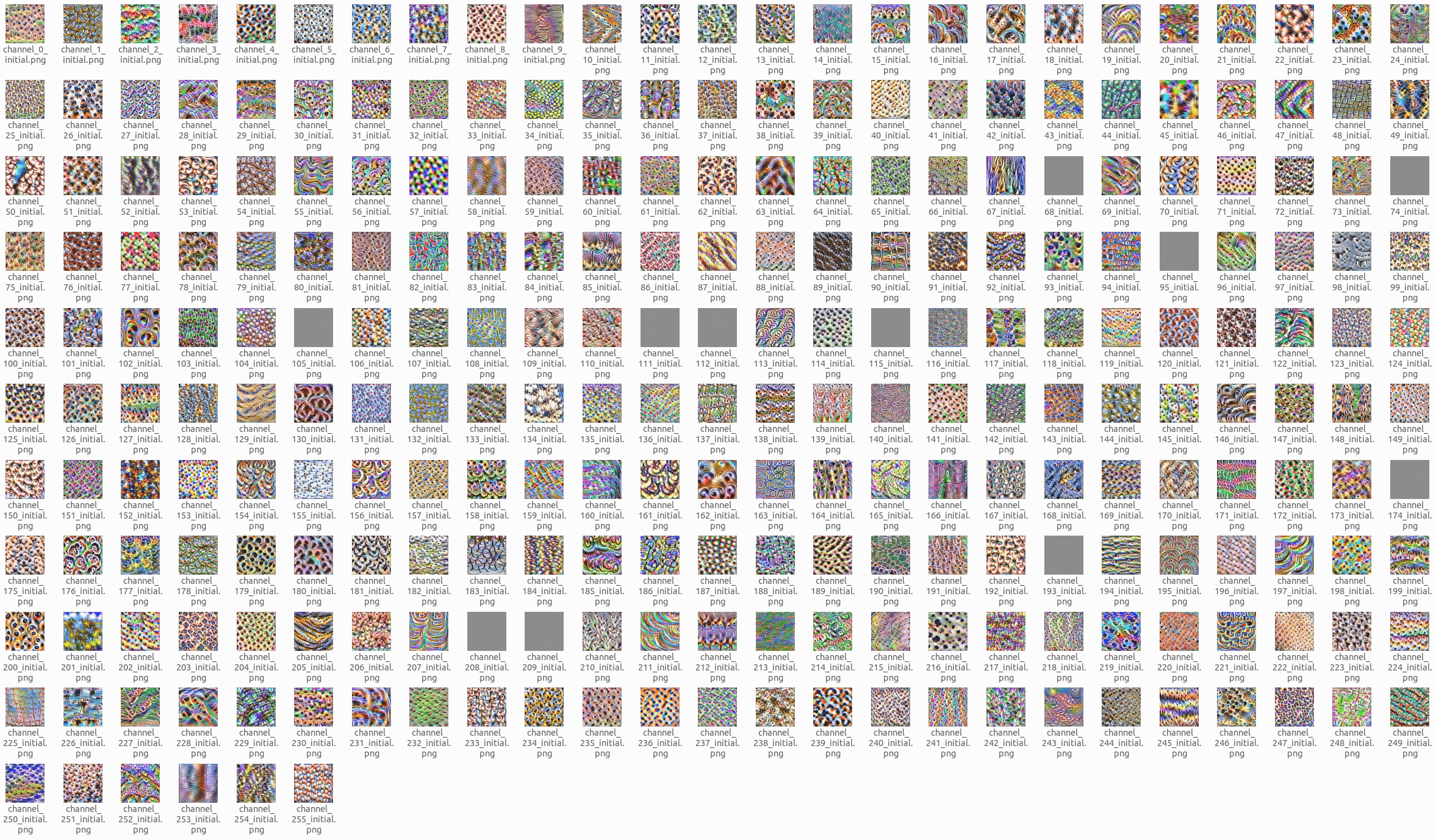}
\caption{
Initial synthetic images of conv4 of AlexNet. 
}\label{fig:synthetic_images_conv4_initial}
\end{figure}

\subsection{Visual Inspection of All Synthetic Feature Visualizations of a Layer}\label{app:all_synthetic}
Fig.~\ref{fig:synthetic_images_conv5_initial} and Fig.~\ref{fig:synthetic_images_conv5} respectively show all synthetic feature visualizations generated on layer conv5 the initial model (i.e., before ProxPulse) and on the final model (i.e., after ProxPulse). We do the same for Fig.~\ref{fig:synthetic_images_conv4_initial} and Fig.~\ref{fig:synthetic_images_conv4} on layer conv4. It can be quickly observed that except for the \textit{noisy} ones, which are sometimes those from the random initialization), all the synthetic images have been replaced with visually similar ones. Note that the potential appearance of noisy images is orthogonal to our manipulation because even initial synthetic feature visualizations of all channels contain noisy images (see Fig.~\ref{fig:synthetic_images_conv5_initial} and Fig.~\ref{fig:synthetic_images_conv4_initial}).
\clearpage
\subsection{Results for ProxPulse on ResNet-50}\label{app:ProxPulseResNet50}
\begin{figure*}[!t]
\centering
\begin{subfigure}[]{0.3\linewidth}
\includegraphics[width=\textwidth]{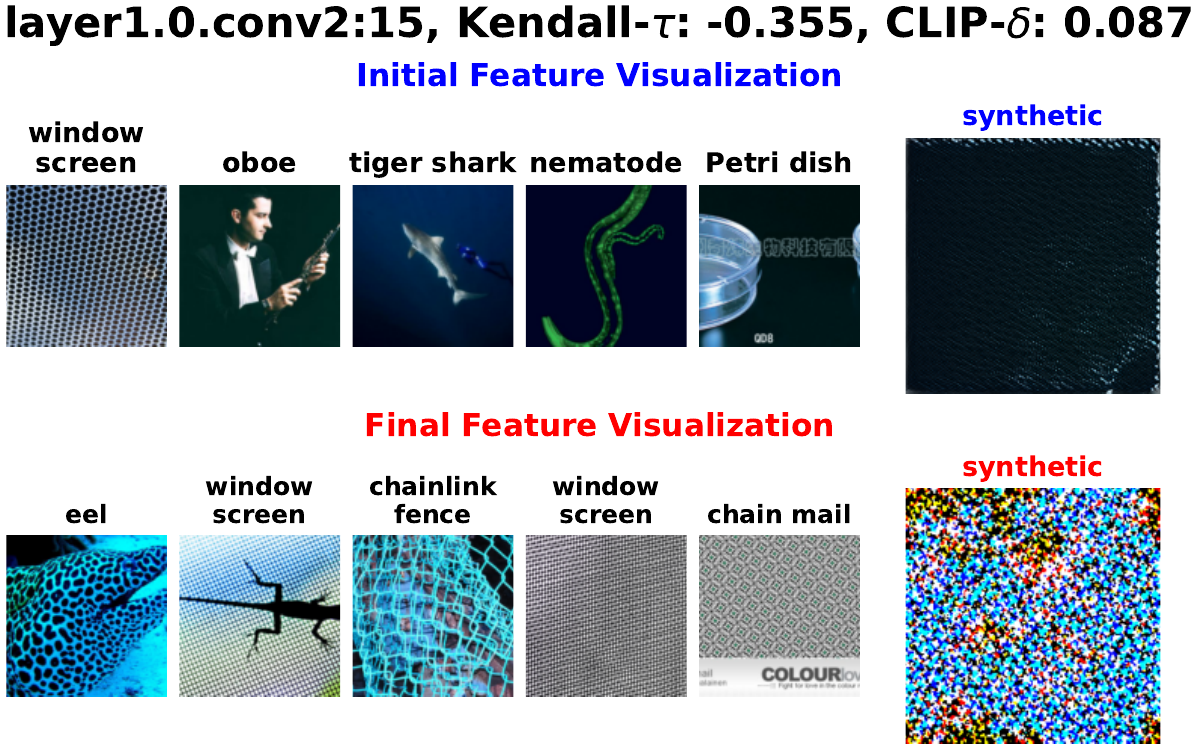}
\end{subfigure}\hspace{.2cm}
\begin{subfigure}[]{0.3\linewidth}
\includegraphics[width=\textwidth]{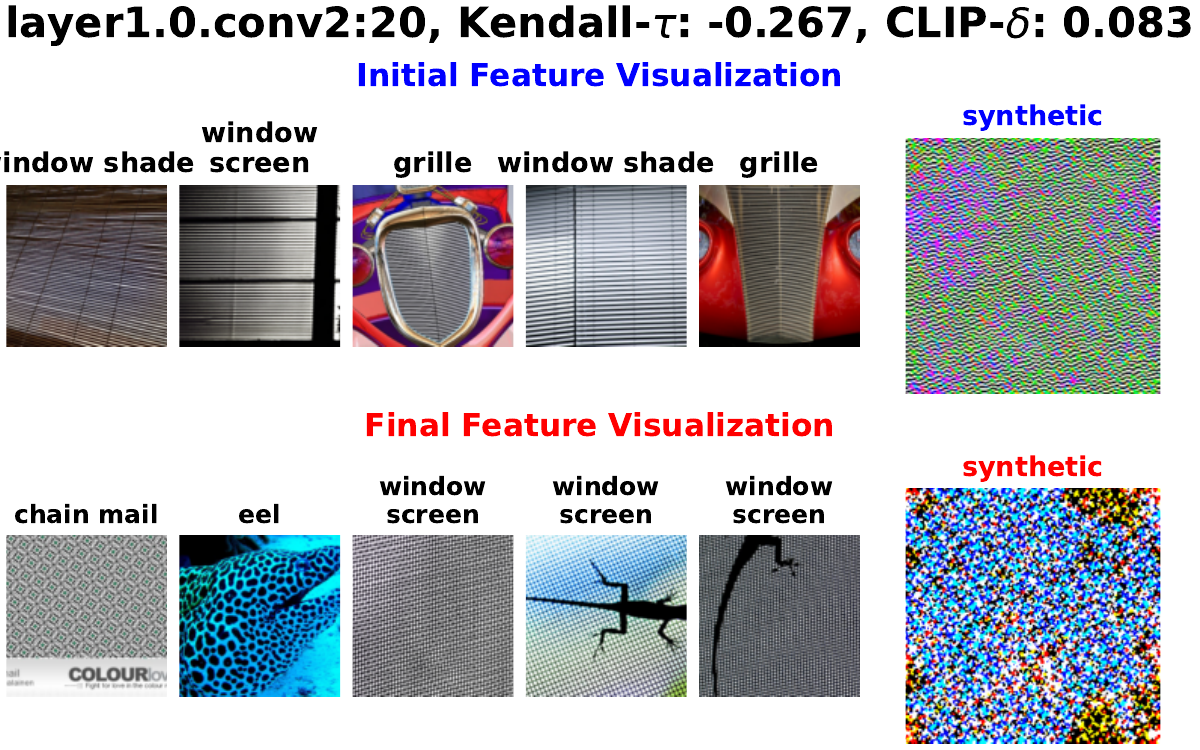}
\end{subfigure}\hspace{.2cm}
\begin{subfigure}[]{0.3\linewidth}
\includegraphics[width=\textwidth]{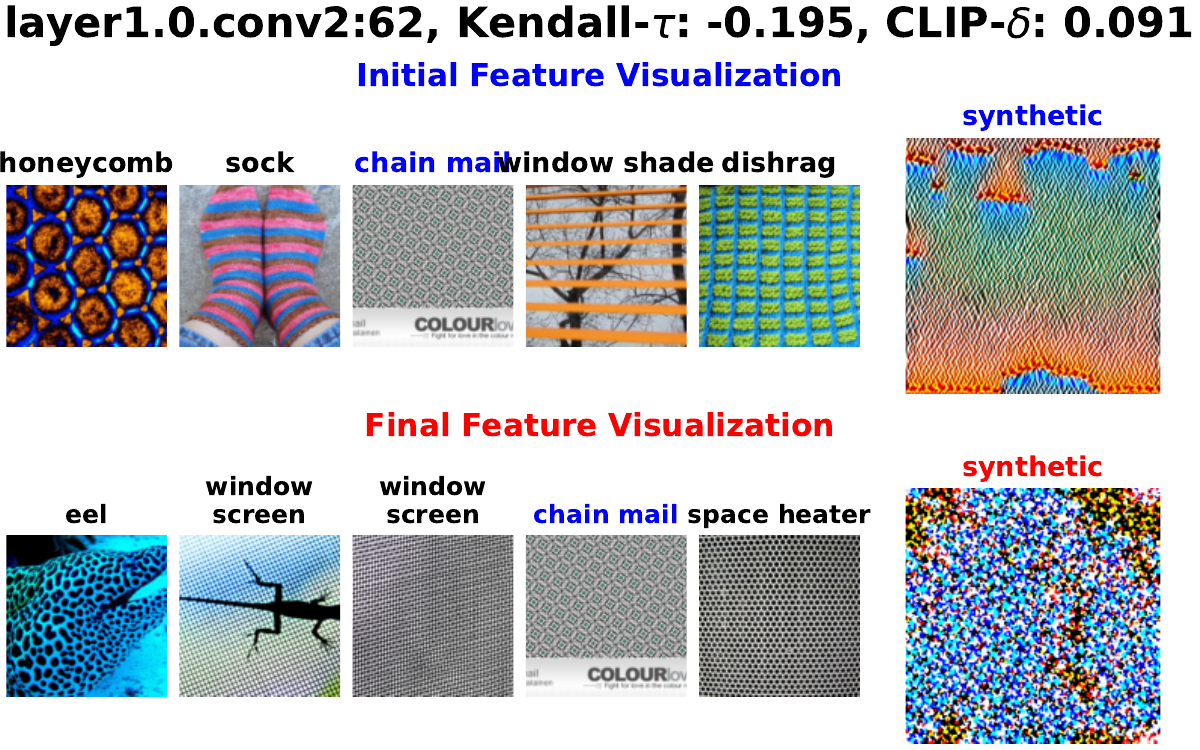}
\end{subfigure}
\caption{
 Illustration of the manipulability of both natural and synthetic feature visualization on Layer1.0.conv2 of ResNet-50. 
}\label{fig:resnet50_ablation_featvis}
\end{figure*}

\begin{figure}[!t]
\centering
\begin{subfigure}[]{0.5\linewidth}
\includegraphics[width=\textwidth]{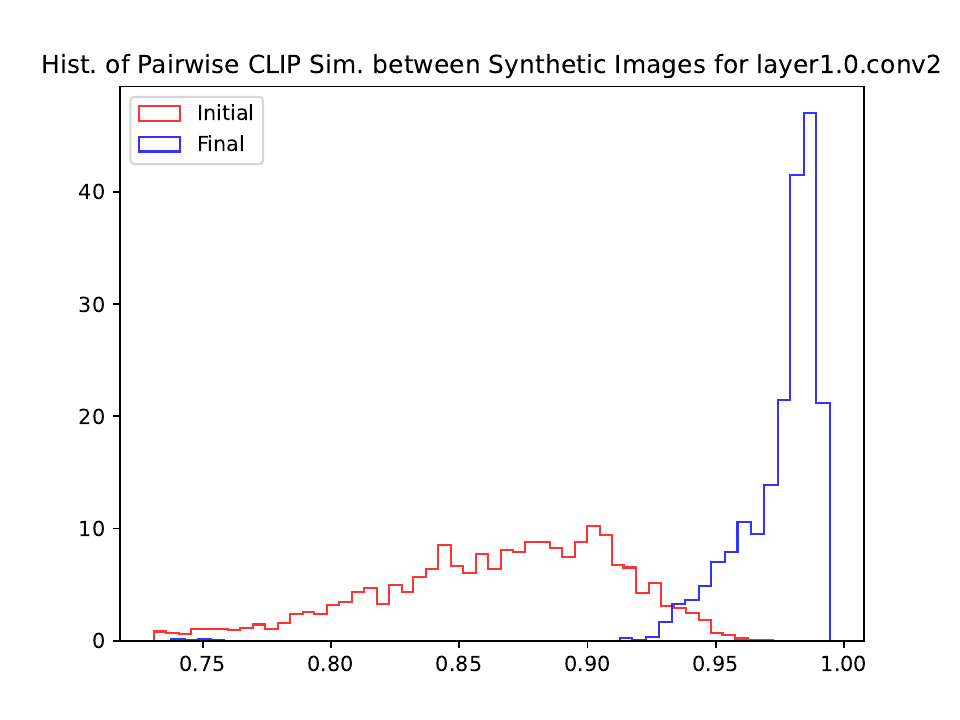}
\end{subfigure}
\caption{Histogram of pairwise cosine similarities between CLIP features of non-noisy synthetic feature visualization before (red) and after (blue) the ProxPulse manipulation. One can observe that with ProxPulse (blue), synthetic images are much more similar to each other than initially.}
\label{fig:img_similaritieslayer1.0.conv2}
\end{figure}

We ablate the model for experiments done in Section~\ref{sec:ProxPulseEffectiveness} to demonstrate that our ProxPulse attack also works on different types of models. More specifically, we do the similar experiment on layer1.0.conv2 and report the result in Fig.~\ref{fig:resnet50_ablation_featvis} and Fig.~\ref{fig:img_similaritieslayer1.0.conv2}. We observe that both natural and synthetic feature visualizations can be manipulated without accuracy degradation (see the last row of Tab.~\ref{tab:main_overall} to confirm that the fine-tuned model with ProxPulse has the same level of accuracy as ResNet-50 initial performance, which is 80.3\%).

\clearpage
\subsection{Ablation for the Use of a Single Target in ProxPulse Manipulation}\label{app:abblation_on_target}
\begin{figure}[!h]
\centering
\includegraphics[width=\linewidth]{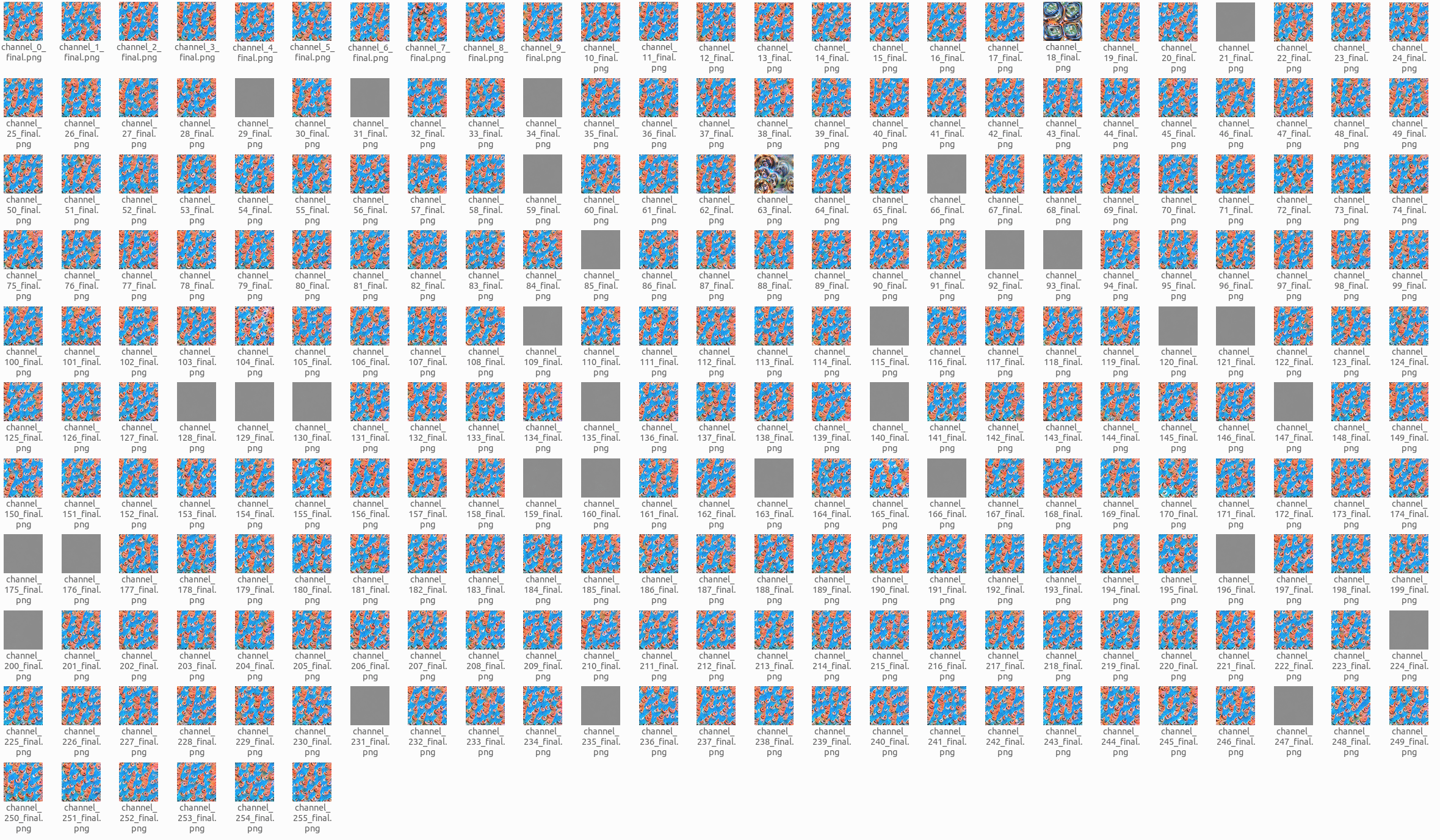}
\caption{
Final synthetic images with one target image.}\label{fig:synthetic_images_on_targetl}
\end{figure}

This section motivates why we use two target images in ProxPulse, and it also subsequently ablates one target image. Fig.~\ref{fig:synthetic_images_on_targetl} shows that some of the final synthetic images have not been substantially changed, motivating therefore the use of two target images. 

\clearpage
\subsection{Further Experiments on the Non-Effectiveness of ProxPulse to Attack Visual Circuits}\label{app:ProxPulseFutherAlexNet}
\begin{figure*}[!t]
\begin{subfigure}[]{0.45\linewidth}
\includegraphics[width=\textwidth]{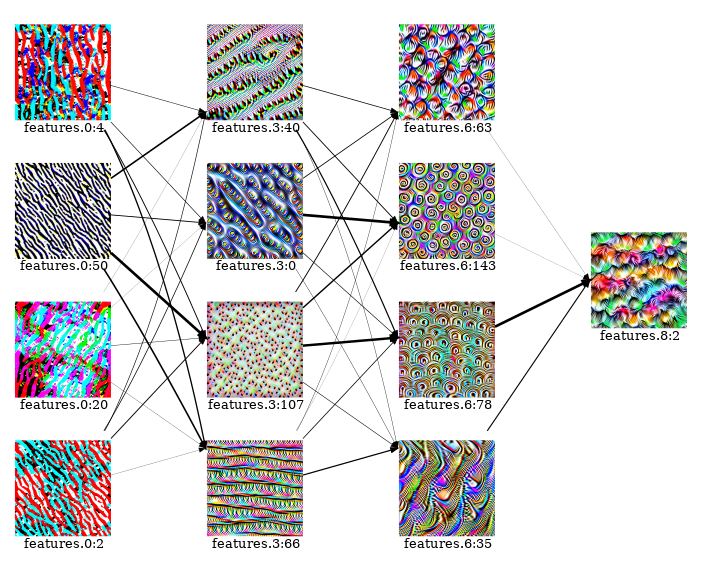}
\caption{Initial circuit with sparsity: 1.}
\end{subfigure}
\begin{subfigure}[]{0.45\linewidth}
\includegraphics[width=\textwidth]{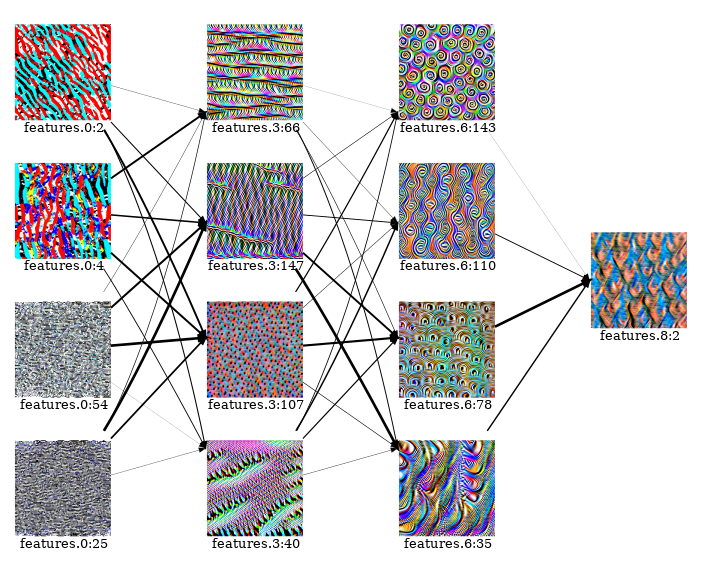}
\caption{Final circuit with sparsity: 1.}
\end{subfigure}

\begin{subfigure}[]{0.45\linewidth}
\includegraphics[width=\textwidth]{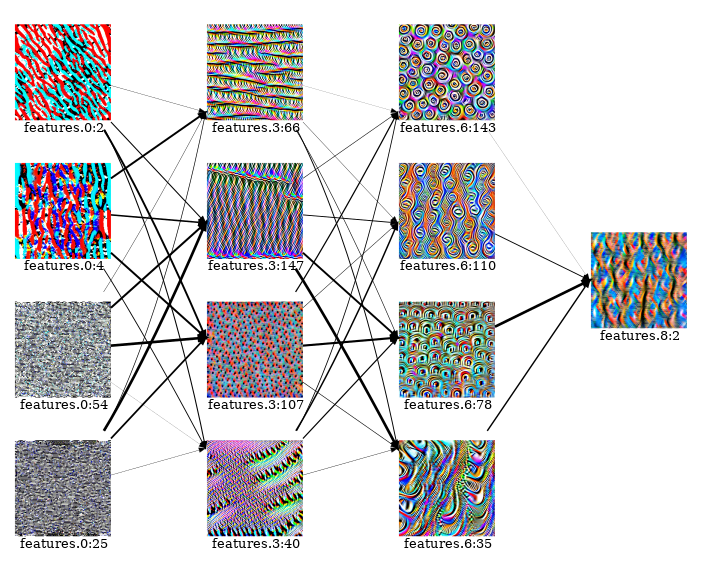}
\caption{Fianal circuit with sparsity: 0.3.}
\end{subfigure}
\begin{subfigure}[]{.45\linewidth}
\includegraphics[width=\textwidth]{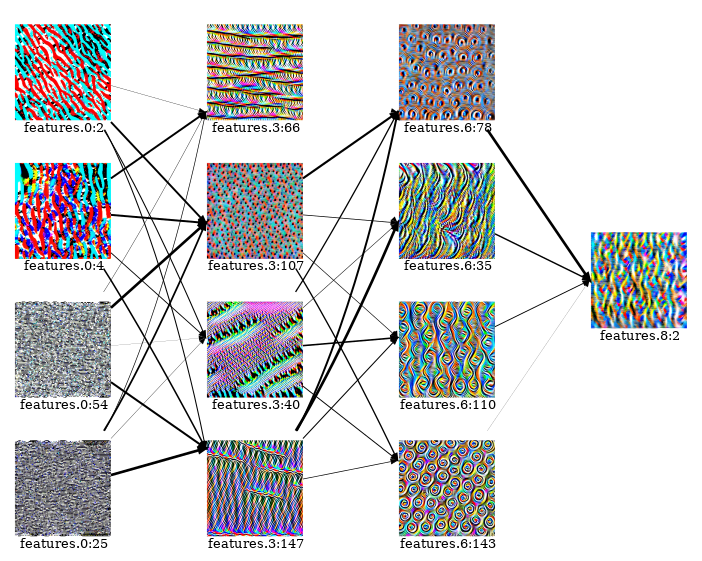}
\caption{Fianal circuit with sparsity: 0.1.}
\end{subfigure}

\caption{
 Illustration of the non-effectiveness of the ProxPulse fooling to manipulate the circuit.
 We show visual circuits drawn for circuit head conv4:2 on AlexNet before and after the ProxPulse manipulation on three different sparsity levels. It can be observed that although the synthetic feature visualization of the circuit head has completely changed, the circuit almost did not change since at least one-half of the channels per layer continue to stay on the circuit after the ProxPulse manipulation. Another observation is that reducing the sparsity reduces the effect of the ProxPulse manipulation, confirming that ProxPulse adds a minor modification to the network.
}\label{fig:ineffectiveness_of_circuit_conv4}
\end{figure*}

\begin{figure}[!t]
\centering
\includegraphics[width=.4\textwidth]{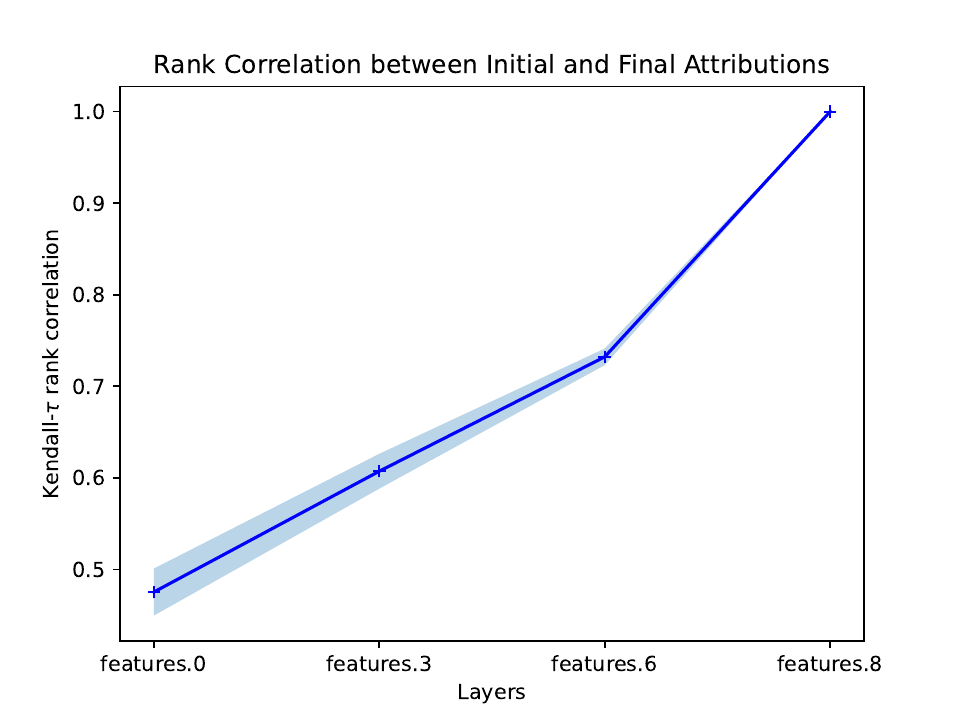} 
\caption{Correlation of attribution scores between the initial and the final (fine-tuned with ProxPulse) model. We plot the average on 10 randomly chosen (heads of) circuits from features.8 (conv4). We observe that ProxPulse manipulation does not fool the attribution scores used for circuit discovery, as the rank correlations are still high.}
\label{fig:ProxPulseCircuit_failure_rank_correlation_further}
\end{figure}

Fig.~\ref{fig:ProxPulseCircuit_failure_rank_correlation_further} and Fig.~\ref{fig:ineffectiveness_of_circuit_conv4} further illustrate the non-effectiveness of ProxPulse to attack visual circuits. It can be observed from Fig.~\ref{fig:ineffectiveness_of_circuit_conv4} that at least one-half of channel indexes continue to stay on the circuit after ProxPulse.
We also observe from Fig.~\ref{fig:ProxPulseCircuit_failure_rank_correlation_further} that the rank correlation scores between kernel attribution scores for circuit discovery are high, suggesting little impact of ProxPulse on the circuit discovery method.

\begin{figure}[!t]
  \vspace{-15pt}
    \centering
    \includegraphics[width=0.5\textwidth]{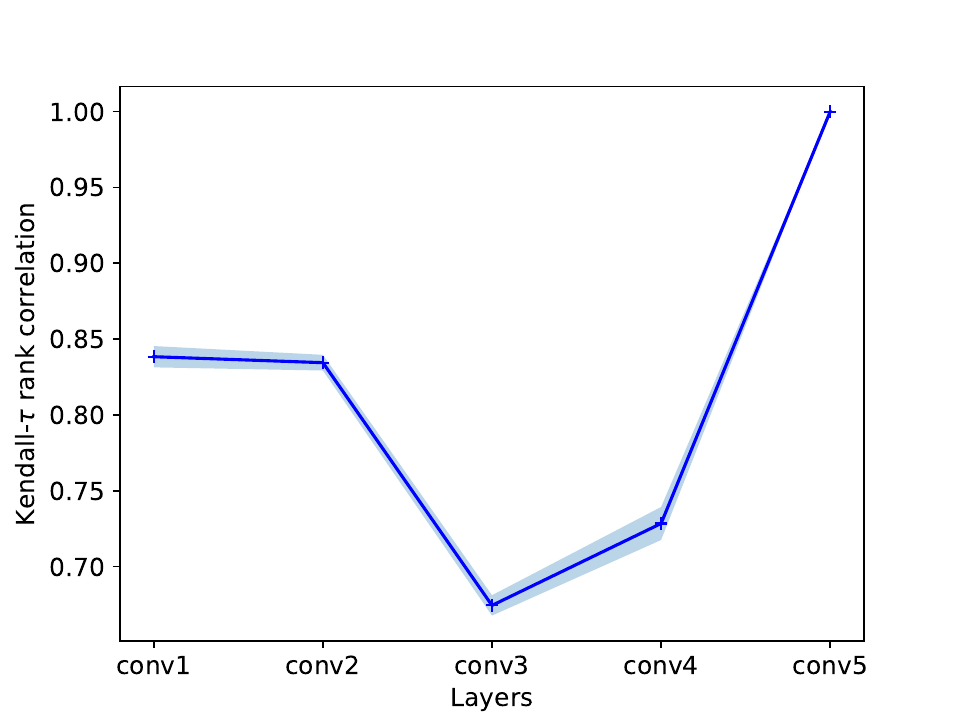}
    \captionof{figure}{Correlation of attribution scores between the initial and the final (fine-tuned with ProxPulse) model. We plot the average on 10 randomly chosen (heads of) circuits from conv5. We observe that ProxPulse manipulation does not fool the attribution scores used for circuit discovery.}
    \label{fig:ProxPulseCircuit_failure_rank_correlation}
\end{figure}

\clearpage
\subsection{Further Visualizations of CircuitBreaker on AlexNet}\label{app:channelcoercionAlexNet}

\begin{figure}[!t]
\centering
\begin{subfigure}[]{0.45\linewidth}
\includegraphics[width=\textwidth]{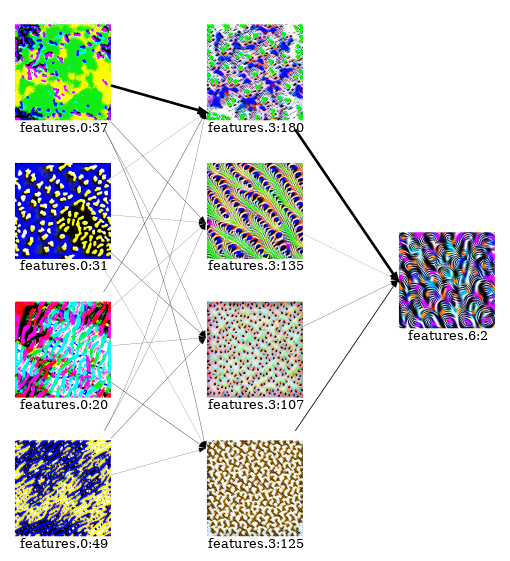}
\caption{With initial model.}
\end{subfigure}\hspace{.2cm}
\begin{subfigure}[]{0.45\linewidth}
\includegraphics[width=\textwidth]{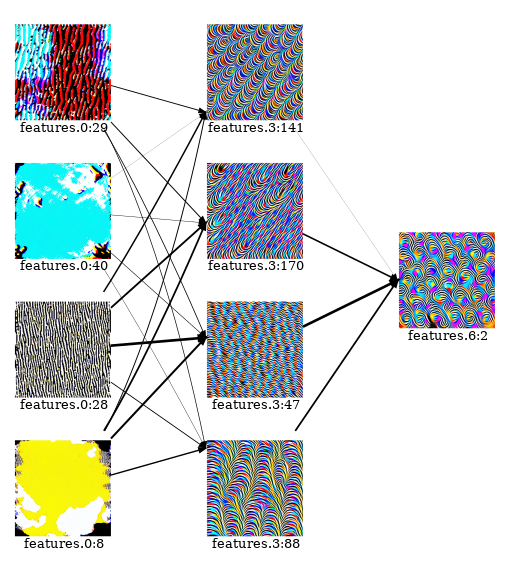}
\caption{After CircuitBreaker.}
\end{subfigure}
\caption{
 Illustration of the effectiveness of CircuitBreaker to manipulate visual circuits on features:8 (conv4) of AlexNet. 
}\label{fig:effectiveness_of_circuit_features_6}
\end{figure}

\begin{figure*}[!t]
\centering
\begin{subfigure}[]{0.45\linewidth}
\includegraphics[width=\textwidth]{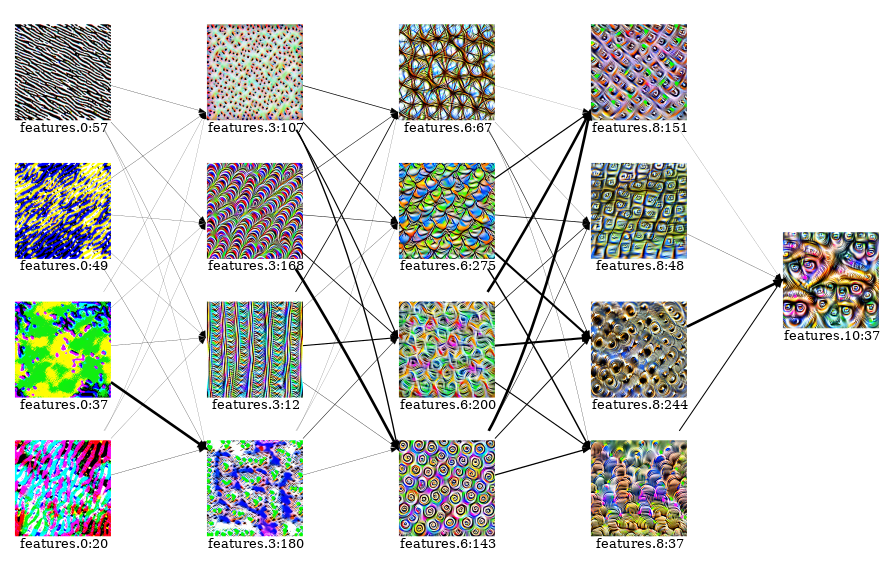}
\caption{Initial circuit.}
\end{subfigure}\hspace{.2cm}
\begin{subfigure}[]{0.45\linewidth}
\includegraphics[width=\textwidth]{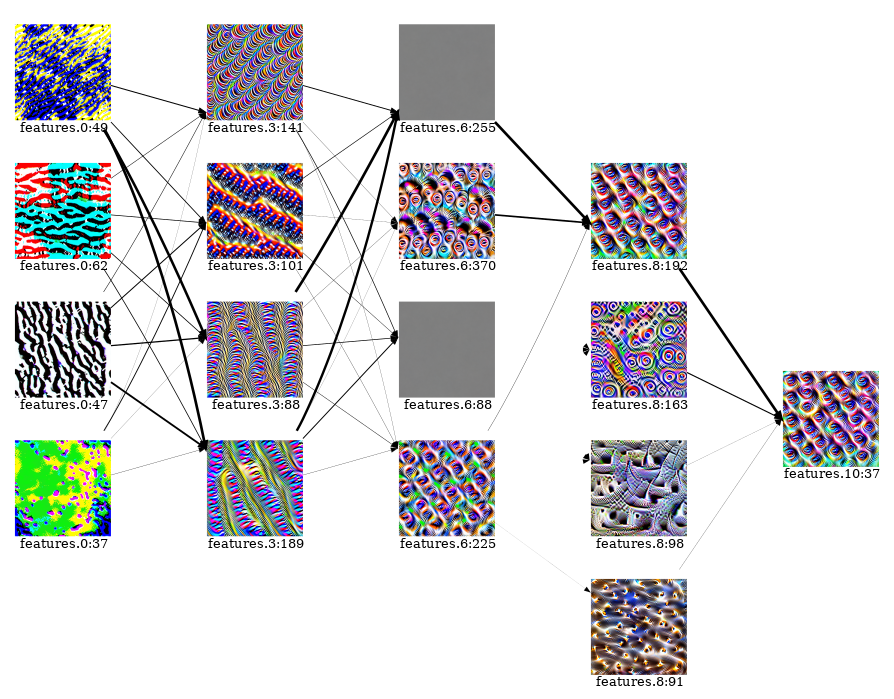}
\caption{Final circuit.}
\end{subfigure}
\caption{
 Illustration of the effectiveness of CircuitBreaker to manipulate the circuit on conv5 of AlexNet.
}\label{fig:effectiveness_of_circuit_features_10_37}
\end{figure*}

\begin{figure*}[!t]
\centering
\begin{subfigure}[]{0.45\linewidth}
\includegraphics[width=\textwidth]{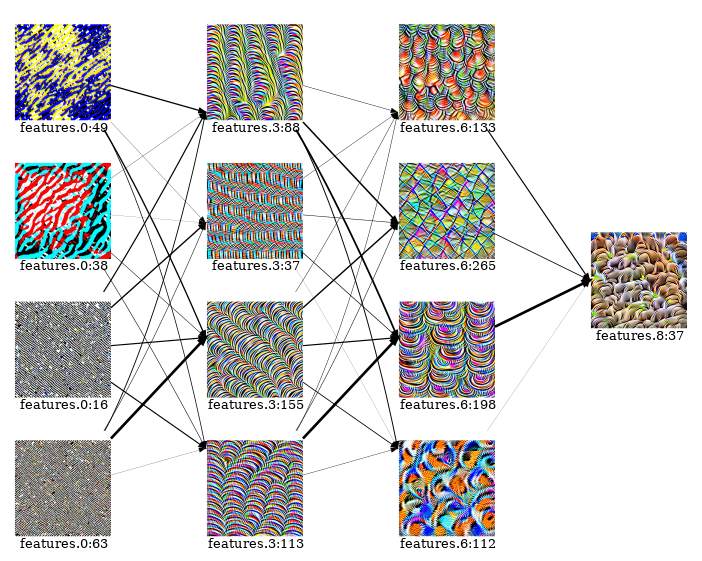}
\caption{Initial circuit.}
\end{subfigure}\hspace{.2cm}
\begin{subfigure}[]{0.45\linewidth}
\includegraphics[width=\textwidth]{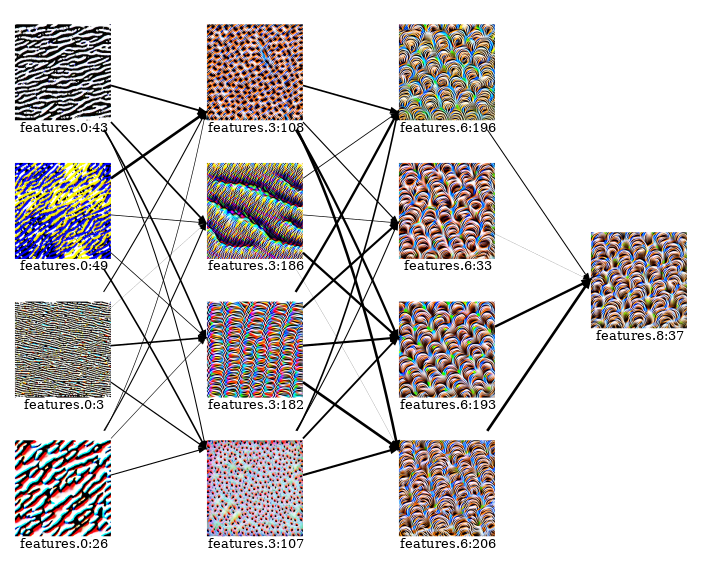}
\caption{Final circuit.}
\end{subfigure}
\caption{
 Illustration of the effectiveness of CircuitBreaker to manipulate the circuit.
}\label{fig:effectiveness_of_circuit_features_8_37}
\end{figure*}

\begin{figure*}[!t]
\centering
\begin{subfigure}[]{0.45\linewidth}
\includegraphics[width=\textwidth]{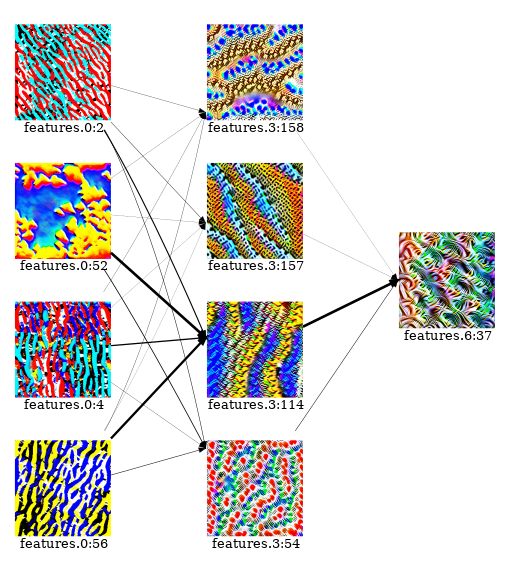}
\caption{Initial circuit.}
\end{subfigure}\hspace{.2cm}
\begin{subfigure}[]{0.45\linewidth}
\includegraphics[width=\textwidth]{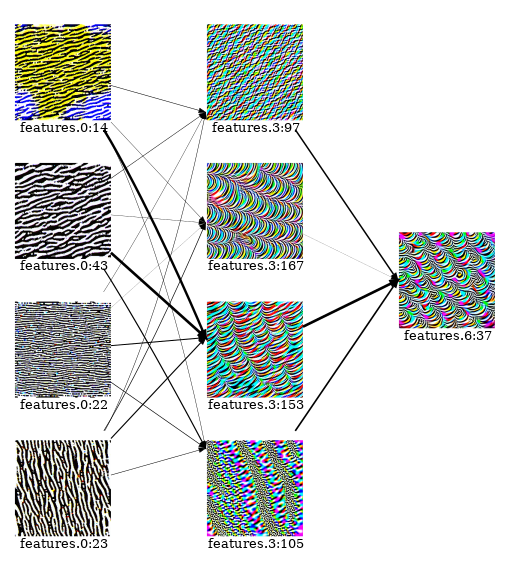}
\caption{Final circuit.}
\end{subfigure}
\caption{
 Illustration of the effectiveness of CircuitBreaker to manipulate the circuit.
}\label{fig:effectiveness_of_circuit_features_6_37}
\end{figure*}

Fig.~\ref{fig:effectiveness_of_circuit_features_10_37}, Fig.~\ref{fig:effectiveness_of_circuit_features_8_37} and Fig.~\ref{fig:effectiveness_of_circuit_features_6_37} demonstrate the visual inspection of the effectiveness of CircuitBreaker to fool initial circuit. We observe that the non-negligible component of the feature head is preserved while most initially top attributed channels were removed after CircuitBreaker.

\clearpage

\subsection{Ablation for Sparsity  for CircuitBreaker}\label{app:ablation_sparsity}
\begin{figure*}[!t]
\centering
\begin{subfigure}[]{0.45\linewidth}
\includegraphics[width=\textwidth]{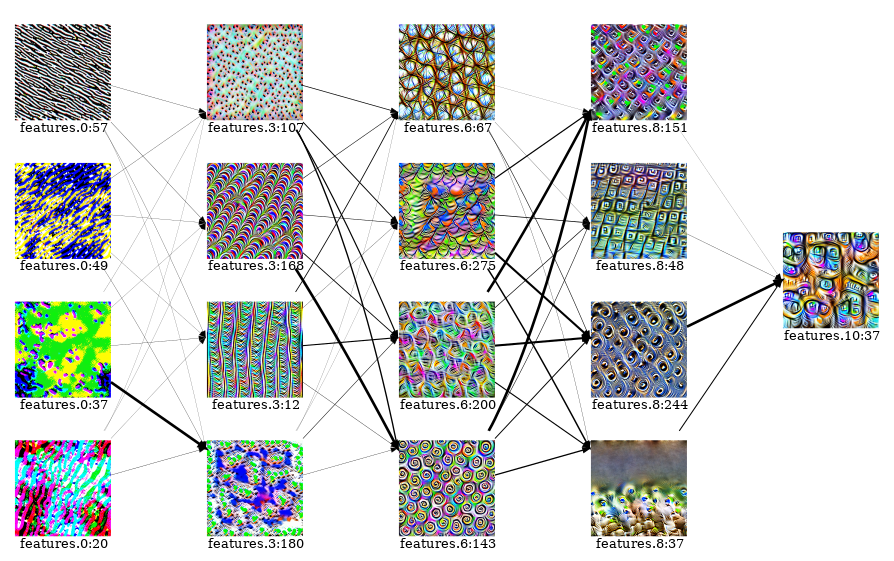}
\caption{Initial circuit.}
\end{subfigure}\hspace{.2cm}
\begin{subfigure}[]{0.45\linewidth}
\includegraphics[width=\textwidth]{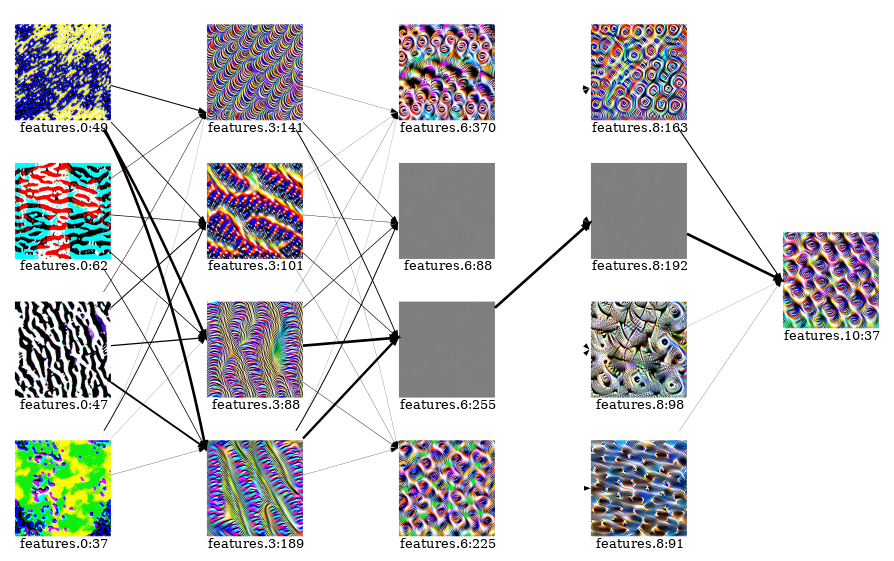}
\caption{Final circuit.}
\end{subfigure}
\caption{
 Illustration of the effectiveness of CircuitBreaker to manipulate the circuit: ablation on the sparsity level. 
}\label{fig:effectiveness_of_circuit_features_10_37_ablation}
\end{figure*}

\begin{figure*}[!t]
\centering
\begin{subfigure}[]{0.45\linewidth}
\includegraphics[width=\textwidth]{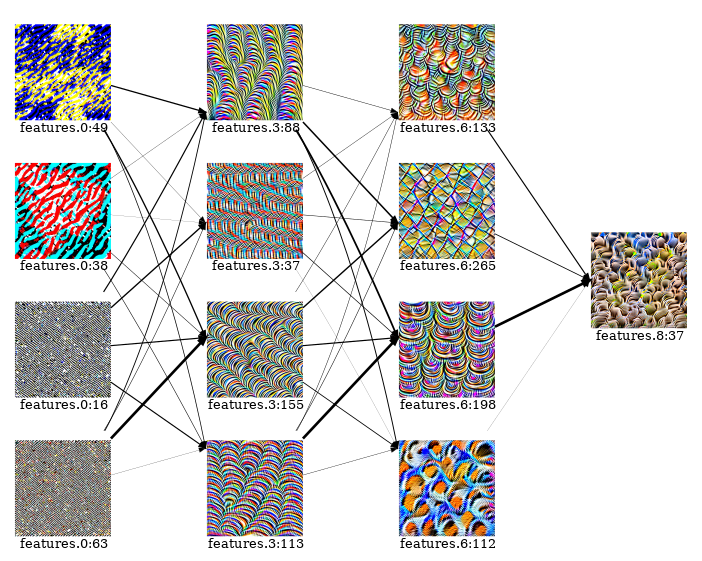}
\caption{Initial circuit.}
\end{subfigure}\hspace{.2cm}
\begin{subfigure}[]{0.45\linewidth}
\includegraphics[width=\textwidth]{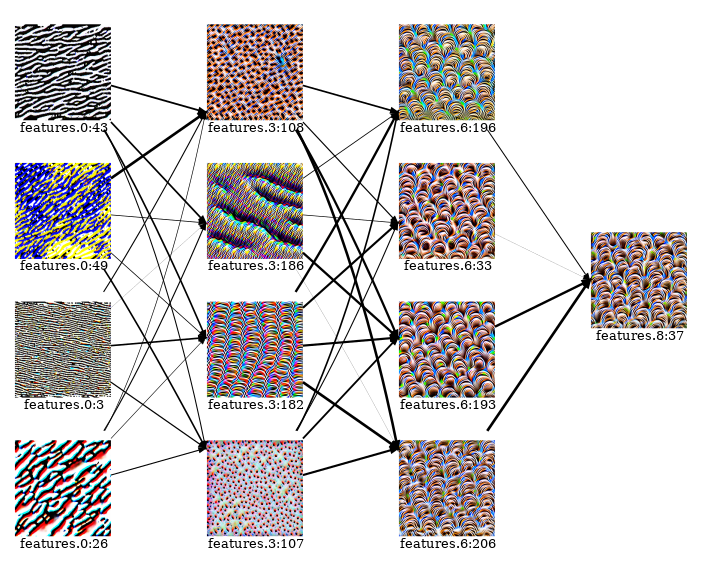}
\caption{Final circuit.}
\end{subfigure}
\caption{
 Illustration of the effectiveness of CircuitBreaker to manipulate the circuit: ablation on the sparsity level. 
}\label{fig:effectiveness_of_circuit_features_8_37_ablation}
\end{figure*}

\begin{figure*}[!t]
\centering
\begin{subfigure}[]{0.45\linewidth}
\includegraphics[width=\textwidth]{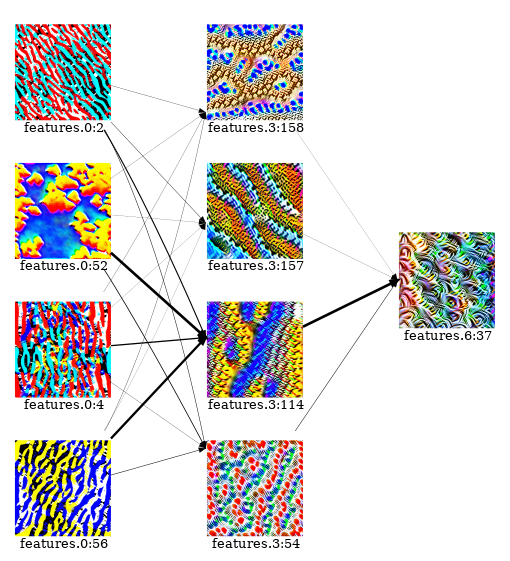}
\caption{Initial circuit.}
\end{subfigure}\hspace{.2cm}
\begin{subfigure}[]{0.45\linewidth}
\includegraphics[width=\textwidth]{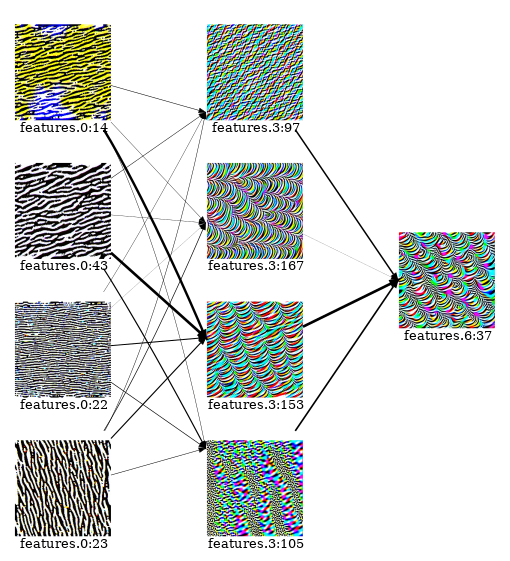}
\caption{Final circuit.}
\end{subfigure}
\caption{
 Illustration of the effectiveness of CircuitBreaker to manipulate the circuit: ablation on the sparsity level. 
}\label{fig:effectiveness_of_circuit_features_6_37_ablation}
\end{figure*}

Fig.~\ref{fig:effectiveness_of_circuit_features_10_37_ablation}, Fig.~\ref{fig:effectiveness_of_circuit_features_8_37_ablation} and Fig.~\ref{fig:effectiveness_of_circuit_features_6_37_ablation} show different circuits with different sparsity levels. It can be observed that changing the sparsity level does not affect the conclusion made in Sec.~\ref{sec:manipulation_channelcoercion}.

\begin{figure}[!t]
\centering
\includegraphics[width=.4\textwidth]{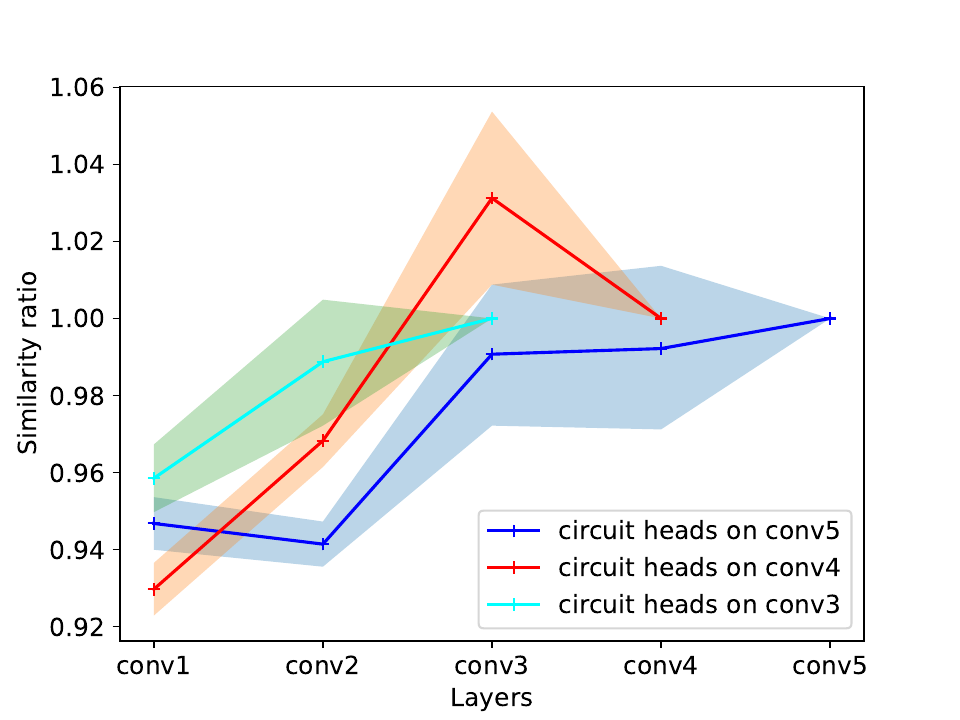}
\caption{Similarity ratio on synthetic feature visualization: ablation on the sparsity level.}
\label{fig:similarity_ration_alexnet_ablation}
\end{figure}

\begin{figure}[!t]
\centering
\includegraphics[width=.35\textwidth]{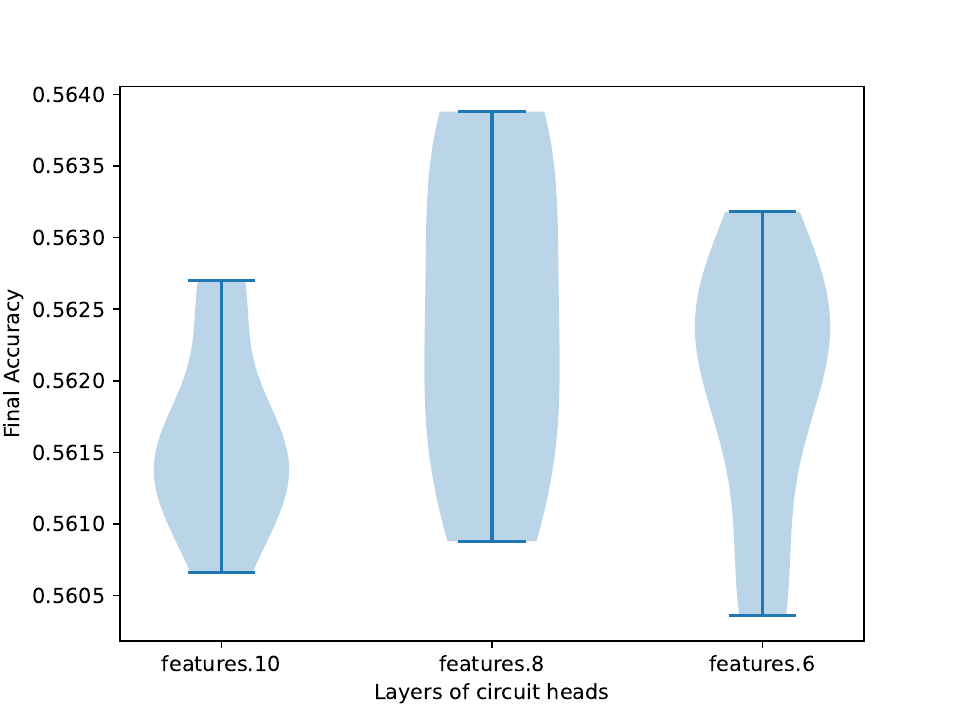}
\caption{Final accuracy after fine-tuning with CircuitBreaker on AlexNet. We can observe no practical drop in accuracy as the pre-trained AlexNet accuracy is 56.52\%.\vspace{-10pt}}
\label{fig:condensed_accuracy_alexnet}
\end{figure}

\subsection{Results for CircuitBreaker on ResNet-50}\label{app:channelCoercionResNet50}
\begin{figure*}[!t]
\centering
\begin{subfigure}[]{0.45\linewidth}
\includegraphics[width=\textwidth]{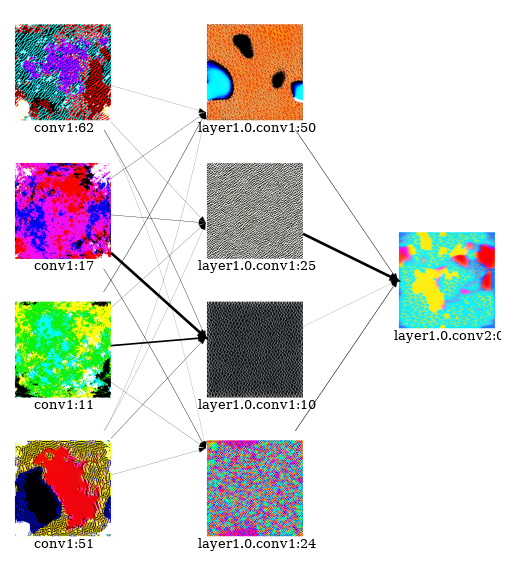}
\caption{Initial circuit.}
\end{subfigure}\hspace{.2cm}
\begin{subfigure}[]{0.45\linewidth}
\includegraphics[width=\textwidth]{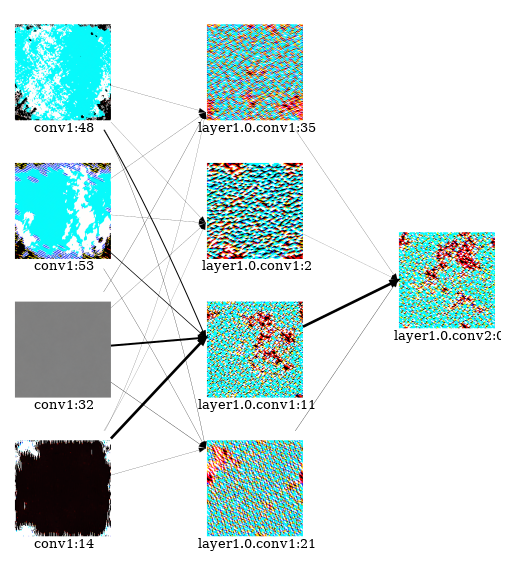}
\caption{Final circuit.}
\end{subfigure}
\caption{
 Illustration of the effectiveness of CircuitBreaker to manipulate the circuit: ablation on the sparsity level. 
}\label{fig:effectiveness_of_circuit_layer1_0_conv2_ablation_resnet50}
\end{figure*}
Fig.~\ref{fig:effectiveness_of_circuit_layer1_0_conv2_ablation_resnet50} shows ablation results on visual circuits on the ResNet-50 model, with a circuit head on layer1.0.conv2. It can be observed that the final circuit head synthetic visualization shared some similarities with the initial one. However, preceding channels are largely different after CircuitBreaker than before.

\vspace{2cm}
\subsection{Additional Results on DenseNet-201 and ResNet-152 for ProxPulse Attack}\label{app:densenet_resnet}
\begin{figure*}[!h]
\centering
\begin{subfigure}[]{0.3\linewidth}
\includegraphics[width=\textwidth]{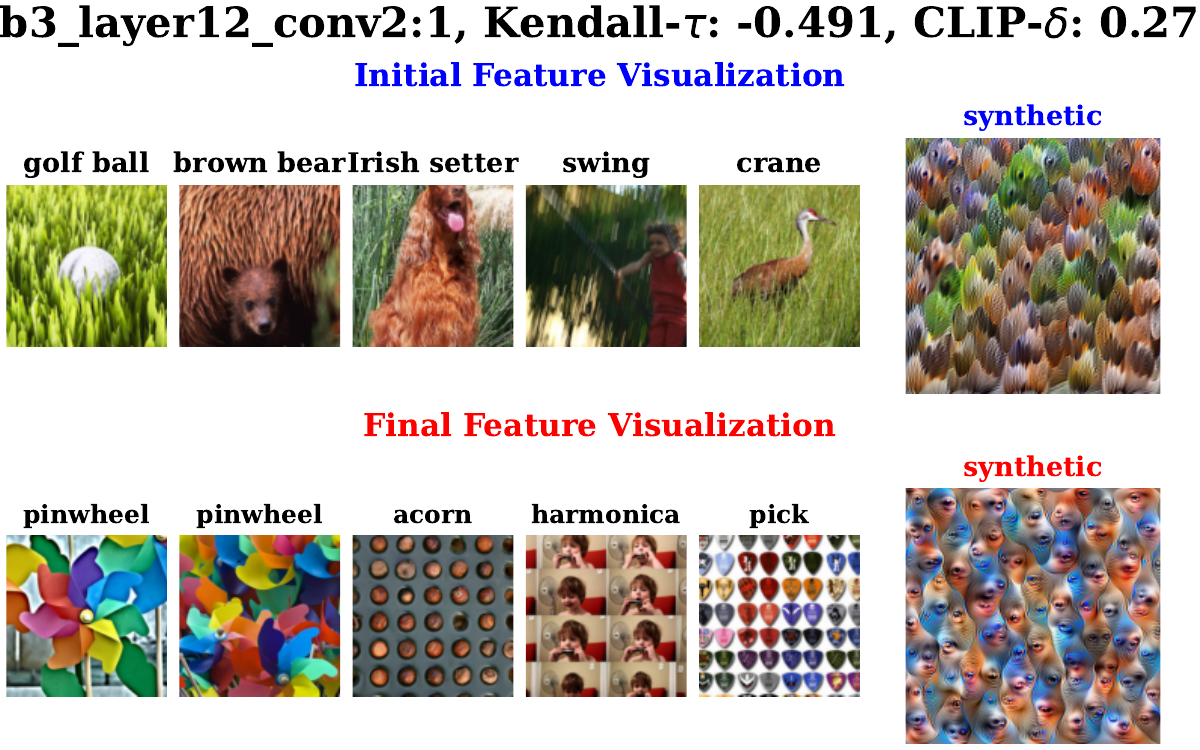}
\end{subfigure}\hspace{.2cm}
\begin{subfigure}[]{0.3\linewidth}
\includegraphics[width=\textwidth]{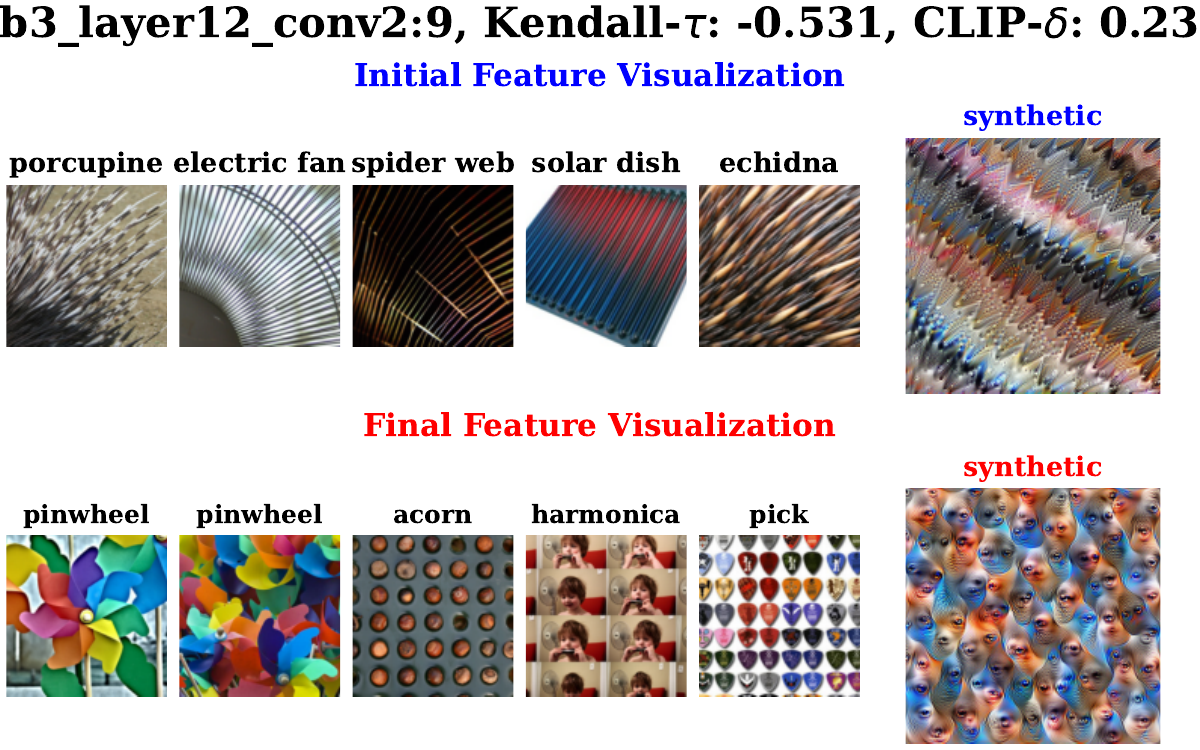}
\end{subfigure}\hspace{.2cm}
\begin{subfigure}[]{0.3\linewidth}
\includegraphics[width=\textwidth]{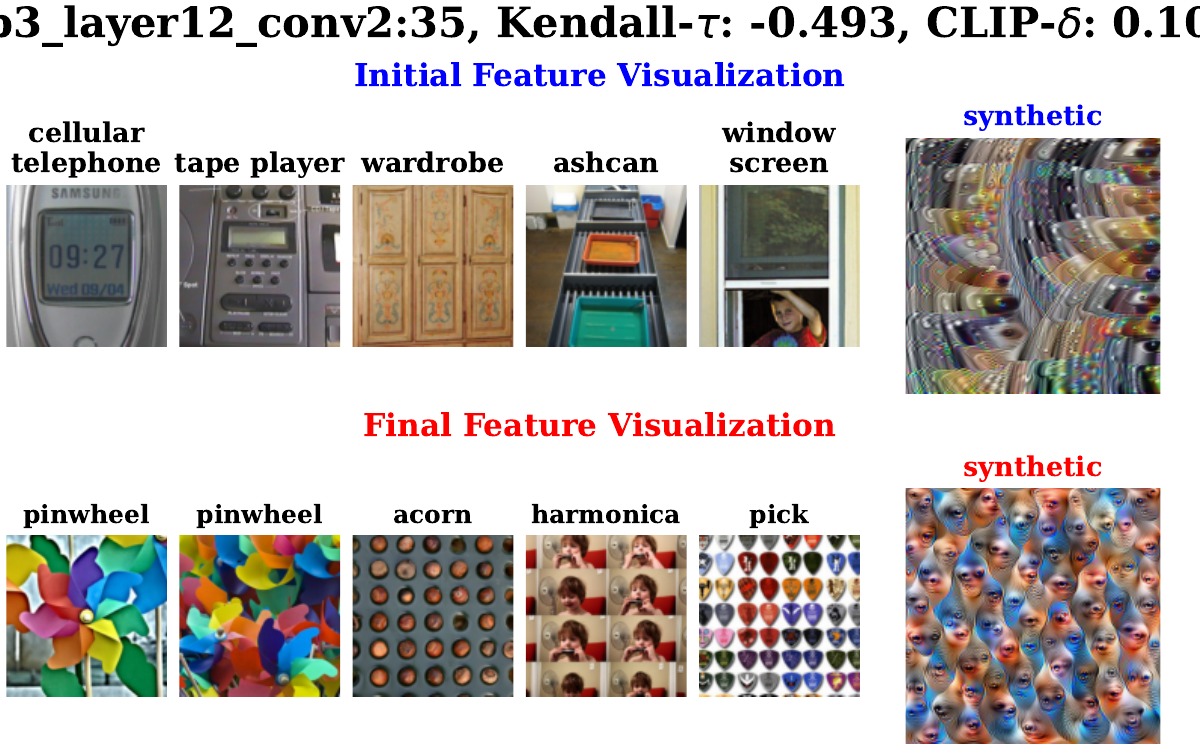}
\end{subfigure}\\
\begin{subfigure}[]{0.3\linewidth}
\includegraphics[width=\textwidth]{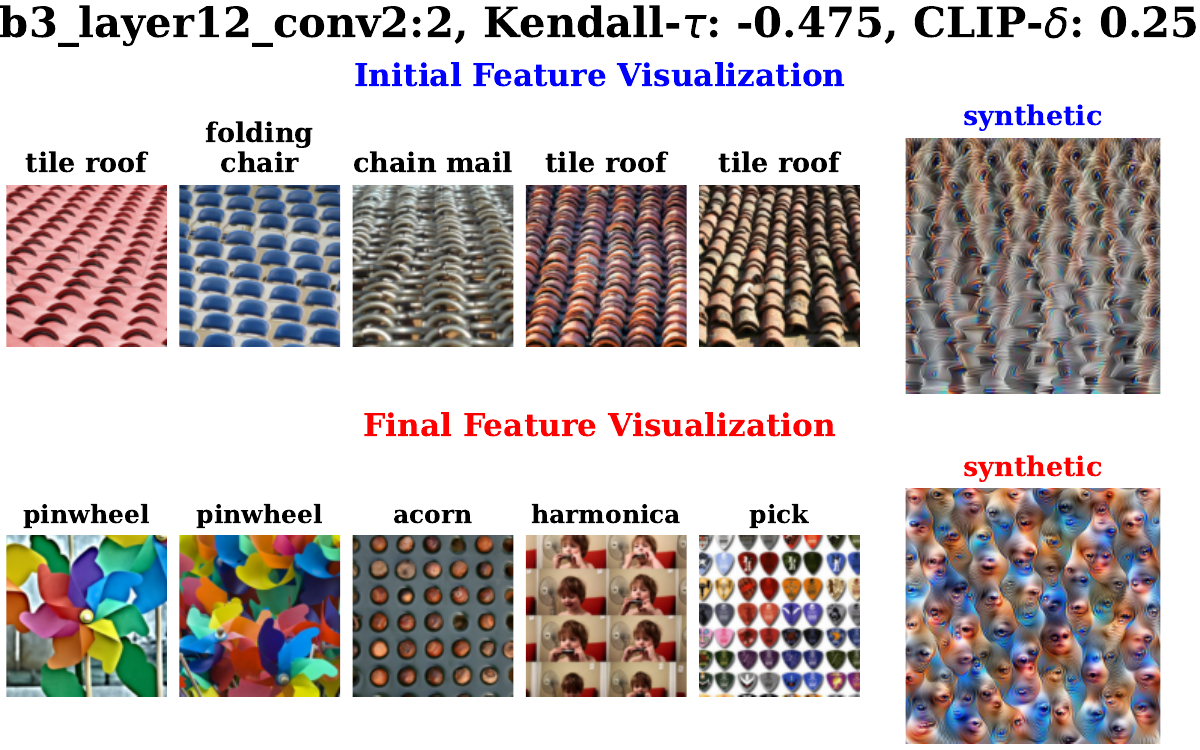}
\end{subfigure}\hspace{.2cm}
\begin{subfigure}[]{0.3\linewidth}
\includegraphics[width=\textwidth]{images/feat_vis_attack/densenet201/features_denseblock3_denselayer12_norm2_channel35_db3_layer12_conv2.pdf}
\end{subfigure}\hspace{.2cm}
\begin{subfigure}[]{0.3\linewidth}
\includegraphics[width=\textwidth]{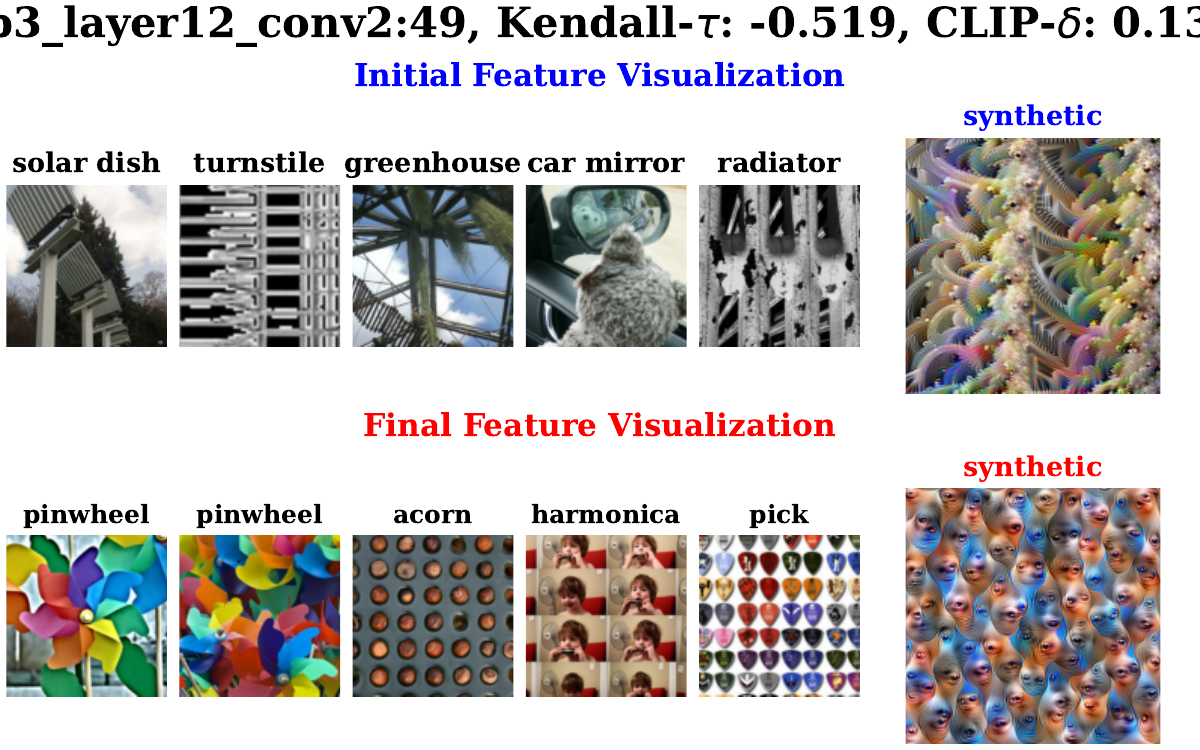}
\end{subfigure}\\
\caption{
 Illustration of the manipulability of both natural and synthetic feature visualization using ProxPulse on Block\_3\_Layer\_12\_conv2 of DenseNet201. The manipulated model has an accuracy of $76.52\%$ (vs $76.9\%$ for the initial model): the drop in accuracy is less than $0.4\%$. The first row (resp. second row) shows the natural initial (resp. final) feature visualization and initial (resp. final) synthetic feature visualizations. On the image title, we report the corresponding metrics to evaluate change in top activating inputs. One can observe that both natural and synthetic feature visualization have completely changed, to very similar images for the synthetic one. Target images are shown in Fig.~\ref{fig:target_images2}.
}\label{fig:densenet201}
\end{figure*}

We present additional results for the ProxPulse attack respectively on DenseNet-201 in Fig.~\ref{fig:densenet201} and on ResNet-152 in Fig.~\ref{fig:resnet152}. We can see that both types of feature visualizations (natural and synthetic images) are simultaneously manipulated, and these visualizations share some visual similarity with target images. 

\newpage

\begin{figure*}[!t]
\centering
\begin{subfigure}[]{0.3\linewidth}
\includegraphics[width=\textwidth]{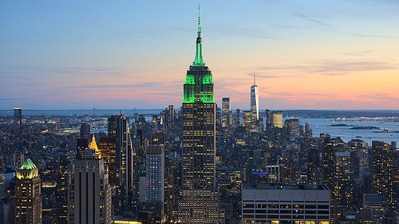}
\end{subfigure}\hspace{.2cm}
\begin{subfigure}[]{.24\linewidth}
\includegraphics[width=\textwidth]{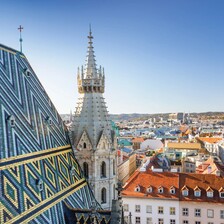}
\end{subfigure}\hspace{.2cm}
\caption{
Target images ($\mathcal{D}_{\text{fool}}$) for ProxPulse on ResNet-152: NewYork and Vienna images taken from Wikipedia and Cntraveller websites. 
}
\label{fig:target_images2}
\end{figure*}

\begin{figure*}[!h]
\centering
\begin{subfigure}[]{0.3\linewidth}
\includegraphics[width=\textwidth]{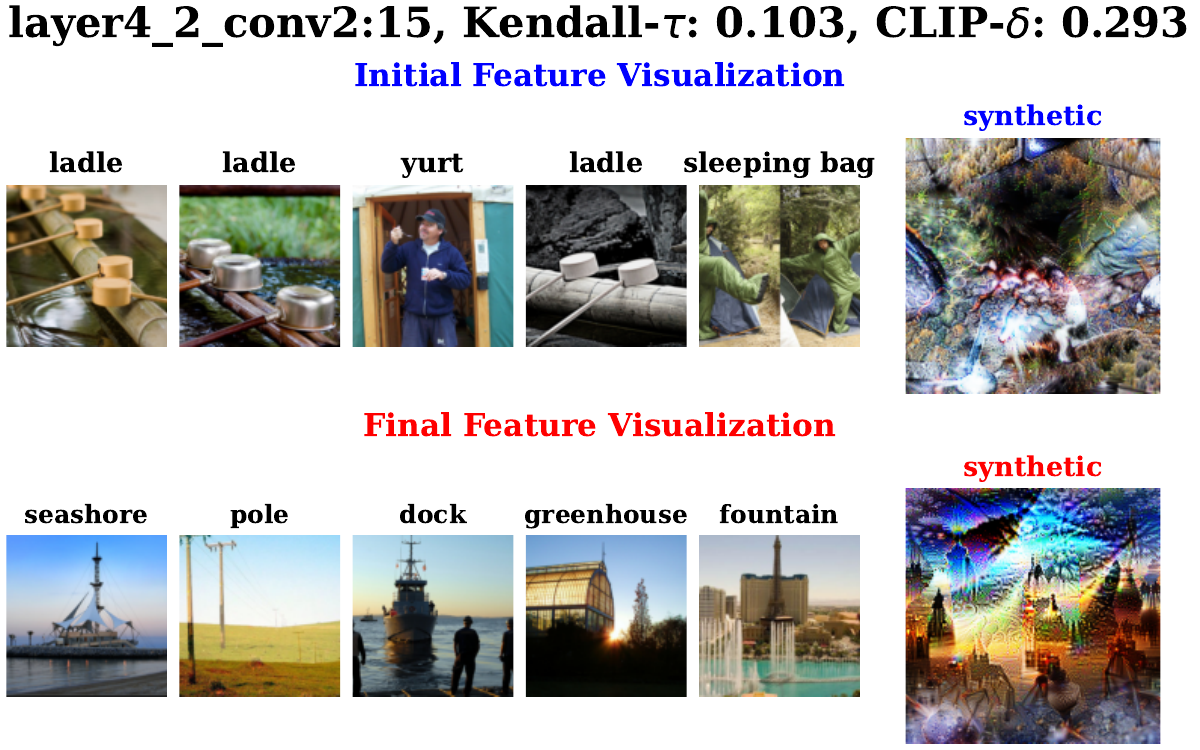}
\end{subfigure}\hspace{.2cm}
\begin{subfigure}[]{0.3\linewidth}
\includegraphics[width=\textwidth]{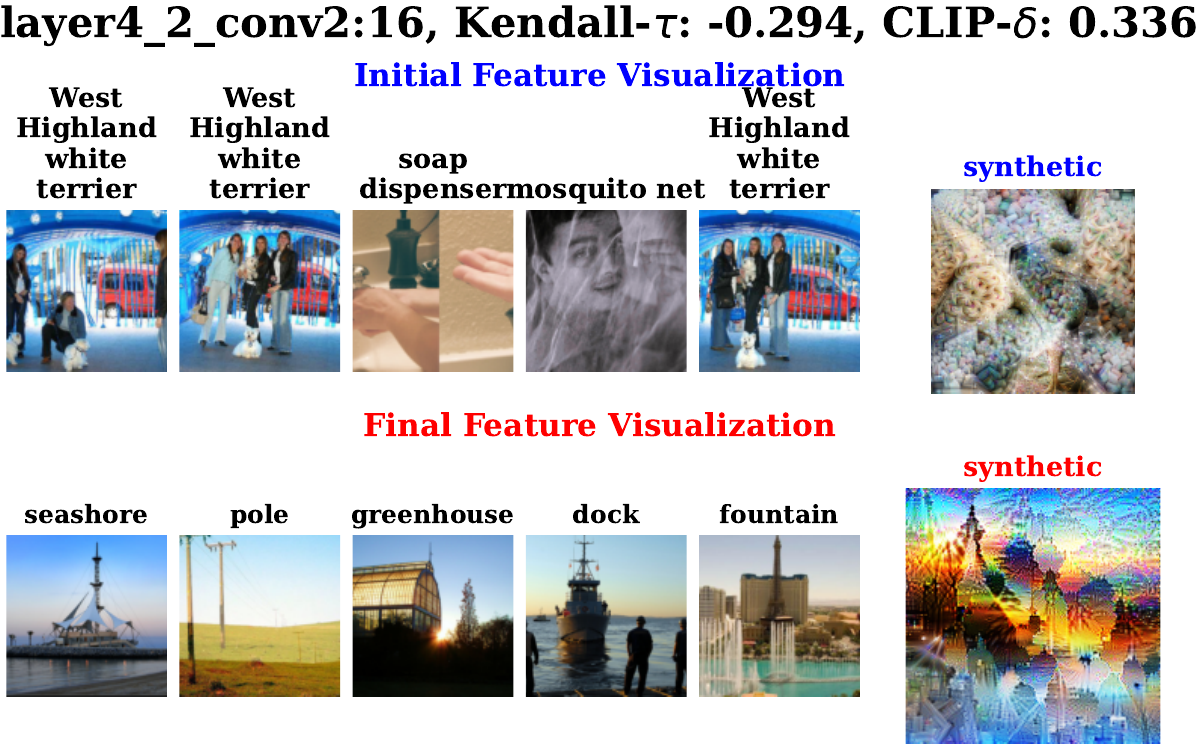}
\end{subfigure}\hspace{.2cm}
\begin{subfigure}[]{0.3\linewidth}
\includegraphics[width=\textwidth]{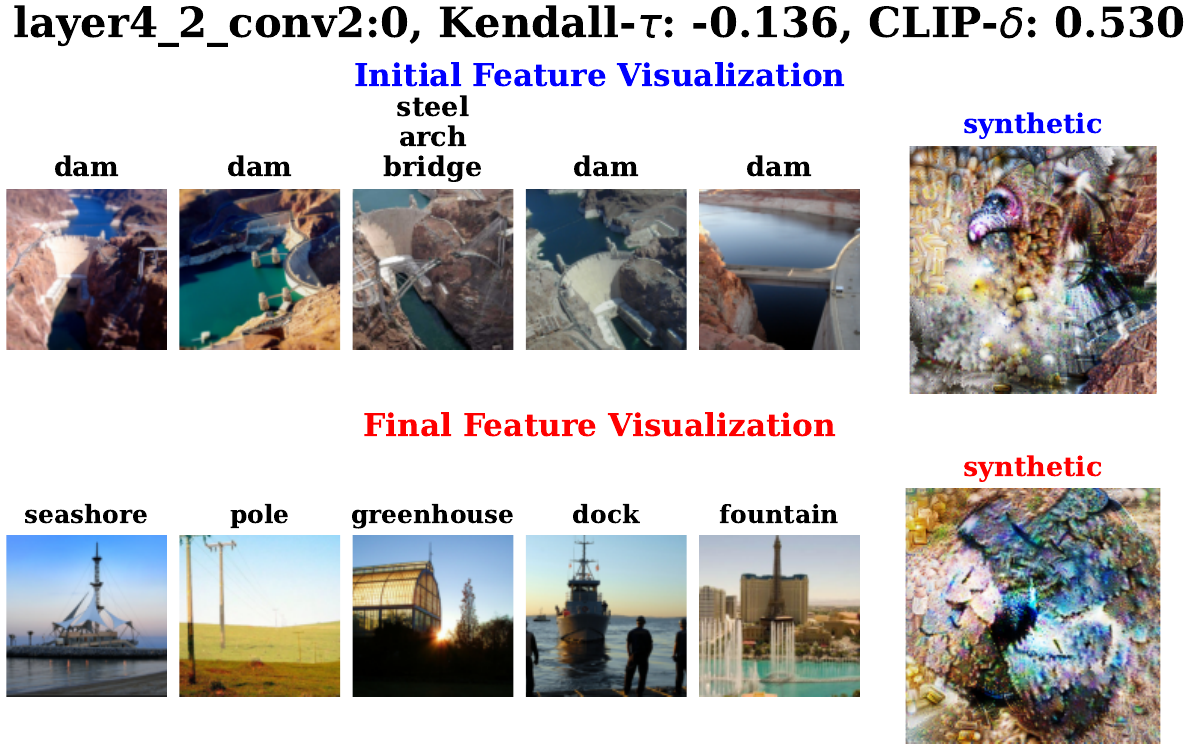}
\end{subfigure}\\
\begin{subfigure}[]{0.3\linewidth}
\includegraphics[width=\textwidth]{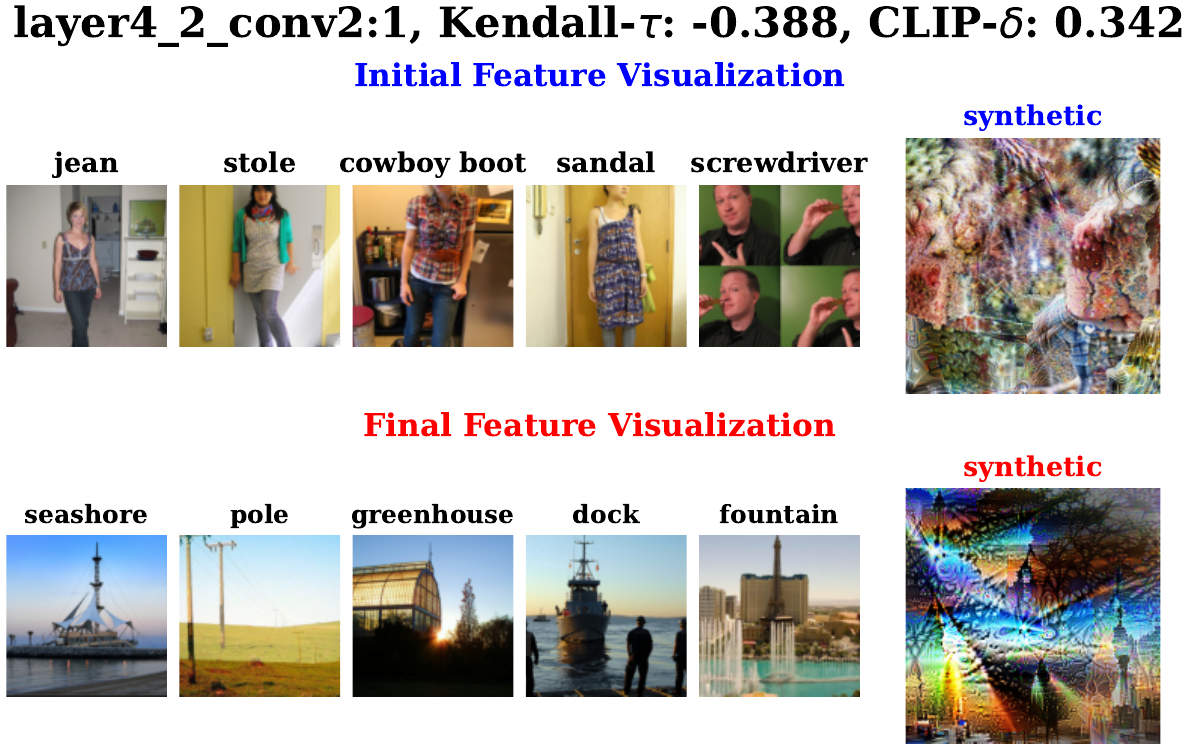}
\end{subfigure}\hspace{.2cm}
\begin{subfigure}[]{0.3\linewidth}
\includegraphics[width=\textwidth]{images/feat_vis_attack/resnet152/layer4_2_bn2_vienna_channel16_layer4_2_conv2.pdf}
\end{subfigure}\hspace{.2cm}
\begin{subfigure}[]{0.3\linewidth}
\includegraphics[width=\textwidth]{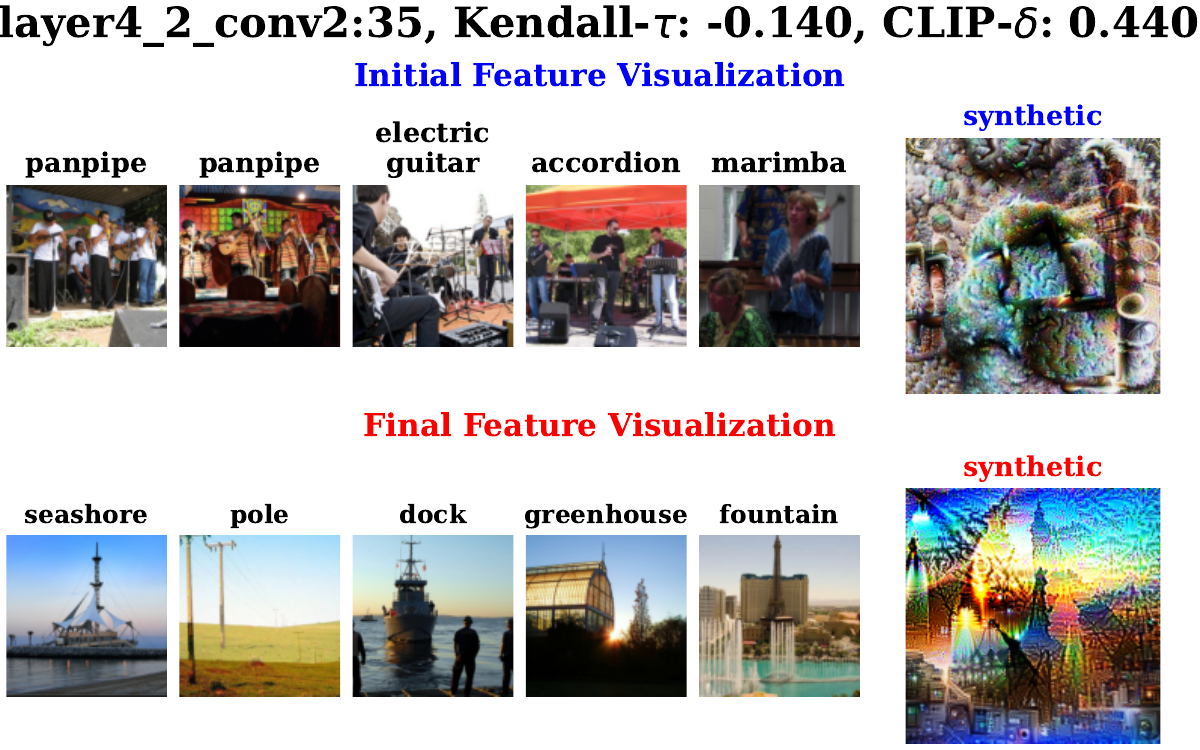}
\end{subfigure}\\
\caption{
 Illustration of the manipulability of both natural and synthetic feature visualization using ProxPulse on Layer\_4\_2\_conv2 of ResNet-152. The manipulated model has an accuracy of $82.27\%$ (vs $82.284 \%$ for the initial model): the drop in accuracy is less than $0.1\%$. The first row (resp. second row) shows the natural initial (resp. final) feature visualization and initial (resp. final) synthetic feature visualizations. On the image title, we report the corresponding metrics to evaluate change in top activating inputs. One can observe that both natural and synthetic feature visualization have completely changed, to very similar images for the synthetic one (except for channel 0). Target images are shown in Fig.~\ref{fig:target_images2}.
}\label{fig:resnet152}
\end{figure*}

\newpage

\subsection{Additional Results on Simulatenously Fooling Several Circuits with Feature Heads on Features:8 (Conv4) of AlexNet}\label{app:simultaneous_manipulation}
In this section, we simultaneously run the CircuitBreaker manipulation on the first 30 circuits with feature heads on features.8 (conv4) of AlexNet. 

\begin{figure*}[!h]
\centering
\begin{subfigure}[]{0.33\linewidth}
    \includegraphics[width=\textwidth]{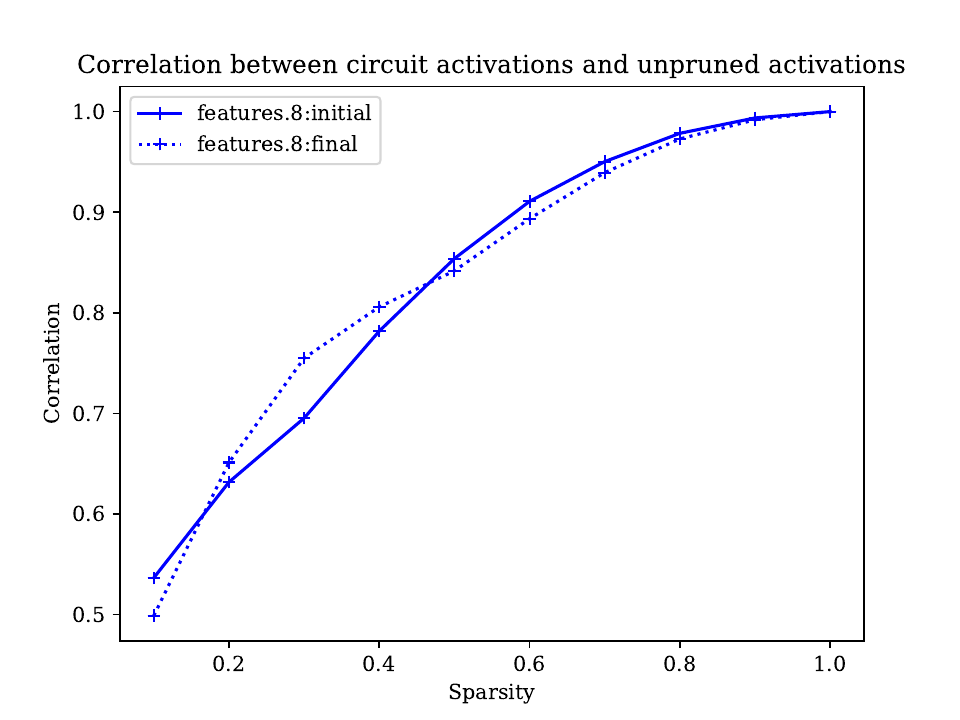}
\caption{Pearson correlation between activations on circuits (with pruning) for the (i) considered model and (ii) the initial model without structured pruning, i.e., with sparsity 1.}
\label{fig:circuit_attack_alexnet_pearson_correlation_rebuttal}
\end{subfigure}\hfill
\begin{subfigure}[]{0.33\linewidth}
\includegraphics[width=\textwidth]{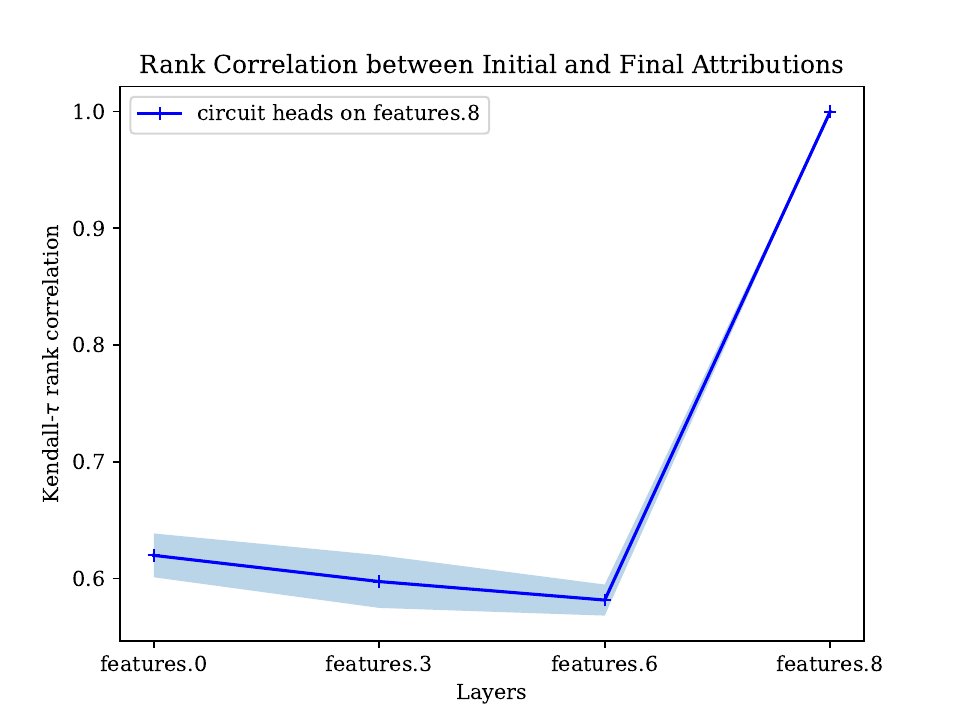}
\caption{Rank correlation between (i) kernel attribution scores for circuits on (i) the initial model and (ii) on the fine-tuned model with CircuitBreaker.}
\label{fig:circuit_attack_alexnet_rank_correlation_rebuttal}
\end{subfigure}
\begin{subfigure}[]{0.32\linewidth}
\vspace{-26.5pt}
\includegraphics[width=\textwidth]{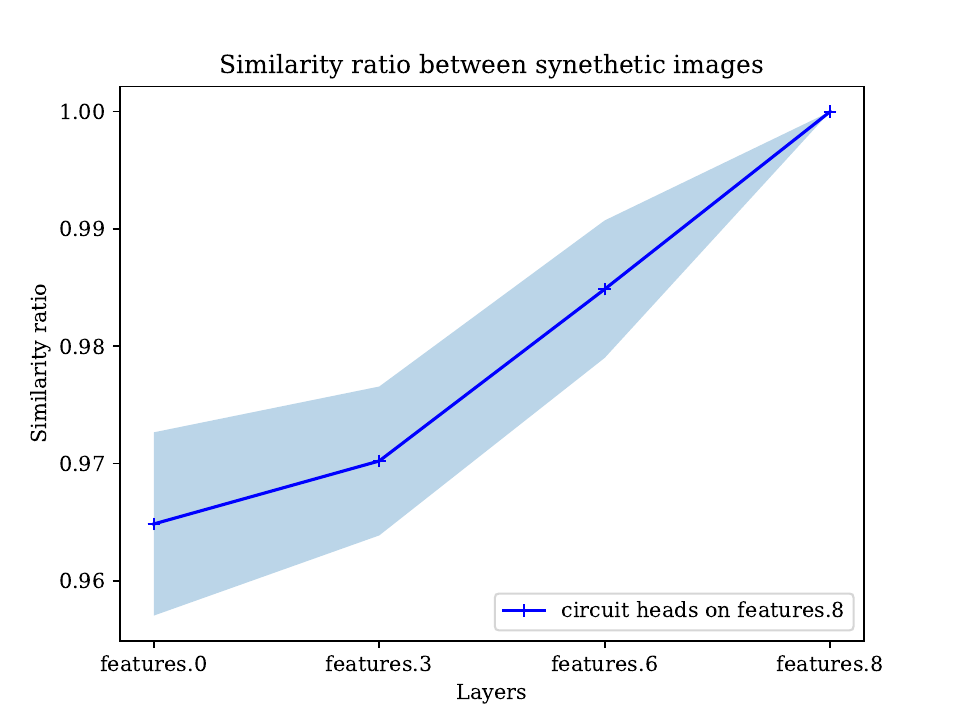}
\caption{Similarity ratio with CircuitBreaker.}
\label{fig:similarity_ratio_rebuttal}
\end{subfigure}
\vspace{-5pt}
\caption{
Results obtained when simultaneously fooling 30 circuits with heads on  features.8 (conv4) of AlexNet. \vspace{-8pt}
}\label{fig:rebuttal_condensed_results}
\end{figure*}

According to the criteria evaluated in Section 5.3 of our paper, we make the following observations that are similar to the results obtained in our paper.
First, by looking at Fig.~\ref{fig:circuit_attack_alexnet_pearson_correlation_rebuttal}, we also observe high functional preservation on moderate to higher sparsity. 

Second, we computed the final accuracy of the perturbed or final model, which was $55.83\%$ (a drop of less than $.7\%$ as the initial accuracy of AlexNet is $56.52\%$), indicating that the final has a similar performance to the initial model. 

Third, from Fig.~\ref{fig:circuit_attack_alexnet_rank_correlation_rebuttal}, we observe that Kendall-$\tau$ rank for layers before the feature heads are around $.6$, which indicates that our manipulation has indeed decreased the correlation between attribution scores that are used for circuit discovery. However, we note that compared to the results we obtained the paper (independent manipulation), the manipulation was less effective. 

Fourth, as seen in Fig.~\ref{fig:rebuttal_condensed_results}, we observe that the similarity ratio is usually less than $1$. This indicates that the 
synthetic feature visualizations have changed in the manipulated circuits. Note that the similarity ratio which is equal to $1$ on feature heads means that the synthetic feature visualizations have almost not changed.

\begin{figure*}[!t]
\centering
\begin{subfigure}[]{0.41\linewidth}
\includegraphics[width=\textwidth]{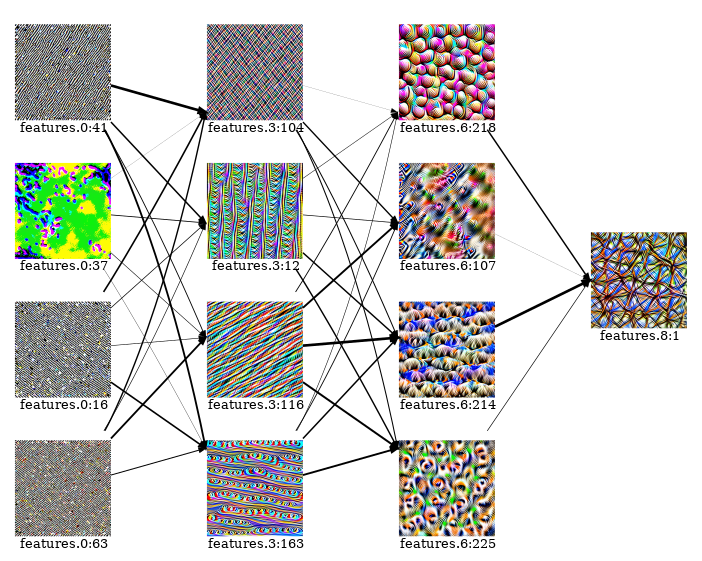}
\caption{With initial model.}
\end{subfigure}\hspace{1.3cm}
\begin{subfigure}[]{0.41\linewidth}
\includegraphics[width=\textwidth]{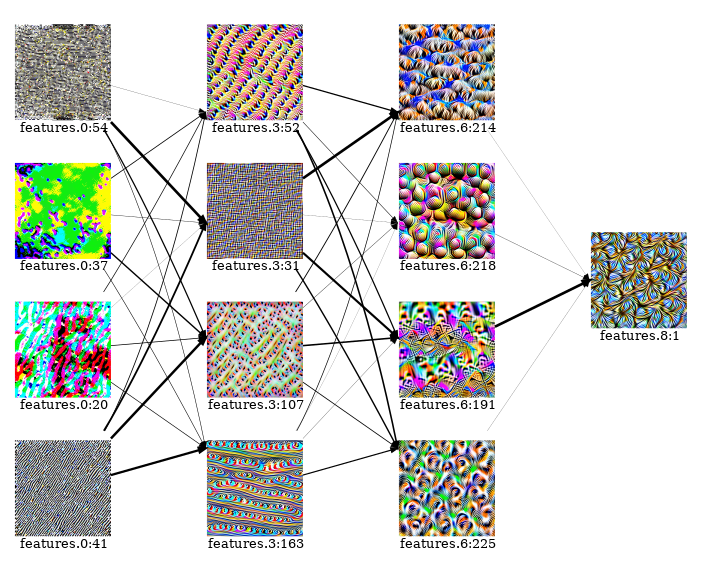}
\caption{After CircuitBreaker.\vspace{-10pt}}
\end{subfigure}
\vspace{-5pt}
\caption{
 Illustration of the effectiveness of CircuitBreaker to manipulate visual circuits on features:8 (conv4) of AlexNet. We observe that the circuit visualization is severely distorted while the network outputs change minimally. \vspace{-15pt}
}\label{fig:effectiveness_of_circuit_features_8_channel_1_rebuttal}
\end{figure*}

\begin{figure*}[!t]
\centering
\begin{subfigure}[]{0.41\linewidth}
\includegraphics[width=\textwidth]{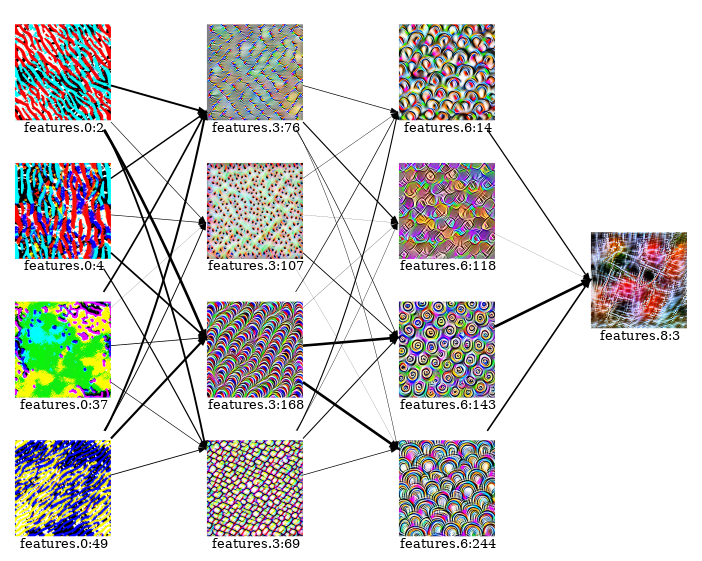}
\caption{With initial model.}
\end{subfigure}\hspace{1.3cm}
\begin{subfigure}[]{0.41\linewidth}
\includegraphics[width=\textwidth]{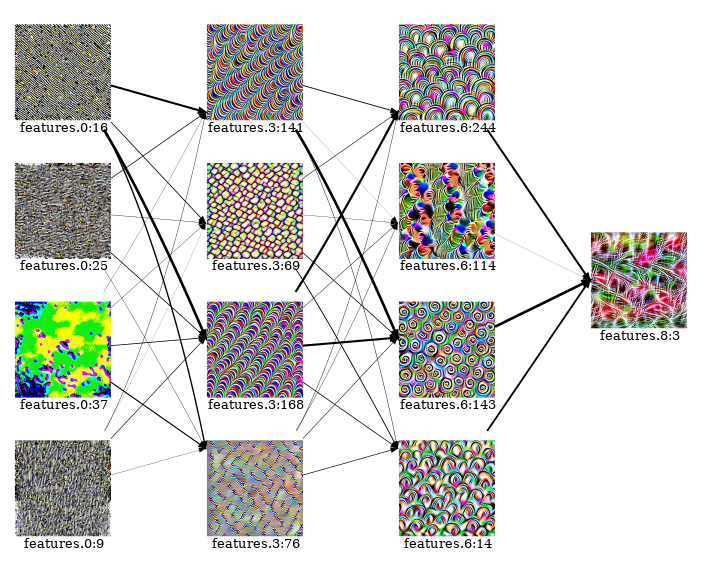}
\caption{After CircuitBreaker.\vspace{-10pt}}
\end{subfigure}
\vspace{-5pt}
\caption{
 Illustration of the effectiveness of CircuitBreaker to manipulate visual circuits on features:8 (conv4) of AlexNet. We observe that the circuit visualization is severely distorted while the network outputs change minimally. \vspace{-15pt}
}\label{fig:effectiveness_of_circuit_features_8_chanel_3_rebuttal}
\end{figure*}

Finally, we depicted in Fig.~\ref{fig:effectiveness_of_circuit_features_8_channel_1_rebuttal} and Fig.~\ref{fig:effectiveness_of_circuit_features_8_chanel_3_rebuttal} two circuits that were part of the simultaneously manipulated circuits. We observe that while the first circuit in Fig.~\ref{fig:effectiveness_of_circuit_features_8_channel_1_rebuttal} has undertaken some changes (the most effective way is to compare layer by layer in particular features:3), we observe that the second one in Fig.~\ref{fig:effectiveness_of_circuit_features_8_chanel_3_rebuttal}  has marginally changed.

\begin{figure}[!h]
\centering
\includegraphics[width=.5\textwidth]{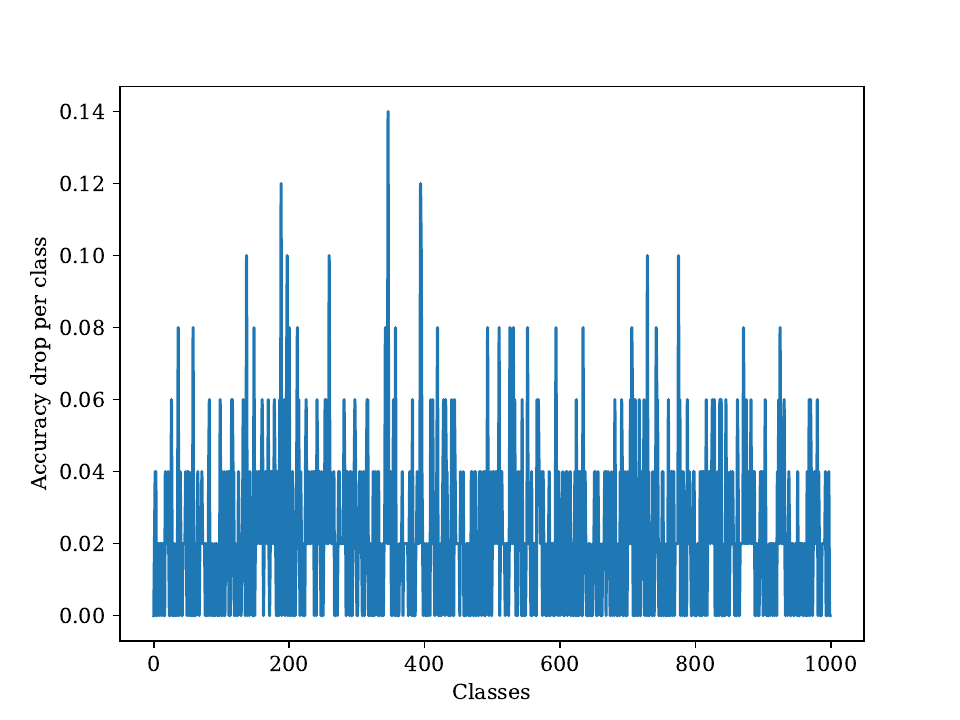}
\caption{Accuracy Drop Per Class. We do not observe a significant drop only in a few classes.}
\label{fig:accuracy_drop}
\end{figure}

\end{document}